\documentclass[preprint,5p,times,twocolumn]{elsarticle}

\usepackage{amssymb}
\usepackage{graphicx}%
\usepackage{adjustbox}
\usepackage{cite}
\usepackage{amsmath,amsfonts}%
\usepackage{amsthm}%
\usepackage{makecell}
\usepackage{bbding}
\usepackage{float}
\usepackage[title]{appendix}%
\usepackage{colortbl}
\usepackage{textcomp}%
\usepackage{siunitx}
\usepackage[dvipsnames]{xcolor} % <<<
\usepackage{tikz}
\usepackage{array}
\usepackage{manyfoot}%
\usepackage{afterpage} 
\usepackage{longtable}
\usepackage{booktabs}%

\usepackage{algorithm}%
\usepackage{algorithmicx}%
\usepackage{algpseudocode}%
\usepackage{amssymb}% http://ctan.org/pkg/amssymb
\usepackage{pifont}% http://ctan.org/pkg/pifont
\usepackage{listings}%
\usepackage{comment}
\usepackage[utf8]{inputenc}
\usepackage{array}
\usepackage{multirow}
\newcommand{\xmark}{\ding{55}}%

\usepackage{hyperref}

\journal{Information Fusion}

\begin{document}

\begin{frontmatter}



\title{Trajectory Prediction for Autonomous Driving: Progress, Limitations, and Future Directions}

\author[a,d]{Nadya Abdel Madjid}
\author[a,d]{Abdulrahman Ahmad}
\author[a,d]{Murad Mebrahtu}
\author[a,d]{Yousef Babaa}
\author[a,d]{Abdelmoamen Nasser}
\author[a,d]{Sumbal Malik}
\author[b]{Bilal Hassan}
\author[a]{Naoufel Werghi}
\author[c,d]{Jorge Dias}
\author[a,d]{Majid Khonji}

\affiliation[a]{organization={Khalifa University, Computer Science}, city={Abu Dhabi}, country={UAE}}
\affiliation[b]{organization={New York University Abu Dhabi, Computer Science}, city={Abu Dhabi}, country={UAE}}
\affiliation[c]{organization={Khalifa University, Computer and Information Engineering}, city={Abu Dhabi}, country={UAE}}
\affiliation[d]{organization={Khalifa University, KUCARS-KU Center for Autonomous Robotic Systems, Department of Computer Science}, city={Abu Dhabi}, country={UAE}}            

\begin{abstract}
As the potential for autonomous vehicles to be integrated on a large scale into modern traffic systems continues to grow, ensuring safe navigation in dynamic environments is crucial for smooth integration. To guarantee safety and prevent collisions, autonomous vehicles must be capable of predicting the future trajectories of surrounding traffic agents. Over the past decade, significant efforts from both academia and industry have been dedicated to designing solutions for precise trajectory forecasting. These efforts have produced a vast range of approaches, raising questions about how they differ, which challenges they address, and what limitations still remain in trajectory prediction. This paper reviews a substantial portion of recent trajectory prediction methods, proposing a taxonomy to classify existing solutions. It also provides a general overview of the prediction pipeline, covering input and output modalities, modeling features, performance evaluation metrics, and existing prediction paradigms. In addition, the paper discusses active research topics within the forecasting realm, addresses the posed research questions, and highlights the remaining research gaps and challenges.
\end{abstract}


%\item An extensive overview of the prediction pipeline, which includes discussions of input modalities, modeling features and approaches, output modalities, evaluation metrics, and prediction paradigms.

%\item Taxonomy of modeling approaches, offering a structured classification of methods. Each category includes a comparative table summarizing key models and their characteristics for quick reference.

%\item Literature review of active research areas in the trajectory prediction domain, covering topics such as uncertainty estimation, integration of driving knowledge, planning-conditioned prediction, vision-language model (VLM)-guided prediction models, and collaboration between vehicles.

%\item An open code repository featuring two prediction pipelines to facilitate experimentation, benchmarking, and the development of new methods.


\end{frontmatter}


\section{Introduction}
\label{sec:intro}

\begin{figure*}
\centering
   \includegraphics[width=1.0\textwidth]{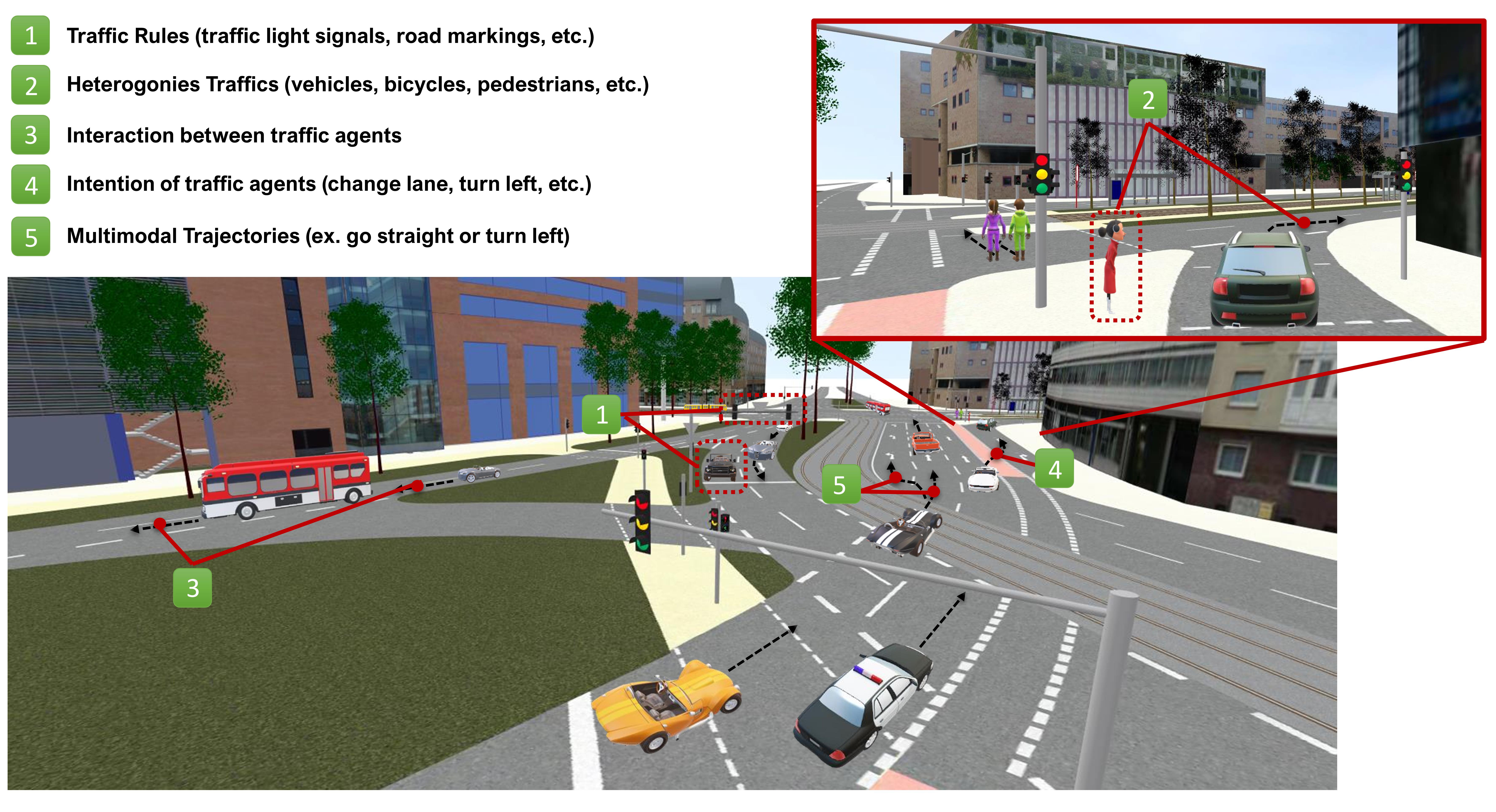}
   \caption{Factors impacting Trajectory Prediction: 1) cars waiting at a red light, demonstrating adherence to traffic rules, 2) a pedestrian waiting to cross the road, indicative of heterogeneous traffic, as the pedestrian waits for the car to pass, 3) interaction between vehicles, such as a car potentially overtaking a bus if travelling at a higher speed, since the bus limits its way, 4) detecting intention, like a car planning to take a right shoulder and turn right, 5) multi-modal trajectories, exemplified by a car having several options at a given point, such as going straight or changing to the left lane.}
   \label{fig:intro_fig}
\end{figure*}

The era of autonomous driving is on the verge of revolutionizing the modern transportation system. Automotive companies like Tesla, Waymo, and Cruise are pioneers in today’s self-driving vehicle technology. The practical, innovative solutions developed by these companies, in conjunction with academic research, have led to rapid advancements in perception, prediction, planning, and control algorithms for autonomous driving. Predicting the future trajectories of surrounding traffic participants is essential for safe navigation and serves as a core task of the prediction module in the autonomous stack. The predicted trajectories are passed as input to planning algorithms to eliminate paths that may lead to potential collisions and to select a safe course of action. Due to the diversity of traffic scenarios, road layouts, interactions between traffic participants, various maneuvers, inherent randomness in human actions, and computational time requirements, the trajectory forecasting problem presents a range of challenges (refer to Figure \ref{fig:intro_fig} for factors impacting trajectory forecasting).

Autonomous vehicles (AVs) operate in heterogeneous traffic environments, and their prediction modules must account for the motion dynamics of all traffic participants, including vehicles, cyclists, pedestrians, and motorbikes. Among these, the motion of vehicles is influenced not only by their past motion, road topology, and physical constraints, but also by the behavior of nearby agents. As human drivers intuitively understand such interdependencies, e.g., anticipating that a neighboring car being cut off may respond with a sudden maneuver, AVs must be equipped with a similar interaction awareness mechanism. To equip AVs with this capability, the prediction module needs to build a representation of each target agent’s surroundings and account for the interdependencies among traffic participants’ decisions. 

As for human motion, beyond the complexity of interaction and inter-dependency, predicting human trajectories involves accounting for complex human behavior. Pedestrian motion is influenced not only by past trajectory, road topology and the actions of surrounding agents but also by social norms, destination intent, and inherent behavioral randomness, e.g., stopping abruptly to check a phone, changing direction to greet someone, crossing at undesignated locations, or hesitating mid-crossing due to uncertainty or distraction. For precise pedestrian trajectory forecasting, impacting factors can be inferred from scene contextual information and general knowledge, e.g., a child and an adult exhibit different motion patterns. However, incorporating scene context increases computational demand, as it requires additional architecture blocks to extract relevant semantic and situational cues.

Additionally, prediction models that incorporate traffic rules can benefit from a reduced set of plausible future trajectories that comply with these constraints. However, human drivers may violate traffic rules by over-speeding, changing lanes over a solid line, turning from a non-designated lane, ignoring stop signs, or even running a red light. Returning to the aspect of human behavioral randomness, prediction models must account for the possibility of such violations to ensure robust and reliable performance in real-world deployment. Lastly, trajectory prediction is inherently multimodal, i.e., the same past trajectory can lead to multiple plausible futures. For example, a vehicle approaching an intersection in a lane marked for both straight and left-turn movements may proceed forward or make a left turn. Ideally, the prediction model should capture this multimodality and produce outputs in a format that can be easily integrated into downstream planning modules.

%%%%%%%%%%%%%%%%%%%%%%%%%%%%%%%%%%%%%%%%%%%%%%%%%%%%%

Thus, the trajectory prediction task is shaped by a set of impacting factors that must be captured from the input data and represented in the model. To address these challenges, the research community has focused on designing prediction models capable of learning complex motion patterns and incorporating relevant contextual cues. Among the top-performing solutions is Trajectron++ \citep{10.1007/978-3-030-58523-5_40}, which integrates environmental context through a graph-structured recurrent architecture and Spatio-Temporal Transformer Network (S2TNet) \citep{pmlr-v157-chen21a}, leveraging transformer layers to capture both spatial and temporal dependencies. Further, from the intersection between academia and industry, the MultiPath++ model \citep{10.1109/ICRA46639.2022.9812107}, released by the Waymo research group, employs multi-context gating fusion and dynamically learns reference points (anchors) to support goal-oriented predictions. Collectively, these models combine diverse techniques, such as graph-based interaction modeling, transformer architectures, and dynamic goal sampling, leading to more accurate and robust trajectory predictions. 

On the other side, the commercial research prioritizes real-time execution, cost-effectiveness, handling uncertainty at all stages of the autonomous stack and generalization for robust operating. This leads each company to use its own combination of sensors, compatible with the set of algorithms it employs. Tesla’s prediction algorithm is an end-to-end vision-based deep learning model\footnote{\url{https://www.tesla.com/support/autopilot}}, forecasting trajectories based on the visual sequences and aggregated contextual information over time. The model continuously learns from the vast amount of data collected by Tesla's global fleet of vehicles. In contrast, Waymo utilizes a combination of high-fidelity sensors (LiDAR, cameras, and radar). The sensor data is processed by the perception module and then passed to motion prediction and planning components. Waymo adopts a modular structure, ensuring that each component can be thoroughly tested independently and then integrated into a single pipeline\footnote{\url{https://waymo.com/research/}}. It is challenging to determine the specific algorithms employed in real operations, but Waymo has a strong R\&D focus, actively publishing its latest research findings. Aurora company focuses on autonomous solutions for trucks serving logistics purposes. Its prediction module\footnote{\url{https://blog.aurora.tech/}} performs joint object detection and trajectory forecasting, combining learning-based methods with rule-based constraints to align predictions with driving rules. To ensure safety across a wide range of scenarios, from highway driving delivering cargo between cities to urban logistics, Aurora’s system is tested extensively through a combination of virtual testing suites and real-world operations. Each company’s strategy reflects a balance between predictive accuracy and operational constraints, tailored to the specific purpose of their vehicles. 

On the testing side of autonomous driving algorithms, two robust open-source platforms provide an environment for evaluation: Autoware\footnote{\url{https://github.com/autowarefoundation/autoware}} and Apollo\footnote{\url{https://github.com/ApolloAuto/apollo}}. Autoware’s perception stack includes functionalities such as object recognition (2D/3D detection using LiDAR, cameras, and radar), obstacle segmentation (distinguishing drivable areas via point cloud processing), occupancy grid mapping (representing occupied and free spaces), and traffic light recognition (detecting light states and directions). However, it lacks an integrated prediction module. Starting from release 5.0, Apollo incorporates an integrated prediction module\footnote{\url{https://github.com/ApolloAuto/apollo/blob/master/modules/prediction/README.md}}, supporting a range of predictors: Free Move Predictor for non-lane-constrained obstacles, Single Lane Predictor for vehicles on a single lane, and Lane Sequence Predictor for multi-lane and behavior-based trajectory predictions. Later versions, such as 6.0.0 and 7.0.0, introduced advanced prediction models, including a semantic map-based pedestrian prediction model and the Inter-TNT prediction model with interactive prediction and planning evaluation. Thus, Apollo provides a comprehensive platform for testing prediction algorithms in the closed-loop setting --- within the full autonomous driving stack. For iterative integration, a lightweight platform that couples perception and prediction is needed for rapid experimentation. It can expedite the development of error-handling mechanisms and help standardize the evaluation of prediction models under perception noise. Adding a joint learning feature would enable the prediction module to adapt to perception uncertainties while providing feedback to improve perceptual representations.

For researchers entering the field of trajectory prediction, the volume of conducted works and proposed methods might be overwhelming, raising questions about how to categorize and classify existing approaches and which areas still offer room for improvement. For professionals experienced in the realm, keeping track of new emerging designs that lead to significant performance improvements, considering how rapidly the field is evolving, might be challenging. To assist both groups, this review covers a wide range of topics relevant to the trajectory prediction field, provides comprehensive overviews, and lists relevant references for readers to fetch additional information, which is either out of scope or briefly mentioned in the paper. In this review, we pose the following research questions:

\begin{itemize}
\item \textbf{RQ1:} \textit{What are the existing solutions for trajectory prediction?} This question forms the foundation of the survey of modeling approaches. It seeks to identify the techniques and methodologies developed by the research community for trajectory prediction, covering the main modeling approaches used in recent years. \\

\item \textbf{RQ2:} \textit{Is trajectory prediction problem solved or there is a room for improvement?} To answer this question, we reflect on the current state of trajectory prediction research, outlining the challenges and research gaps guiding ongoing development, as well as the limitations that hinder real-world deployment and integration. \\

\item \textbf{RQ3:} \textit{What are the emerging research trends in the trajectory forecasting realm?} This question aims to discern the current focus of researchers and which pressing challenges they are addressing. 
\end{itemize}

To address the posed research questions, this study surveys the latest advancements in prediction models and offers a structured framework for understanding the trajectory forecasting landscape. We review key factors that influence the performance and robustness of prediction models. This analysis aims to assist researchers in identifying gaps in the current literature and guiding future developments in the field. Our contributions are as follows:

\begin{itemize} 
\item Provide an extensive overview of the prediction pipeline covering input modalities, modeling features and approaches, output modalities, evaluation metrics, and prediction paradigms.

\item Extend the taxonomies presented in \citep{9756903, BHARILYA2024100733}, offering a structured classification of modeling approaches and positioning the broad set of methods found in the literature. Each category is supported by a comparative table summarizing key models and their defining characteristics for quick reference.

\item Provide a review of active research areas in the trajectory prediction domain, covering topics such as uncertainty estimation, integration of driving knowledge, planning-conditioned prediction, vision-language model (VLM)-guided prediction models, and collaboration between vehicles.

\item Provide an open repository\footnote{\url{https://github.com/AV-Lab/Trajectory_Prediction_Models_Zoo}} featuring two prediction pipelines to facilitate experimentation and benchmarking. The first pipeline includes multiple learning-based models that can be trained on SOTA or custom datasets and the second pipeline implements a detect–track–predict approach that supports several sensor configurations (e.g., frontal camera, stereo camera, LiDAR, LiDAR coupled with cameras).  

\end{itemize}

Further, we discuss the related surveys conducted for trajectory prediction task and elaborate on the scope and structure of this study.

\subsection{Related Surveys}
\label{sec:rel_work}

\begin{table*}[t]
\caption{Comparison with Existing Review Papers: RP1 - \citet{9739407}, RP2 - \citet{DBLP:conf/ijcai/TeetiKSBC22}, RP3 - \citet{math9060660}, RP4 - \citet{9158529}, RP5 - \citet{9756903}, RP6 - \citet{10100881}, RP7 - \citet{9756845}, RP8 - \citet{10.1177/0278364920917446}, RP9 - \citet{s21227543}, RP10 - \citet{9899358}, RP11 - \citet{10149114}, RP12 - \citet{GomesARO}, RP13 - \citet{BHARILYA2024100733}}
\centering
\begin{adjustbox}{max width=\textwidth}
\begin{tabular}{|p{4.5cm}|c|c|c|c|c|c|c|c|c|c|c|c|c|c|}
\hline 
\textbf{\begin{tabular}{@{}l@{}} Discussed Topics \end{tabular}} & RP1 & RP2 & RP3 & RP4 & RP5 & RP6 & RP7 & RP8  & RP9 & RP10 & RP11 &  RP12 & RP13 & Our Review  \\ \hline 

Trajectory prediction problem formulation &  \checkmark  & \checkmark  & \checkmark  & \checkmark & \checkmark &  \checkmark & \xmark &  \xmark  &  \xmark & \xmark  & \checkmark & \checkmark & \checkmark & \checkmark\\ \hline

Overview of prediction pipeline & \xmark & \checkmark  & \xmark & \xmark & \xmark  &  \checkmark & \xmark &  \xmark  &  \xmark & \xmark  & \checkmark & \xmark & \xmark & \checkmark \\ \hline

Overview of input and output modalities &  \checkmark & \checkmark  & \xmark & \checkmark & \xmark  &  \checkmark & \checkmark &  \xmark  & \checkmark & \xmark  & \checkmark & \xmark & \xmark &  \checkmark\\ \hline

Summary of prediction datasets &  \checkmark & \checkmark  & \xmark & \xmark & \checkmark  &  \checkmark & \checkmark &  \checkmark & \checkmark & \checkmark & \checkmark & \checkmark & \checkmark & \checkmark \\ \hline

Survey of physical modeling approaches & \xmark & \xmark & \xmark & \xmark & \checkmark  & \xmark & \xmark & \checkmark  &  \xmark &  \xmark & \xmark & \xmark & \checkmark & \checkmark \\ \hline

Survey of machine learning modeling approaches & \xmark & \xmark & \xmark & \xmark & \checkmark  & \xmark & \xmark &  \checkmark &  \xmark & \xmark & \xmark & \checkmark & \checkmark &  \checkmark\\ \hline

Survey of deep learning modeling approaches &  \checkmark &  \checkmark & \checkmark & \xmark & \checkmark &  \checkmark & \checkmark  & \checkmark  & \checkmark  & \checkmark & \checkmark & \checkmark & \checkmark &\checkmark \\  \hline

Survey of reinforcement learning modeling approaches  & \xmark  & \xmark & \xmark & \xmark & \checkmark  & \xmark & \xmark & \checkmark  & \xmark & \xmark & \xmark & \checkmark & \checkmark & \checkmark \\ \hline

Discussion of prediction paradigms & \xmark  & \xmark & \xmark  & \xmark  &  \xmark  & \xmark  & \xmark  & \xmark  & \xmark  & \xmark  & \xmark  &  \xmark  &  \xmark & \checkmark \\ \hline

Discussion on intention-aware approaches & \xmark  & \xmark & \xmark  & \xmark  &  \xmark  &  \checkmark  & \xmark  & \xmark  & \xmark  & \xmark  & \checkmark  &  \xmark  &  \checkmark & \checkmark \\ \hline

Discussion on interaction-aware approaches & \xmark  & \xmark & \xmark  & \xmark  &  \xmark  &  \checkmark  & \xmark  & \xmark  & \xmark  & \xmark  & \xmark  &  \xmark  &  \checkmark & \checkmark \\ \hline

Discussion of active research topics & \xmark  & \xmark & \xmark  & \xmark  &  \xmark  & \xmark  & \xmark  & \xmark  & \xmark  & \xmark  & \xmark  &  \xmark  &  \xmark & \checkmark \\ \hline

Discussion of research gaps & \xmark  & \checkmark & \xmark & \checkmark & \checkmark  &  \checkmark & \xmark & \checkmark  & \xmark & \xmark & \checkmark & \checkmark & \checkmark & \checkmark \\ \hline

\end{tabular}
\label{tab:comparison_With_surveys}
\end{adjustbox}
\end{table*}

\begin{figure*}[h!]
\centering
   \includegraphics[width=1.0\textwidth]{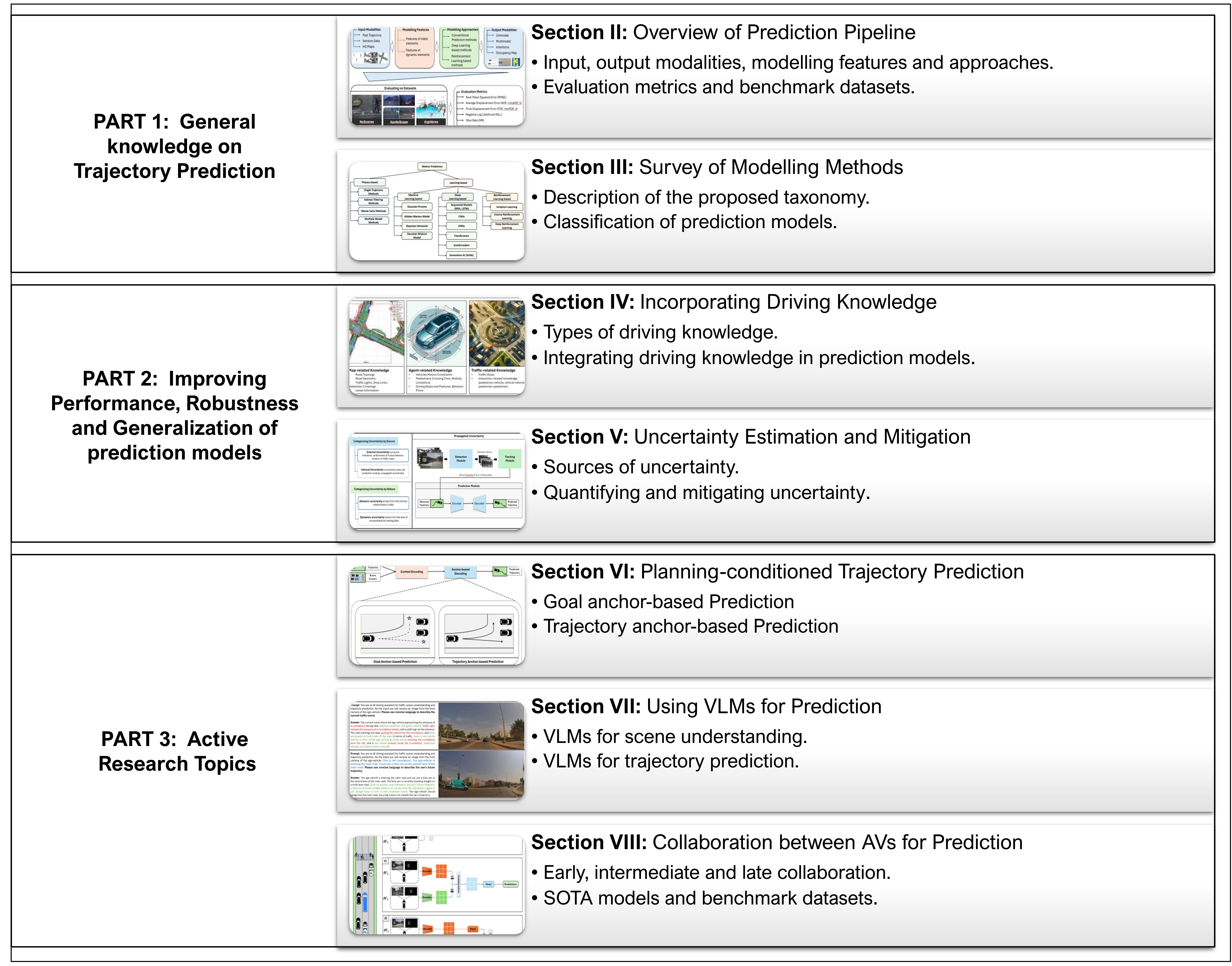}
   \caption{Structure of the review paper: the study comprises seven main sections. The first two sections - an overview of the prediction pipeline and a survey of modeling approaches, provide general knowledge of the trajectory prediction domain. Next two sections are dedicated to incorporating driving knowledge and addressing uncertainty focusing improving performance, robustness and generalization of prediction models. Last three sections cover planning-conditioned prediction, VLM-guided prediction, and collaboration between intelligent vehicles for trajectory prediction.}
   \label{fig:structure_of_paper}
\end{figure*}

This section reviews key survey papers related to the trajectory prediction domain. Each paper is briefly discussed, emphasizing its major focus points. Table \ref{tab:comparison_With_surveys} provides a comparative overview of the topics covered in these surveys, contrasting them with the focus areas of this study. This comparison delineates the scope and layout of this review.

\citet{9739407} review learning-based methods for trajectory predictions, considering aspects of representation, modeling, and learning. For representation, the authors examine agent state and scene context. In terms of modeling, the survey categorizes methods into feature encoding, interaction modeling, and generative models. \citet{DBLP:conf/ijcai/TeetiKSBC22} focus their survey on vision-based intention and trajectory prediction approaches. All methods are summarized while detailing their architectures for input encoding, temporal and social encoding, decoding and loss function. The survey also discusses goal-conditioned methods and prediction in stochastic environments. As prospective research directions, it outlines real-time inference and the shortage of self-supervised solutions. \citet{math9060660} conducted a survey on tracking and motion prediction methods. The authors argue that for effective trajectory prediction, a system should rely on physical and traffic rules constraints to narrow down the set of possible trajectories and select a reasonable subset. To classify prediction methods, it categorizes the existing approaches into neural networks-based, methods using stochastic techniques, and mixed methods that combine stochastic techniques with neural networks.

Next, \citet{9158529} review deep learning-based prediction methods, categorizing based on input representation, prediction method, and output type. For input representation, the paper identifies four classes: the past trajectory of the target vehicle, the past trajectories of the surrounding vehicles, a simplified bird's-eye view (BEV) image with information about the road geometry, and raw sensor data. 
As research gaps, the paper highlights running time evaluation of prediction algorithms, assumption of full observability of BEV images, lack of integration of traffic rules, and generalizability of the prediction models across diverse scenarios. \citet{9756903} base their classification on the prediction method, modeling of contextual factors, and generated output. For contextual factors, the paper distinguishes physics-related, road-related, and interaction-related factors. For prediction methods, it categorizes into physics-based, classic machine learning, deep learning, and reinforcement learning. The suggested potential research directions are inclusion of more interaction-related information and improving model robustness towards uncertainties and perturbations caused by tracking, location, and map errors. \citet{10100881} review deep learning (DL) based models with a focus on the inclusion of driving knowledge. The study reviews how SOTA models incorporate the driving knowledge to enhance performance, explainability, and interpretability. \citet{9756845} focus their survey on motion prediction and state estimation, through thoroughly analyzing and discussing sensory perception capabilities. The authors divide the discussion of perception capabilities into two categories: ego-vehicle perception and cooperative perception. 

The next line of work focuses on pedestrian trajectory prediction. This task introduces an additional layer of complexity as pedestrian motion is less constrained compared to that of vehicles and is influenced by a broader range of factors. These include an individual's own goal intent, actions of surrounding agents, the topology of the environment, social relations between agents, and social rules and norms. Some of these impacting factors are not directly observable and must be inferred or modeled from contextual information. \citet{10.1177/0278364920917446} conducted a survey on pedestrian trajectory prediction and proposed to categorize prediction methods based on the motion modeling approaches and contextual cues. The authors divided modeling approaches into ``sense-predict" (physics-based approaches), ``sense-learn-predict" (pattern-based approaches) and ``sense-reason-predict" (planning-based approaches). The contextual cues are categorized into cues of the static environment and dynamic agents. \citet{s21227543} concentrate their review on the types of sensors used for acquiring data for pedestrian trajectory prediction and the deep learning-based approaches employed. The authors offer an overview of LiDAR, RADAR, and camera sensors, discussing their comparative features and the associated processing methodologies to transform acquired data into predicted trajectories. In terms of prediction approaches, the paper distinguishes CNN-based solutions, sequential prediction algorithms (based on RNN or LSTM), and generative-based approaches. Next, the survey conducted by \citet{9899358} compares knowledge-based approaches, used to simulate crowd dynamics, with deep learning-based methods. They explore the relevance of using knowledge-based methods for prediction, given the high accuracy of data-driven algorithms and their ability to learn collective dynamics. \citet{10149114} conducted a review of deep learning-based approaches for predicting pedestrian trajectory and intention in urban scenarios. They categorize existing works by prediction tasks (trajectory, intention, and joint prediction), types of pedestrian behavior features considered by prediction models, and the network structures of these models (sequential and non-sequential). As the research gaps, the authors mention integration of appearance and skeleton behavioral features into prediction models and the lack of models predicting from raw sensor data without assumptions of perfect detector and tracker. Lastly, \citet{GomesARO} conducted a survey focused on intention and interaction-aware solutions. Within the realm of deep learning solutions, the survey highlights methods that utilize attention mechanisms, transformers-based solutions, and continual learning approaches. 

The current survey aims to fill the gap by extensively analyzing recent research on trajectory prediction solutions. Given its comprehensive scope, structured approach, and focus on emerging topics, we believe this survey is timely and will serve a broad audience in the field. 

\subsection{Scope and Structure}
The scope of this review is defined by the objective to offer a comprehensive overview of the trajectory prediction field for novice researchers and provide discussion of active topics for already experienced practitioners. Figure \ref{fig:structure_of_paper} illustrates the structure of our review, detailing the sections and main points discussed.

The first part consists of two sections aimed at providing general knowledge of the trajectory prediction realm. Section \ref{sec:overview_of_pipline} provides an overview of a trajectory prediction pipeline and covers problem formulation and prediction paradigms. Section \ref{sec:survey} lays out the taxonomy and categorization of the modeling approaches and elaborates on the models available in the literature from each category. The next part is dedicated to the discussion on improving models' performance, robustness and generalization. This part also comprises two sections. Section \ref{sec:incorporating_knowledge} reviews the inclusion of driving knowledge in prediction models, covering different types of driving knowledge and the ways they are incorporated into prediction models. Section \ref{sec:uncertainty} is dedicated to uncertainty estimation and mitigation in prediction models. It elaborates on the sources of uncertainty and approaches to mitigate the risks associated with them. The last part depicts active research areas and is presented by three sections. Section \ref{sec:planning-driven} discusses planning-prediction-coupled frameworks, where for generating predictions the ego-vehicle models internal planning intentions of surrounding agents. Next, Section \ref{sec:vlm} discusses the application of Vision-Language models in scene understanding and trajectory prediction. Further, Section \ref{sec:collaboration} discusses the concept of collaborative perception, elaborating on different types of fusion and different types of challenges caused by collaboration. Lastly, Section \ref{sec:discussion} provides a discussion on the posed research questions, outlines the research gaps, and highlights future research directions.

For a comprehensive survey of trajectory prediction methods, we recommend Section \ref{sec:survey}. Practitioners may find Section \ref{sec:uncertainty} particularly valuable, where sources of uncertainty are discussed. Researchers interested in specific topics within the domain of trajectory prediction should refer to Sections \ref{sec:planning-driven}, \ref{sec:vlm} and \ref{sec:collaboration}.

\section{Overview of Prediction Pipeline}
\label{sec:overview_of_pipline}

\begin{figure*}[h!]
\centering
   \includegraphics[width=1\textwidth]{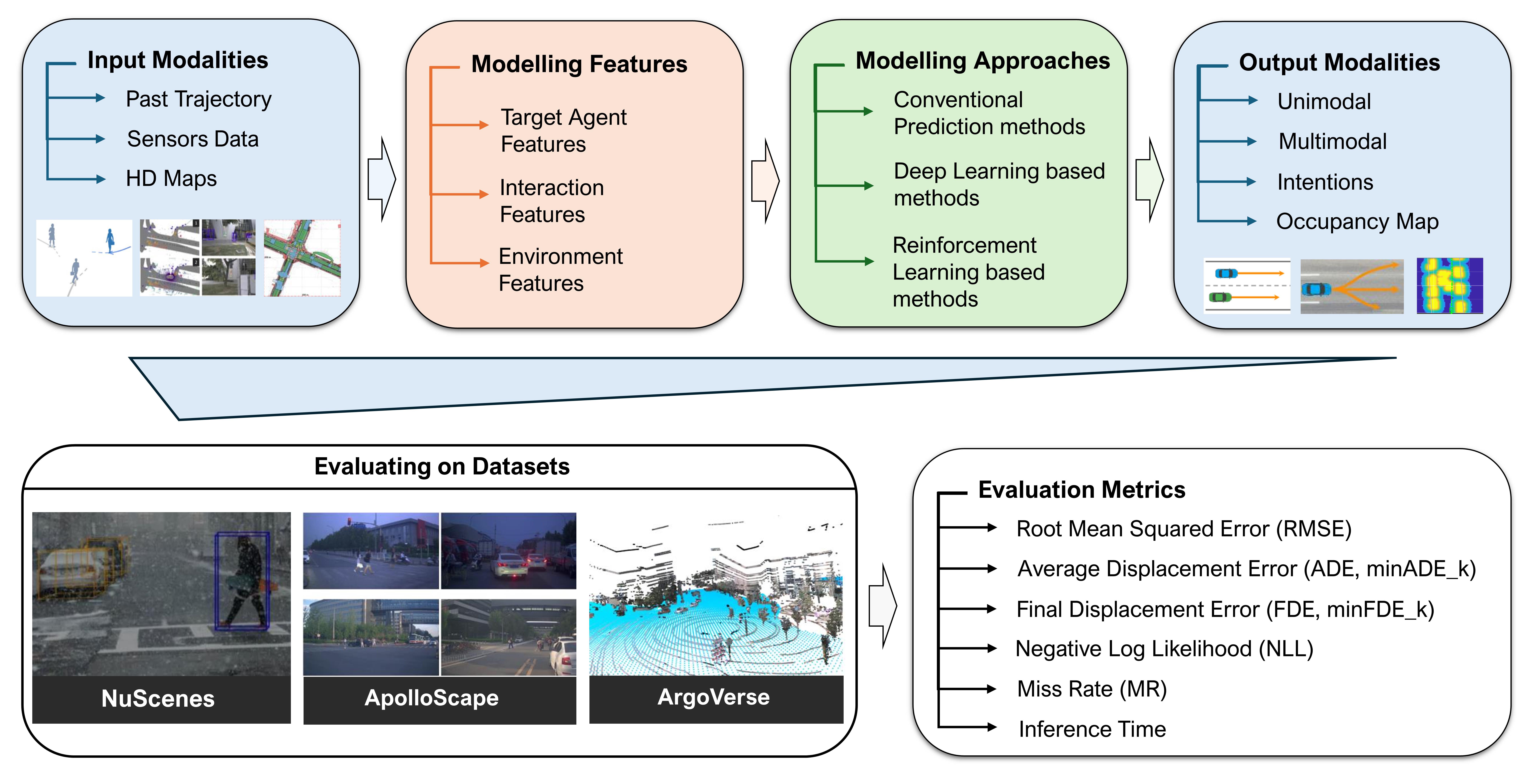}
   \caption{Overview of trajectory prediction pipeline: including input modalities (past trajectories, sensor data, and HD maps), key modeling features (target agent, interaction, and environment features), and modeling approaches (conventional, deep learning-based, and reinforcement learning-based methods). The outputs of forecasting models can be unimodal and multimodal predictions, in the form of intentions, and occupancy maps.}
   \label{fig:overview_pipeline}
\end{figure*}

This section aims to provide a comprehensive overview of a prediction pipeline. Figure \ref{fig:overview_pipeline} illustrates the general prediction pipeline flow, highlighting its main components. The overview begins with the problem formulation of trajectory prediction task, elaborates on input modalities, provides a discussion of modelling features and approaches, and iterates through possible output modalities. The section continues by elaborating on metrics used to evaluate trajectory forecasting models and listing benchmark datasets with leader-board solutions. It concludes with a discussion of existing in the literature trajectory prediction paradigms.

\subsection{Problem Formulation}

A traffic scenario is characterized by an ego vehicle $ev$ and a set of agents $A = \{a_1, a_2, ..., a_n\}$. Each agent $a_i$ is defined by a state vector $\tau_i$ $\in$ $\mathbb{R}^{t \: \times \: m}$, where $t$ indicates the length of observation and $m$ is an arbitrary dimension and in the most reduced form $(m=2)$.
The state vector $(\tau) \in \mathbb{R}^{t \: \times \: 2}$ for each agent is defined by its position $(x,y)$ over time:

\begin{equation*}
    \tau_i = [(x_i^1, y_i^1),  (x_i^2, y_i^2), ... , (x^{t-1}_i, y^{t-1}_i), (x_i^t, y_i^t)], \:\:\:\: t \in \mathbb{N}
\end{equation*}

Depending on the type of sensors observations, the state vector ($\tau$) can additionally include orientation and speed of the agent. The prediction task is to estimate trajectory sequences $\zeta_i^{t+1: t + \Delta t}$ for each traffic agent $a_i$ and specified time horizon $\Delta t$ given the previous states:

\begin{equation*}
\zeta_i^{t+1: t + \Delta t} = \{(x^{t+1}_i, y^{t+1}_i), (x^{t+2}_i,y^{t+2}_i)  ..., (x^{t+\Delta t}_i,y^{t+\Delta t}_i)  | \tau_i\}
\end{equation*}

in which $t$ is the current time frame and $(x_i^k,y_i^k)$ is the prediction state of the \textit{i-th} agent. The goal of trajectory predictor ($\mathcal{P}$) is to estimate trajectory sequences $\zeta_i^{t+1: t + \Delta t}$ for $i \in \{1, 2, ..., n-1, n\}$ given the historical state of the traffic agents (T): 
\begin{equation*}
    \mathcal{P}(T) = \{ \zeta_1, \zeta_2, ..., \zeta_{n-1}, \zeta_n \}, \:\:\:\: n \in \mathbb{N}
\end{equation*}

In the given formulation, the predictor ($\mathcal{P}$) is a function of a single parameter, past trajectories ($\mathcal{T}$). This can be expanded to make $\mathcal{P}$ a function of multiple parameters if additional inputs are available, e.g., road geometrical features ($G$) or interaction modeling features ($I$). Additionally, the coordinate component of $\tau$ can be represented in either a local or global coordinate system. Sensor observations are inherently centered on the ego-vehicle, meaning all raw measurements are initially recorded in the vehicle's local frame. As such, predicting trajectories in the local frame does not require coordinate transformations. However, local frames introduce ego-motion distortions like stationary vehicles may appear to move, and turning maneuvers can be misrepresented due to the ego-vehicle's own motion. To ensure consistent learning of motion patterns, ego-motion must be compensated, e.g., through frame-to-frame undistortion using odometry interpolation or local frame stabilization using ego-motion data. Alternatively, trajectories can be transformed to a global frame using ego-pose information, as done in \citep{nuscenes2019}, though this requires precise localization.

Representing trajectories in Cartesian coordinates ties them to the underlying road geometry, making it more challenging for prediction models to generalize across scenes where structurally different layouts exhibit similar behaviors. The Frenet coordinate frame addresses this by encoding each trajectory point relative to the ego-vehicle’s path as \((s, d)\), where \(s\) denotes the arc length along the lane centerline and \(d\) the perpendicular lateral offset. This decouples motion patterns from the global layout and enables learning behaviors relative to the road structure~\citep{Ye_2023}.

\subsection{Input Modalities}
\label{sec:input_modalities}

In this subsection, we discuss the classification of input modalities and review relevant studies in the literature for each category. Input modalities are classified into three groups: past trajectories of agents, sensor data from LiDAR, radar, or cameras, and high-definition (HD) maps describing static environmental context. Explicitly encoding tracked past trajectories is one way to capture temporal dynamics; alternatively, temporal information can be inferred from sequences of raw sensor inputs. Regardless of the format, providing input from which dynamics can be extracted is a fundamental cornerstone of trajectory prediction models.

The combination of input modalities, e.g., coupling past trajectories with HD maps for road layout information and camera feeds for traffic light detection, is shaped by the system’s sensing capabilities, operational constraints, and design choices. Models trained solely on past trajectories may require complementary context under noisy conditions, which can be addressed by processing additional sensor data to extract semantic and environmental cues. While HD maps enrich predictions with high-level semantics, they also introduce dependencies on precise localization and map availability. Each modality has inherent limitations, and combining them increases processing complexity. In the following subsections, we discuss each input modality in detail, highlighting its strengths and limitations.

\subsubsection{Past Trajectories}

Predicting future trajectories solely based on past trajectories \citep{ 10.1109/IVS.2017.7995919, 8290702, Zyner2018NaturalisticDI, Park2018SequencetoSequencePO} forces the models to learn patterns by relying purely on spatio-temporal dependencies. Predicting future trajectories solely based on past motion is efficient in terms of prediction speed and computational cost. One way to incorporate additional cues is to implicitly encode interaction of the target agent. This can be done by including the past trajectories of its surrounding agents in the prediction process, as shown in the studies \citep{Phillips2017GeneralizableIP, 8672889, Ding2019OnlineVT, Diehl2019GraphNN, Hu2018ProbabilisticPO, Ding2019PredictingVB, 8917228}. This enables predictions that better reflect real-world behaviors, such as reactions to nearby vehicles or pedestrians. However, when forecasting models are trained on a single input source under the assumption of error-free observations, their performance will degrade in real-world deployments. In practice, the past trajectories available to prediction models are often noisy and incomplete. Therefore, models should be evaluated under such conditions (e.g., by simulating tracker errors) to quantify the magnitude of prediction error, assess sensitivity, and evaluate the model’s robustness to inaccurate inputs. On representations, typically the agent state in past trajectories is represented as $(x,y)$ coordinates and models are trained on relative distances. Additionally, heading and speed can be added complementarily. Studies \citep{9043898, ma2017, Carrasco2021SCOUTSA} improved prediction accuracy by including velocity, acceleration, and orientation as input features. 

\subsubsection{Sensors Data}
Sensor data from LiDAR, radar, or cameras is used as prime or additional input to prediction models. This section first examines each sensor individually and then discusses approaches of integrating multi-modal inputs.

\begin{figure*}[t!]
\centering
   \includegraphics[width=1\textwidth]{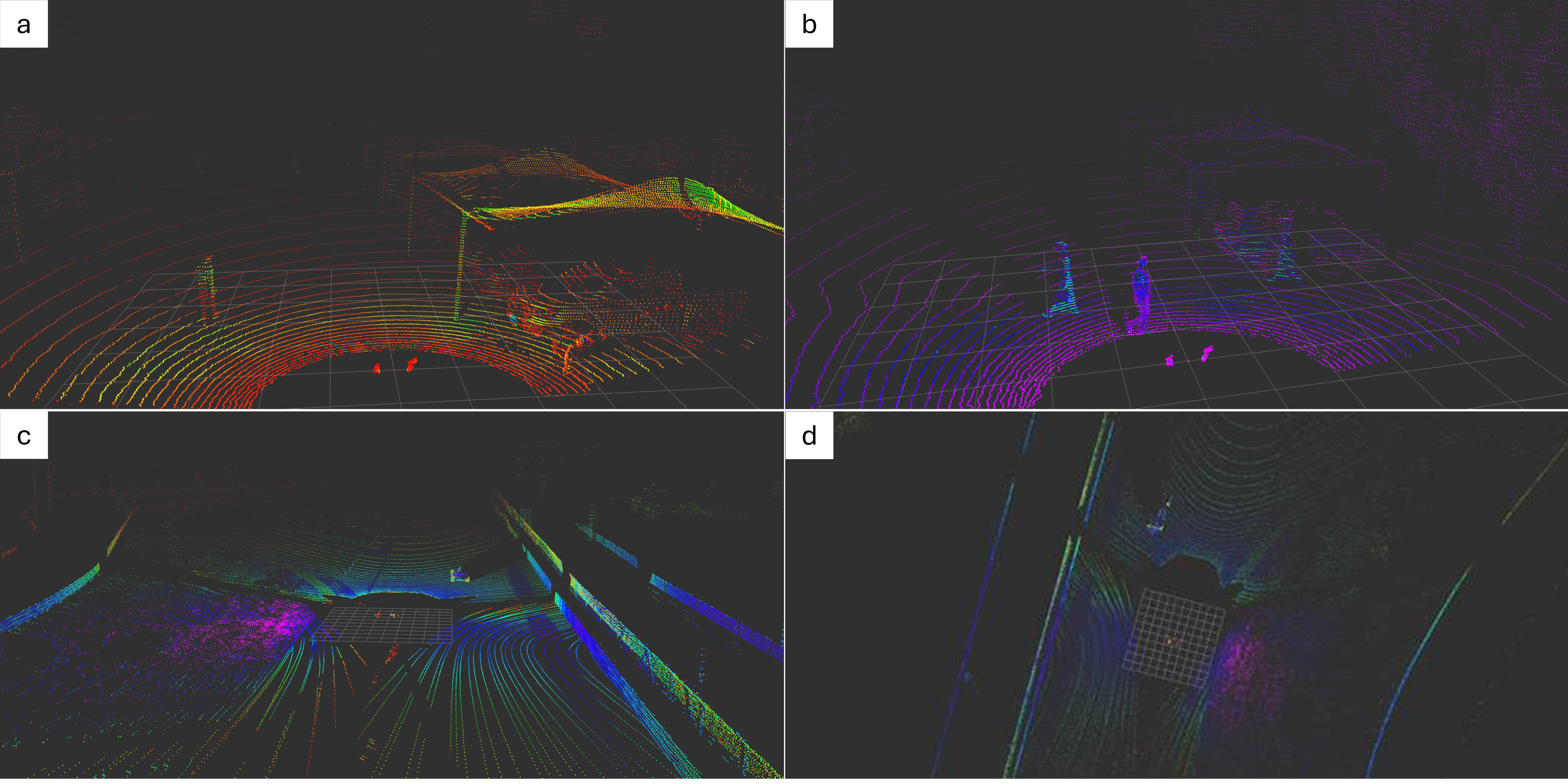}
   \caption{Point clouds captured using a 64-layered Ouster LiDAR and 64-layered Kinetic LiDAR: (a) in a parking area, (b) on a campus, (c) during an Autonomous Racing Competition (A2RL) with the LiDAR mounted on a Formula 1 car, and (d) at A2RL in a top-down view.}
   \label{fig:lidar_fig}
\end{figure*}

\textbf{\textit{LiDAR.}}
LiDAR sensors enable environmental perception through laser-based distance measurements, making them indispensable for autonomous driving systems. Two types of LiDAR are commonly employed: perimeter LiDAR and $360^\circ$ LiDAR. Perimeter LiDAR, mounted on the vehicle's edges or corners, is optimized for monitoring close-range objects, blind spots, and collision risks, particularly in scenarios like parking or navigating tight spaces. In contrast, $360^\circ$ LiDAR, typically roof-mounted, provides a complete $360^\circ$ field of view, supporting long-range detection and situational awareness. Current SOTA datasets predominantly include data from $360^\circ$ LiDARs (refer to Figure \ref{fig:lidar_fig} for examples), while perimeter LiDARs are used by automotive manufacturers as complementary sensors to enhance overall perception. One of LiDAR's key strengths is its ability to function reliably in all lighting conditions, including complete darkness and bright sunlight, due to its light-invariant nature. However, its performance degrades in adverse weather, such as heavy rain, or snow, where laser beam scattering reduces detection range and accuracy along with highly reflective surfaces such as glass, potentially creating blind spots. Despite these challenges, LiDAR remains highly effective for detecting obstacles \citep{DBLP:journals/corr/abs-2103-05423}, vehicles \citep{QIAN2022108796, Mao2023, e25040635}, lane markers \citep{8953739,8205955, WANG201771, Wang2018}, and forecasting \citep{9880029}.

\citet{8578474} proposed a deep-learning model for 3D detection, tracking, and motion prediction using LiDAR data as input. The model processes point clouds with 3D convolutions to extract spatial and temporal information, completing all tasks within 30 milliseconds on a custom dataset recorded with LiDAR-equipped vehicles. Further,  \citet{7795975} developed a LiDAR-based model for classifying pedestrian intentions at urban crosswalks. This approach projects LiDAR point clouds into a 2D representation based on angular coordinates. The resulting 2D images are passed to CNNs to extract spatial and motion-related features, which are further combined with pedestrian velocity and distance to the crosswalk for intention classification.

Two limitations hinder the widespread adoption of LiDAR technology: its relatively high cost and its accuracy in detecting pedestrians. Although LiDAR technology is advancing and prices are gradually declining, it remains expensive and impractical for consumer-level mass-market adoption. The second limitation is its lower 
accuracy in recognizing  fine-grained classes such as pedestrians and cyclists. This disparity stems from its sparse, geometry-centric output and lack of rich semantic information, which makes interpretation of complex objects challenging. For example, on the KITTI leaderboard, the top pedestrian detection score is only 56\% \citep{casa2022}, compared to a 93\% vehicle detection score achieved by \citep{10205191}. As a result, LiDAR systems are ideally complemented by cameras, which provide the necessary semantic context to enhance pedestrian detection performance.

\textbf{\textit{Camera.}} 
Cameras play a key role in perception of autonomous vehicles in detecting traffic participants, road signs, and traffic lights. These sensors are characterized by their resolution, field of view, and configuration. The vision community is highly advanced, providing efficient algorithms for object detection \citep{9004469}, semantic segmentation \citep{Neumann_2021_CVPR}, and lane recognition \citep{osti_10304295}. Camera setups in AV systems are generally categorized as single frontal cameras \citep{technologies10040090} or multi-view configurations \citep{YAN2021106}. Single frontal camera setups primarily rely on 2D perception and are mostly used in driving assistance systems. In contrast, multi-view setups \citep{Loukkal_2021_WACV} arrange cameras to ensure complementary, non-overlapping fields of view, providing 360-degree observability. Additionally, camera systems employ either monocular or stereo setups. Monocular cameras \citep{9004469, Neumann_2021_CVPR, osti_10304295, Loukkal_2021_WACV} are lightweight and cost-effective but restricted to 2D image data, relying on advanced algorithms for depth estimation. Stereo cameras utilize dual lenses to calculate depth through stereo triangulation, enabling 3D perception \citep{8972435}. 

\citet{Bhattacharyya2017LongTermOP} proposed a prediction model that uses data from an onboard monocular camera. Pedestrian bounding boxes detected in the camera feed are combined with the ego-vehicle's odometry information and processed by a two-stream Bayesian RNN encoder-decoder architecture. The encoded bounding boxes and odometry data are used to predict pedestrian trajectories over a one-second horizon. Similarly, \citet{8972435} designed a system utilizing a stereo camera setup to estimate 3D human poses and predict pedestrian trajectories. A twin network processes the video streams from each camera independently to estimate 2D poses. These 2D poses are then aligned and fused using a stereo consistency constraint to produce accurate 3D pose estimates, which are subsequently processed by an adapted SocialGAN framework to generate 3D pedestrian trajectories.

RGB cameras, however, degrade in performance under low-light or adverse weather conditions. To address these limitations, thermal and infrared cameras can serve as complementary sensors. Thermal cameras\footnote{\url{https://www.flir.com/oem/adas/adas-dataset-form/}}, which detect heat signatures and can be utilized in detecting pedestrians and animals in complete darkness or adverse weather conditions. However, their shorter range and lack of fine detail limit their utility for tasks like detecting road signs and require specialized deep learning solutions. Infrared cameras enhance visibility in low-light conditions and are suitable for detecting road signs, detecting and tracking pedestrians and vehicles \citep{madjid2023multi}. Nonetheless, they are sensitive to bright light sources, which can cause glare and blooming artifacts. This issue can be mitigated with filters designed to reduce glare and enhance image quality in challenging scenarios. Despite their individual limitations, combining thermal or infrared cameras with RGB cameras creates a more robust sensor suite, ensuring reliable operation across diverse environments.

\textbf{\textit{Radar.}} 
Radar sensors are deployed to estimate the relative velocity, distance, and direction of targets by analyzing reflected electromagnetic waves \citep{Meinel2014EvolvingAR}. Modern automotive radar systems provide coverage with angles up to 150° and distances of up to 250 meters \citep{s150614661}, making them suitable for detecting objects in diverse driving scenarios. Frequency-Modulated Continuous Wave (FMCW) radar is a key technology in these systems, enabling precise measurements. By comparing transmitted and reflected signals, the distance of a target is calculated from the beat frequency, while the relative velocity is determined through Doppler shift analysis. For direction, FMCW radar pairs with digital beam-forming \citep{Hasch2015DrivingT2} or multi-antenna setups to estimate the angle of arrival of reflected waves.

\citet{Kim2018MovingTC} presented a model to classify agent types (pedestrian, cyclist, or vehicle) using radar data. Radar signals are transformed into range-velocity (RV) images that capture an object’s distance and motion. These RV images are processed by a convolutional LSTM to learn spatial and temporal features, followed by a fully connected layer to map these features to target classes. \citet{Steinhauser2019MicroDopplerEO} proposed a system focused on individual limb movements for pedestrian behavior prediction. Their method employs a deconvolution algorithm combined with a clustering technique to extract and analyze reflections from pedestrian body parts, decomposing the range-Doppler spectrum to identify the torso, arms, and legs. Radar typically complements LiDAR and camera inputs in SOTA datasets, providing velocity and heading cues.

\textbf{\textit{Rasterization.}}
Rasterization converts raw sensor data into structured, grid-like representations. For example, LiDAR point clouds can be transformed into grid-based formats, such as intensity or height maps, that capture spatial geometry. Radar data adds dynamic attributes such as velocity and direction \citep{Major2019VehicleDW}, which can be encoded either through color coding (e.g., higher speeds represented by brighter colors) \citep{9699098} or as separate velocity channels, storing velocity values in an additional data layer for each grid cell. Similarly, camera images can be rasterized to map rich semantic information, such as object classes and bounding boxes. This unified rasterized format facilitates the fusion of multimodal sensor data.

The most common type of rasterization for autonomous driving is Bird’s Eye View (BEV) rasterization \citep{Lee2017ConvolutionNN, 10.1109/ICRA.2019.8793868, Deo_2018_CVPR_Workshops, DBLP:conf/cvpr/ZhaoXMCBZ0W19, Lee2017DESIREDF, li2023powerbevpowerfullightweightframework}, where sensor data is transformed into a top-down perspective of the environment. In BEV rasterization, dynamic parameters such as positions and velocities are mapped into a unified plane aligned with the ego vehicle. This format often employs color-coded semantics to differentiate objects and incorporates temporal information through stacked frames or brightness variations to represent motion over time. Simplified BEV representations extend this concept by depicting static and dynamic elements, e.g., road lanes, vehicles, and pedestrians, using polygons and lines.

\subsubsection{HD Maps}

\begin{figure*}[t!]
\centering
   \includegraphics[width=1\textwidth]{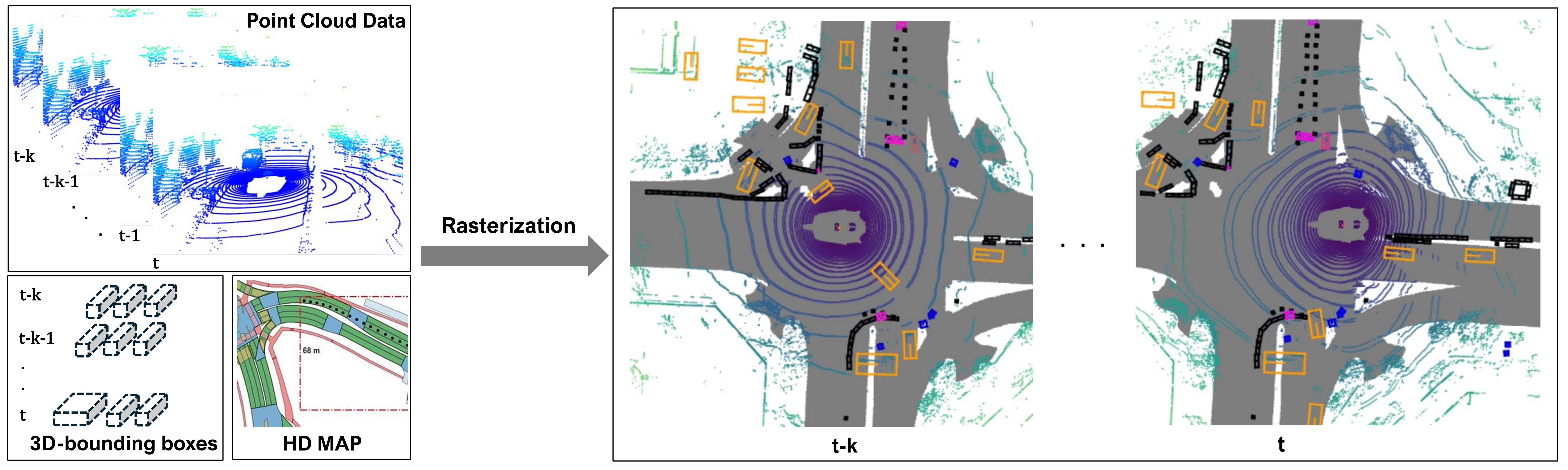}
   \caption{BEV representation aligning dynamic and static elements: raw point cloud data, HD map, and ground truth 3D bounding boxes.}
   \label{fig:hd_map_rasterization}
\end{figure*}

Latest datasets supplement collected data with High Definition (HD) maps, which provide detailed representations of static road features, such as lane markers, lane edges, pedestrian crossings, road layouts, and stop signs. Typically, the map is presented as a single large file and requires pre-processing to extract locally relevant environmental information. A Region of Interest (ROI), usually encompassing 50–100 meters around the vehicle's position, is extracted to capture nearby lanes, crosswalks, traffic signs, and other relevant features. Some SOTA datasets, such as NuScenes \citep{nuscenes2019} and Argoverse \citep{Argoverse, Argoverse2}, provide SDKs to query map data and efficiently extract ROIs. This ROI is then transformed into a BEV image, where semantic elements like drivable areas and lane boundaries are encoded using distinct colors or layers. Prediction models can process this BEV image independently, combining it with past trajectories for improved trajectory forecasting \citep{9157523, 10.1007/978-3-031-43424-2_24, Lee2017DESIREDF}, or use it in conjunction with sensor data to create a combined rasterized representation \citep{Djuric2020MultiXNetMM, 10.1007/978-3-030-58523-5_40} (refer to Figure \ref{fig:hd_map_rasterization} for illustration).

Incorporating information about the driving environment from HD maps and geometrically constraining future trajectories can boost forecasting accuracy in structured environments such as urban roads, intersections, and highways. Regarding large empirical drops in displacement error with map information: VectorNet \citep{Gao2020VectorNetEH} achieves a 28\% reduction at a 3-second horizon by incorporating lane polylines, and LaneGCN \citep{liang2020learning} reduces 6-second error from 1.90 to 1.35 meters by encoding lane centerlines. More moderate gains are reported by the model QCNet \citep{zhou2023query}, where the addition of a map-based attention mechanism contributed to a reduction in displacement error from 0.82 to 0.78 meters. The effectiveness of integrating map data depends on how it is represented --- whether as vectorized lane graphs, local semantic priors, or attention-weighted sequences, and how it is integrated with agent dynamics.

As for limitations, this approach requires precise localization of the ego-vehicle and is sensitive to localization errors. Additionally, constructing HD maps involves substantial costs and ongoing efforts to update them. To alleviate the human labor associated with constructing HD maps, the research community is actively exploring methods for automatically constructing them from raw sensor data \citep{10.5555/3618408.3619338, Li2021HDMapNetAL, philion2020lift, 9156806, 10656243}. Combining HD maps with sensor data offers a promising solution to mitigate localization errors and handle changes in the road environment, such as road closures or modifications to traffic signs, however, it requires additional computational pre-processing, which impacts the inference time of the prediction pipeline.

\subsection{Modelling Features}

Modeling features in trajectory prediction can be broadly categorized into static and dynamic components. Static features refer to the physical layout of the environment, such as road geometry, sidewalks, or walkable areas. Models that include this information can incorporate spatial constraints, e.g., road boundaries, lanes, or walkable regions, into the forecasting process, allowing predicted trajectories to better align with the underlying scene layout. Dynamic features include features related to the target agents and their interactions. Agent’s past motion is a dynamic feature, where the number of past steps is a design choice and varies across models. For example, physics-based models typically rely on the last one or two steps, while deep learning models with attention mechanisms can learn temporal dependencies from longer sequences. Interaction modeling can be beneficial in the scenarios of dense traffic and intersections, where agent behaviors are tightly coupled. In the following sections, we elaborate on each category and describe how different models employ these features.

\subsubsection{Environment Features}
\label{subsec:env}

Traffic agents rely on the environment's topology to plan their potential routes, and this geometric information can be integrated to enhance prediction accuracy \citep{9577563, 7995734, Ballan2016KnowledgeTF, 8813889, 8460203}. The environmental features can be integrated through explicit modelling. \citet{ma2017} incorporated the distance to the boundaries of crossroads into their prediction model, while \citet{yang2019} included the presence of traffic lights, stop signs, and zebra crossings. Similarly, \citet{8813889} incorporated static features for cyclist path prediction by estimating the distance of cyclists to the nearest intersection. This estimation was passed as a contextual feature to a GRU-based RNN, enabling the model to account for the impact of intersections on trajectory predictions. \citet{8460203} presented a more advanced approach by constructing a topological map that organizes semantic features, such as roads, sidewalks, and obstacles, into an occupancy grid. A recurrent Mixture Density Network (rMDN) predicts potential pedestrian destinations, which are integrated into the topological map to encode spatial relationships and constraints.

\subsubsection{Target Agents}

As discussed in Section \ref{sec:input_modalities}, it is common for prediction methods to generate forecasts based on the past trajectories of target agents \citep{Bock2017SelflearningTP, 8317941, nikhil2018convolutionalneuralnetworktrajectory, giuliari2020transformernetworkstrajectoryforecasting}, and/or incorporate velocity history \citep{s20061776, 7795975, 9197257}, as well as heading and orientation \citep{Blaiotta2019LearningGS, Kooij2018ContextBasedPP}. In heterogeneous models, including the category of the object (e.g., person, truck, cyclist) enables prediction models \citep{8317913, COSCIA201881} to learn category-specific dynamics, although these methods may be prone to errors caused by the incorrect target agent classification by the detector. For vehicles, learning individual behavioral patterns historical data \citep{10411104} can improve prediction accuracy by capturing unique driving styles. Similarly, in pedestrian trajectory prediction, appearance cues and posture can offer insights into pedestrians' intentions. Studies \citep{Lorenzo2020RNNbasedPC, 8265243, Liu2020SpatiotemporalRR} have focused on extracting visual cues to capture behavioral patterns to improve prediction accuracy. To estimate posture, two main approaches are employed: using skeleton key-points \citep{8972435, 9423518, fang2018pedestriangoingcrossanswering, Piccoli2020FuSSINetFO} or leveraging pretrained CNNs \citep{9008118, Kotseruba2020DoTW, 9423518}.

Incorporating additional semantic attributes requires either integrating a subnetwork within the model to extract these attributes from the input modalities or utilizing networks pre-trained for these tasks, e.g., estimating head direction and gestures in pedestrian behavior prediction. In the first case, the availability of annotated data containing such attributes is often limited, and errors in predicting these attributes can propagate. In the second case, the trade-off between the accuracy boost from including these attributes and the added time complexity to the inference pipeline needs to be evaluated while model testing. 

\subsubsection{Interaction Features}
The integration of autonomous vehicles into human-driven traffic requires AV to understand socially compatible communication among traffic agents and the ability to interpret these interactions. Accounting for interaction-related factors \citep{Ma2016ForecastingID, 7838128, 7313129, Kooij2014ContextBasedPP} is beneficial in dense, interaction-intensive environments, where avoiding collisions depends on accurately modelling the interplay between traffic agents. A simple approach to integrating interactions involves using hand-crafted features, such as relative position and speed of neighboring agents, the distance between vehicles and pedestrians, and time to collision \citep{ma2017, volz2016, yang2019, 9423436}. While straightforward, this method often fails to capture the complexity and variability of interaction patterns in real-world scenarios. More sophisticated techniques for modelling interactions include social field models, game theory and graph-based approaches and computation cognition methods \citep{Liao2024ACT, Cuzzolin_Morelli_Cîrstea_Sahakian_2020, 9646493}. 

Further, we briefly elaborate on Social Pooling methods, which are widely adopted learning-based method in the literature. GNN-based modelling of interactions between traffic agents is discussed in Section \ref{sec:dl}. For a comprehensive review of other approaches and a detailed study on interaction modelling for autonomous driving, readers are referred to the extensive review available on this topic \citep{9964137}.

\textit{Social pooling} is one of the most common learning-based approaches for modelling interactions between agents. It involves extracting relevant features from neighboring agents and aggregating them into a pooled representation, typically using max or average pooling. Social-LSTM \citep{7780479} is a pioneering model in this type of interaction modelling, introducing a social pooling layer that integrates the hidden states of individual LSTMs (for each pedestrian in a scene) to account for social interactions. Subsequent studies have refined this concept by incorporating attention mechanisms to model interactions more effectively \citep{Liu2020CoLGANPA, 10.1007/978-3-030-29911-8_34}. \citet{8578651} introduced a weighted spatial affinity function, where weights are computed based on spatial features to represent interaction strength. Similarly, SoPhie model \citep{8953374} prioritized attention to closer objects, reflecting the idea that agents are more influenced by their immediate neighbors. These advancements highlight the evolution of social pooling methods, moving from uniform aggregation to more context-aware interaction modelling. \citet{Amirian2019SocialWL} included attention-based social pooling block, which encodes geometric features such distance, bearing angle into interaction features via a fully connected layer. The attention weights are computed through scalar products with LSTM-encoded hidden states.

On more advanced pooling mechanism architectures,  \citet{Gupta2018SocialGS} use relative positions of agents and process them through a Multi-Layer Perceptron (MLP) followed by max pooling to create a compact and globally aware representation, capturing both local and distant interactions. Convolutional Social Pooling (CSP), introduced by \citet{Deo_2018_CVPR_Workshops}, leverages spatial grid representations and convolutional operations to model interactions. Each vehicle is represented by an LSTM, and the interaction and spatial features are captured through a CNN that processes a spatial grid relative to the target vehicle. This grid encodes the states of surrounding vehicles' LSTMs, enabling the pooling block to hierarchically extract global and local interactions. The CNN output, combined with the target vehicle's LSTM encoding, forms the input to a decoder that predicts both future trajectories and maneuvers.

The performance of interaction-aware models is heavily influenced by how interaction learning is integrated, as well as by the diversity of scenarios in the dataset (only a subset of which typically involves densely interactive settings). Quantitatively, AgentFormer \citep{9710708}, a transformer-based model that jointly encodes social and temporal cues, demonstrates through ablations that modeling either cue in isolation leads to degraded performance. Qualitatively, models like Social-LSTM \citep{7780479} and Social-GAN \citep{Gupta2018SocialGS} have shown improvements in generating more socially acceptable trajectories in crowded environments. Regarding the efficiency of social pooling, models such as Social-STGCNN \citep{9156583} and Social-BiGAT \citep{e294141389194a54a05536938fcdd509} by replacing pooling operations with graph convolutions or attention mechanisms, enhanced predictive accuracy and reduced inference latency.

\subsection{Modelling Approach}

The choice of a modelling approach depends on the application requirements, platform characteristics, and environmental complexity, e.g., crowded urban scenarios or off-road autonomy \citep{10610128, Hassan2024PathFormerAT}. Traditional \textit{model-based methods}, such as the Constant-Velocity model or Interactive Multiple Model, explicitly encode motion constraints to generate physically plausible trajectories. These methods are simple in terms of implementation and interpretation. A more detailed discussion is provided in Section~\ref{sec:physics-based}. To capture complex trajectory patterns over longer horizons, the community often adopts data-driven approaches capable of capturing both spatial and temporal contexts. One such category is \textit{sampling-based methods} \citep{8953435, 10203123, 10203823, Ma2020LikelihoodBasedDS, Gilles2021THOMASTH, 9879462, 9812254, Anderson2019StochasticSS}, which first model the distribution of future trajectories conditioned on observed past paths and then generate samples from this distribution to produce future predictions. Sampling can be \textit{parametric}, such as Gaussian or Gaussian Mixture Models, or \textit{learning-based}, like GANs or Autoencoders. For parametric methods, \citet{8953435} introduced a two-stage framework that employs a modified Evolving Winner-Takes-All (EWTA) loss during sampling to produce diverse and representative hypotheses from Gaussian distributions. A second network in the framework is used to refine the drawn samples to better align them with the true data distribution. In learning-based methods \citep{10203123, 10203823, Ma2020LikelihoodBasedDS, Gilles2021THOMASTH, 9879462, 9812254, Anderson2019StochasticSS}, models are trained to map the relationship between past and future trajectories into a latent space. At inference, given a latent variable and past trajectory, the model generates plausible forecasts. \citet{Gilles2021THOMASTH} introduced a heatmap-driven method that encodes spatial reasoning and agent interactions, producing interpretable likelihood maps for socially-aware multi-agent sampling. We elaborate further on learning-based methods with latent spaces in Sections \ref{sec:auutoencoders} and \ref{sec:gans}.

On more recent approaches, \textit{diffusion models} \citep{Gu2022StochasticTP, 10657121, Zhong2022GuidedCD, 10378047, 10203706, 10377771, NEURIPS2023_2e57e2c1, LI2023110990, 10203059, 10.5555/3666122.3668965, liao2024characterizeddiffusionspatialtemporalinteraction} have been actively utilized in trajectory forecasting. These models follow a two-phase process: first, Gaussian noise is added to the data to produce a latent representation aligned with a prior; then, a reverse process iteratively denoises it, conditioned on past trajectories and scene context. During inference, starting from random noise, the model refines predictions into diverse, multimodal, and context-aware trajectories. \citet{10203706} proposed MotionDiffuser, which leverages a permutation-invariant transformer-based denoiser with cross-attention mechanisms to integrate scene context, past trajectories, and agent interactions. It employs PCA to compress trajectories into a latent space for computational efficiency and supports controlled sampling via differentiable cost functions, enabling constraints such as collision avoidance. \citet{10377771} introduced the BeLFusion model, a diffusion framework for human motion prediction, which uses a behavior coupler to conditionally sample from a behavioral latent space, ensuring realistic and behaviorally coherent trajectories. \citet{10378047} developed PhysDiff, integrating a motion projection module with a physics simulator to iteratively guide the denoising process, eliminating artifacts like floating and ground penetration, and producing physically accurate trajectories.

Continuing with data-driven models, learning-based prediction methods \citep{7576681, 9138768} aim to learn motion dynamics directly from driving data. In classical Machine Learning (ML) approaches such as Gaussian Processes, Hidden Markov Models, Dynamic Bayesian Networks, and Gaussian Mixture Models (see Section~\ref{sec:ml}), model parameters, including distribution parameters, transition probabilities, and kernel hyperparameters (for GP) --- are estimated from data using probabilistic inference. While ML methods offer interpretability grounded in probabilistic reasoning and perform well in simpler environments, Deep Learning (DL) methods excel in complex, dynamic settings due to their ability to learn hierarchical and non-linear representations. Nevertheless, their generalization is highly sensitive to the diversity and quality of training data, often requiring architectural enhancements or training strategies to improve robustness, which can increase computational cost. Section~\ref{sec:dl} discusses key DL approaches, including Sequential Models, Convolutional Neural Networks, Graph Neural Networks, Transformers, Autoencoders, and Generative Adversarial Networks. The mentioned methods are reviewed in this paper within the supervised learning setting. \textit{Self-supervised learning (SSL)} approaches \citep{Chen2023TrajMAEMA, 10.1007/978-3-031-20047-2_14, zhou2025smartpretrain} in trajectory forecasting are commonly employed to learn transferable representations of motion patterns, addressing the scarcity of labeled datasets by leveraging large volumes of unlabeled data. SSL is typically used to enhance the quality of learned features, assisting the prediction model in better generalization. A common setup involves multi-task training, where SSL objectives are combined with supervised forecasting tasks. These methods often rely on proxy tasks, which do not require manual labels but encourage the model to learn meaningful structure from the data, e.g., in contrastive learning, the model is trained to bring representations of similar trajectories closer together while pushing apart dissimilar ones, thereby encouraging the encoding of discriminative motion features. The specifics of SSL methods are out of the scope of this review, we refer the reader to the study \citep{wang2025deployablegeneralizablemotionprediction} for more details. Another approach, Reinforcement Learning (RL), assumes agents aim to achieve goals by optimizing policies that maximize rewards or minimize costs. RL is particularly suited for trajectory prediction tasks involving interactions and decision-making, as it estimates underlying cost functions to identify optimal policies. RL approaches are further elaborated in Section \ref{sec:rl}.

In mixed traffic, where autonomous vehicles operate alongside human-driven vehicles, safe navigation requires understanding and interpreting the implicit and explicit ways road users interact. Incorporating interaction-related factors into trajectory prediction models \citep{Ivanovic2018GenerativeMO, 9156688, 10418557, 10656248, 9010989, Zhu2019StarNetPT, 9025431} can improve AV operation in dense, crowded environments. This survey discusses interaction modeling across various approaches, highlighting its benefits in developing socially compatible trajectory prediction systems. Another line of approaches focuses on intention-aware models. These methods \citep{7487409, Deo2018HowWS, 8766889, 7412746, 6766565, girase2021loki, 9879291} incorporate the vehicle's maneuver intentions to predict its future states. Maneuvers are defined as short-term, goal-oriented decisions comprising a sequence of continuous states that collectively lead to achieving a global objective. They can be broadly categorized into two types: lateral maneuvers (e.g., maintaining the current lane, turning left or right, changing lanes) and longitudinal maneuvers (e.g., maintaining speed, accelerating, or decelerating). Integrating intention-awareness introduces an intermediate layer into prediction models, where future forecasts are informed by planning-based reasoning.

\subsection{Output Modalities}
\label{sec:output_modalities}

For \textit{\textbf{object-centric}} trajectory prediction, where forecasts are centered around observed agents, prediction outputs are categorized as \textit{unimodal} and \textit{multimodal}. \textit{Unimodal} outputs are either deterministic trajectory points or parameters of a unimodal trajectory distribution \citep{8672889, 8317913, 8917228, Diehl2019GraphNN}. However, driving behavior is inherently multimodal. For instance, in a lane with arrows indicating both straight and left turns, a vehicle could either proceed straight or turn left. Until the driver signals their intent (e.g., by activating an indicator or beginning to incline left), both options should be considered equally plausible. Ignoring such multimodal characteristics can negatively affect tasks like collision avoidance at intersections or during lane merging. Therefore, models with \textit{multimodal} outputs \citep{Zyner2018NaturalisticDI, Park2018SequencetoSequencePO, Lee2017DESIREDF} generate predictions as a multimodal probability distribution to better reflect real-world traffic behavior. These models can produce a set of plausible future trajectories, each associated with a probability that reflects the likelihood of that path being followed.

A set of studies focuses on predicting \textit{maneuver intentions} as the output \citep{8290702, 10.1109/IVS.2017.7995919, Phillips2017GeneralizableIP, Ding2019PredictingVB, Lee2017ConvolutionNN}. Common maneuver categories include changing lanes, staying in the current lane, going straight, and turning left or right. For the joint prediction of trajectories and maneuvers, two main approaches exist. The first approach involves multi-task models \citep{liang2020}, where maneuvers intentions and trajectories are predicted as separate outputs but share a common feature extraction module. The second approach \citep{Huang2020LongTermPT, Goldhammer2018IntentionsOV, 9294491, 9008118, Kotseruba2020DoTW} predicts maneuvers and trajectories separately, with the predicted maneuvers used to guide trajectory prediction.

For \textit{\textbf{occupancy-centric}} models, the environment is typically divided into a grid by a perception module, and the prediction module \citep{8460874, Schreiber2018LongTermOG, Kim2017ProbabilisticVT} estimates the likelihood of each cell being occupied at future time steps \citep{9879534, 9811830}. \textit{Occupancy grids} are very natural for integration with the planning module; however, the resolution of the grid controls the balance between precision and computational demands. A potential solution is to use a flexible grid resolution that adapts to the environment: finer grids for dense urban traffic and coarser grids for highways. \textit{Occupancy flow} extends the concept of occupancy grids by also capturing the motion dynamics within these cells \citep{10204893, Li2024EndTE, Casas2021MP3AU}. While occupancy grids generate a sequence of static occupancy maps for future time steps, occupancy flow explicitly models dynamics by jointly predicting occupancy and flow fields, encoding how occupied regions are expected to move. Both formats learn from sequences of input frames and do not require tracking, as they derive spatial and temporal information directly from the input data.

\subsection{Performance Evaluation}
\label{subsec:formulas} 

The most common evaluation setting in existing literature for trajectory prediction models is the \textit{open-loop mode}, where predicted trajectories are evaluated by directly comparing them to ground-truth future paths, without quantifying the impact on downstream modules. Further we review evaluation metrics for both unimodal and multimodal methods discussed in the Section \ref{sec:output_modalities}, where $\zeta_{i,t}$ and $\hat{\zeta}_{i,t}$ denote the predicted and ground truth trajectories for the \textit{i-th} agent at time step $t$, respectively.

\begin{itemize}
    \item \textbf{Root Mean Squared Error (RMSE)} computes the average of the squared error between predicted $\zeta_{i,t}$ and ground truth $\hat{\zeta}_{i,t}$ trajectories at time step $t$ for $n$ agents: 
    \begin{equation*}
        RMSE = \sqrt{\frac{1}{n} \sum_{i=1}^{n} |\zeta_{i,t} - \hat{\zeta}_{i,t}|^2}
    \end{equation*}

    \item \textbf{Average Displacement Error (ADE)} estimates the average L2 distance between predicted and ground truth trajectories over the entire prediction horizon $\Delta t$:
    \begin{equation*}
        ADE = \frac{1}{n \times \Delta t} \sum_{i=1}^{n} \sum_{t=1}^{\Delta t} ||\zeta_{i,t}- \hat{\zeta}_{i,t}||
    \end{equation*}

    For multimodal predictions, the metric \textbf{MinADE\_k} computes the L2 distance between the ground truth trajectory and the closest prediction out of $k$ possible trajectories:
    \begin{equation*}
        MinADE\_k = \frac{1}{n \times \Delta t} \min_{k} \sum_{i=1}^{n} \sum_{t=1}^{\Delta t} ||\zeta^k_{i,t}- \hat{\zeta}_{i,t}||
    \end{equation*}

    \item \textbf{Final Displacement Error (FDE)} estimates the average L2 distance between predicted and ground truth trajectories at the final time step $\Delta t$:
    \begin{equation*}
        FDE = \frac{1}{n} \sum_{i=1}^{n} ||\zeta_{i, \Delta t} - \hat{\zeta}_{i, \Delta t}||
    \end{equation*}

    For multimodal predictions, FDE can be extended to \textbf{MinFDE\_k}:
    \begin{equation*}
        MinFDE\_k = \frac{1}{n} \min_{k} \sum_{i=1}^{n} ||\zeta^k_{i, \Delta t} - \hat{\zeta}_{i, \Delta t}||
    \end{equation*}

    \item \textbf{Negative Log Likelihood (NLL)} evaluates the accuracy of probabilistic prediction algorithms by measuring the likelihood of the ground truth trajectory under the predicted distribution $q(\hat{\zeta}_{i,1:\Delta t}| \Theta)$, where $\Theta$ represents the model parameters:
    \begin{equation*}
        NLL = \sum_i -\log (q(\hat{\zeta}_{i,1:\Delta t}|\Theta))
    \end{equation*}
\end{itemize}

When object IDs are preserved, such as in evaluations using ground-truth tracklets (past trajectories), computing prediction metrics is straightforward. However, in scenarios where ground-truth identifiers are unavailable, predicted and ground-truth objects must first be matched, typically using IoU thresholds or nearest-centroid distance. Only then can accuracy metrics be calculated. In such settings, it is essential to report the \textbf{Miss Rate (MR)}, which quantifies the proportion of agents entirely missed by the perception-prediction pipeline and thus never reported to the planner for collision avoidance. Another critical metric that should become standard in benchmarking is \textbf{inference time}, as it directly reflects the model’s suitability for real-time deployment.

In \textit{closed-loop evaluation}, prediction models are assessed within the full autonomy stack \citep{10.5555/3666122.3669245}, capturing how prediction errors propagate to downstream planning and impact ego-vehicle safety. In this setting, the models can be evaluated from the point of inter-frame prediction consistency, by measuring how frequently the model triggers unnecessary rerouting. Another important factor is the model’s level of conservatism, which determines which predictions should be passed to the planner. Overly conservative models may output fully occupied grids, causing the ego-vehicle to stop unnecessarily, even when feasible paths exist. More on discussion on open and closed-loop evaluations we refer the reader to the study \citep{wang2025deployablegeneralizablemotionprediction}.

Finally, for proper cross-dataset evaluation and deployment, it is essential to account for differences in sampling rates to ensure fair performance comparisons. For example, if a model is trained on data sampled at 20 frames per second (FPS) and deployed in a 10 FPS setting, downsampling must be applied. Similarly, if the model is trained to predict trajectories at 5 steps per second, this defines the motion dynamics it learns. To deploy such a model in a 10 Hz setup, resampling or appropriate temporal alignment is required. Some works to avoid these challenges caused by discretizations shift to continuous domains by representing trajectories as continuous stochastic process \citep{li2020scalable}.

\begin{table*}[t]
\caption{Benchmark Datasets for Trajectory Prediction in Autonomous Driving}
\centering
\begin{adjustbox}{max width=\textwidth}
\renewcommand{\arraystretch}{1.2}
\begin{tabular}{|p{3.5cm}|p{2cm}|p{1.5cm}|p{5.5cm}|p{7.8cm}|}
\hline 
\textbf{Dataset} & \textbf{Modality} & \textbf{Size} & \textbf{Description} & \textbf{Leaderboard} \\ \hline 

KITTI \citep{Geiger2013IJRR}, prediction split from \citep{Marchetti2020MANTRAMA}  
& RGB Camera, \: LiDAR, \:\:\: GPS/IMU 
& 1.5 h 
& Includes 6-second segments. Training set: 8,613 top-view trajectories, test set: 2,907 samples. 
& MANTRA \citep{Geiger2013IJRR} (ADE@4s:0.74, FDE@4s:1.68) \\ \hline

Daimler \citep{10.1007/978-3-642-40602-7_18} & Stereo RGB, Camera & 19,612 images & Short sequences less than 2 seconds, contains 19,612 stereo image pairs. & - \\ \hline

Ko-PER \citep{Strigel2014TheKI} & RGB Camera, LiDAR & 4 chunks, 6.28 min each & Seq. 1 has 4 chunks capturing several agents crossing the intersection, Seq. 2 and Seq. 3 featuring single vehicle at the intersection. & - \\ \hline

ApolloScape \citep{10.1609/aaai.v33i01.33016120} & RGB Camera, LiDAR, \hspace{0.3cm} GPS/IMU & 2 h & Each frame annotations contain objects type, position in world coordinates, dimensions, and heading indicating the steering angle. & AI-TP \citep{9723649} (WSADE:1.15, WSFDE:2.12)  \:\:\:\:\:\:\:\: S2TNet \citep{pmlr-v157-chen21a} (WSADE:1.16, WSFDE:2.17)  \\ \hline

PIE \citep{9008118}  & RGB Camera & 6 h & Collected over 6 hours of driving. Contains intention estimation and features to study pedestrians' behaviour. & SGNet \citep{9691856} (MSE@0.5:34, MSE@1.5:442)  \:\:\:\:\:\: Bitrap-D \citep{9345445} (MSE@0.5:41, MSE@1.5:511) \:\:\:\:\:\: PIE\_traj \citep{9008118} (MSE@0.5:58, MSE@1.5:636) \\ \hline

Argoverse I  \citep{Argoverse} &  RGB Camera, LiDAR, \:\:\: GPS/IMU & 324,557  trajectories & Collection of 324,557 scenarios, each 5 seconds long featuring 2D BEV trajectories. The dataset is accompanied with high-definition maps. & HPNet  \citep{tang2024hpnet} (MinADE:1.59, MinFDE:3.46) \:\:\:\:\:\: iDLab-SEPT  \citep{lan2023sept} (MinADE:1.44, MinFDE:3.17)  \:\:\:\:\:\: 	QCNet-AV1  \citep{10203873} (MinADE:1.52, MinFDE:3.34)  \\ \hline

Argoverse II \citep{Argoverse2} \citep{TrustButVerify}  &  RGB Camera, LiDAR, \:\:\: GPS/IMU  & 250,000 trajectories & Collection of 250,000 scenarios, each 11 seconds long featuring 2D BEV trajectories and headings. & 	iDLab-SEPT++ (MinADE:1.41, MinFDE:3.51)   iDLab-SEPT  \citep{lan2023sept} (MinADE:1.48, MinFDE:3.7)  \:\:\: 	GACRND-XLAB (MinADE:1.56, MinFDE:3.91) \\ \hline

nuScenes \citep{nuscenes2019} &  RGB Camera, LiDAR, \:\:\: GPS/IMU, Radar  & 6 h &  1,000 urban scenes, including LiDAR, radar, and cameras data, paired with high-definition maps. & UniTraj \citep{feng2024unitraj} (MinADE@5:1.18, MinFDE@1:6.6) 
FRM \citep{park2023leveraging} (MSE@0.5:41, MSE@1.5:511) \:\: LaPred++ \citep{Kim2021LaPredLP} (MinADE@5:1.24, MinFDE@1:7.58)  \\ \hline

WAYMO \citep{9709630} & RGB Camera, LiDAR, \:\:\: GPS/IMU, Radar & 570 h & 570 hours of unique data across 1,750 km of roadways. Over 100,000 scenes collected. & MTR\_v3 \citep{shi2022mtra} (mAP:0.49, minADE:0.55) \:\:\:\:\:\:\:\:\:\:\:\:\:\: ModeSeq (mAP:0.47, minADE:0.56) \:\:\:\:\:\:\:\:\:\:\:\:\:  RMP\_Ensemble (mAP:0.47, minADE:0.55) \\ \hline

Lyft Level 5 \citep{Houston2020OneTA} &  RGB Camera, LiDAR,\:\:\: GPS/IMU, Radar & 1,118 h & Collection of 170,000 scenes, each 25 seconds long, including vehicles, cyclists, and pedestrians. & - \\ \hline
\end{tabular}
\end{adjustbox}
\label{tab:benchmark_datasets}
\end{table*}

\subsection{Datasets and Leaderboard}
\label{subsec:datasets}

Assessing the effectiveness of trajectory prediction models require diverse test scenarios to evaluate the models' performance under various factors. The test scenarios can vary widely in complexity, location, and traffic density, including intersections, highways, mixed urban streets, crowded public spaces, residential areas, and more. Each scenario presents unique challenges and variability that the model must handle effectively to demonstrate its practicality and potential for real-world autonomous driving applications. Table \ref{tab:benchmark_datasets} lists trajectory prediction datasets that can be utilized to test forecasting models. It provides details on the modalities used to collect the datasets, along with short descriptions and leaderboard solutions. For some datasets, certain details are omitted, encouraging readers to refer to the datasets' official websites for the most up-to-date information.

The KITTI dataset \citep{Geiger2013IJRR} is a well-known resource in the autonomous driving community. The official dataset does not provide an official split for trajectory prediction. The lack of a standardized data split causes inconsistencies in baseline comparisons across different models. Marchetti et al. introduced a well-defined split of the KITTI data. Samples are collected in 6-second segments with 2 seconds of past data and 4 seconds of future data. The dataset comprises 8,613 top-view trajectories for training and 2,907 for testing, supplied with top-view maps. The recent addition by the authors is KITTI-360 \citep{Liao2022PAMI}, collected over a distance of 70 kilometers and geo-localized with OpenStreetMap. The dataset serves as a benchmark for scene understanding and is a step forward toward map-less navigation, encouraging models to use only the global map without relying on the granularity of HD maps. The Daimler dataset \citep{10.1007/978-3-642-40602-7_18} was introduced for predicting pedestrian paths over short time horizons (less than 2 seconds). The dataset includes 19,612 stereo image pairs, where 12,485 images are manually labeled with pedestrian bounding boxes, and 9,366 images contain measurements from a pedestrian detector. The Ko-PER Intersection dataset \citep{Strigel2014TheKI} provides data for intersection analysis. The dataset contains raw LiDAR measurements and undistorted monochrome camera images. It comprises 3 sequences: sequence 1 captures different agents crossing the intersection, sequence 2 contains data of a single car performing a right turn, and sequence 3 features a car crossing the intersection straight ahead. The dataset comes with a viewer and sample code to ease inspection and data transformation for researchers.

Next, the ApolloScape dataset \citep{10.1609/aaai.v33i01.33016120} contains camera-based images and LiDAR point clouds of trajectory data collected in Beijing, China. The dataset was captured under various lighting conditions and features complex traffic flows. Each training file is a sequence 1 minute long. Annotations contain object types, positions, dimensions, and headings. The evaluation protocol defines prediction over the next 3 seconds. The PIE \citep{9008118} dataset was collected using a monocular dashboard camera over 6 hours of driving in Toronto, Canada. The dataset captures various pedestrian behaviors at crosswalks under different urban settings and weather conditions. It is annotated with bounding boxes, occlusion flags, pedestrian actions and intentions, and ego-vehicle information such as speed, heading angle, and GPS coordinates. The NuScenes dataset \citep{nuscenes2019} provides data for trajectory prediction of various traffic participants, such as vehicles, pedestrians, and cyclists. The data is collected from 1,000 urban scenes and includes sensor data from LiDAR, radar, and cameras, paired with high-definition maps.

The WAYMO Motion Prediction Challenge \citep{9709630} introduced a diverse interactive motion dataset. This dataset includes around 100,000 scenes, each 20 seconds long, and features detailed 3D bounding boxes and high-definition maps. The Argoverse dataset is broadly used by trajectory prediction community. It is released in two versions. Version I \citep{Argoverse} is a collection of 324,557 scenarios, each 5 seconds long, featuring 2D BEV trajectories of tracked objects. It includes complex driving scenarios such as intersections and lane changes. The dataset is accompanied by high-definition maps, aiding in the precise prediction of object motion. Version II \citep{Argoverse2,TrustButVerify} was released two years later. It consists of 250,000 scenarios, each 11 seconds long, capturing the 2D trajectories and headings of tracked objects. The dataset includes complex interactions, such as vehicles yielding to pedestrians and cyclists navigating city streets. Compared to the first version, this collection has longer sequences and more diverse scenarios. Lastly, the Lyft Level 5 dataset \citep{Houston2020OneTA} comprises over 1,000 hours of data collected by 20 autonomous vehicles along a set route in California over four months. The dataset includes 170,000 scenes, each 25 seconds long, detailing the precise movements of nearby vehicles, cyclists, and pedestrians, and features a high-definition semantic map and aerial view.

\subsection{Prediction Paradigms}

\begin{figure*}[t!]
\centering
   \includegraphics[width=1\textwidth]{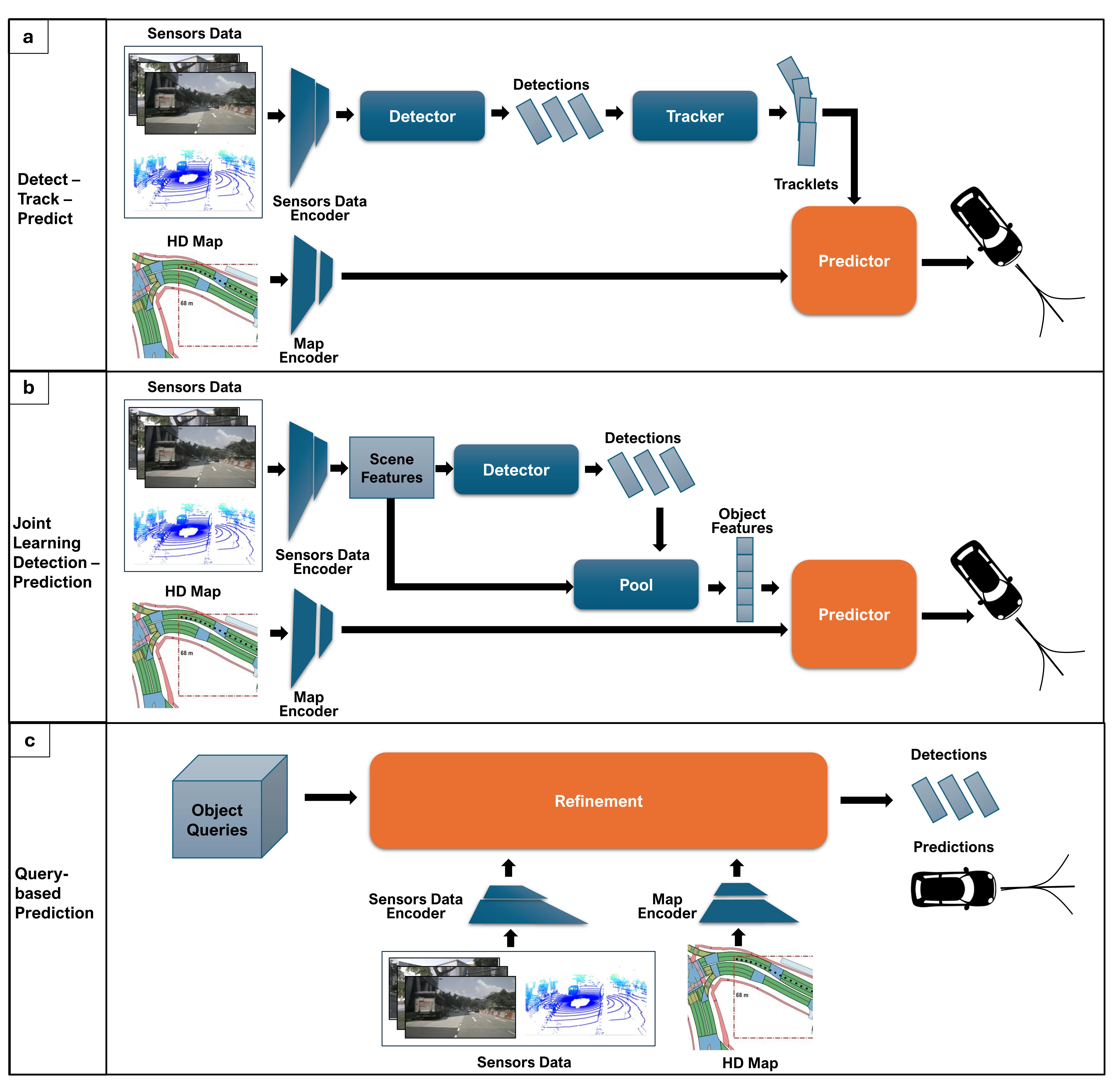}
    \caption{Overview of prediction paradigms in autonomous driving: (a) Detect-Track-Predict: a modular pipeline where detections initialize and update tracklets, which are then passed as past trajectories to the prediction module; (b) Joint Learning Detection–Prediction: an integrated setup where detections are pooled into object-level representations and directly consumed by the prediction module; (c) Query-Based Prediction: a query-driven framework where spatial and temporal object queries attend to encoded scene and map features to jointly produce detections and predictions.}
   \label{fig:paradigms_fig}
\end{figure*}

This section discusses three existing prediction paradigms: Detect-Track-Predict, Joint Learning Approaches, and Query-Based Approaches, which are graphically illustrated in Figure \ref{fig:paradigms_fig}. For each paradigm, we outline the general concept and highlight its advantages and limitations for integration.

\subsubsection{Detect-Track-Predict}

The Detect-Track-Predict paradigm is a modular cascade approach consisting of detection, tracking, and prediction modules. First, the detection module localizes objects from sensor data. Next, the tracking module initializes tracklets for the detected objects and maintains their trajectories consistently over time. Finally, the prediction module takes the active tracklets and forecasts their future positions. This approach is one of the most common implementations in autonomous driving and is broadly used in the literature. Its modular design makes it easier to debug, as most datasets provide ground truth annotations for each of the modules. Additionally, the extensive research output for detectors \citep{Mao2023} and trackers \citep{Agrawal2024}, covering both 2D and 3D modalities, further supports this paradigm development.

\begin{figure*}[t!]
\centering
   \includegraphics[width=1\textwidth]{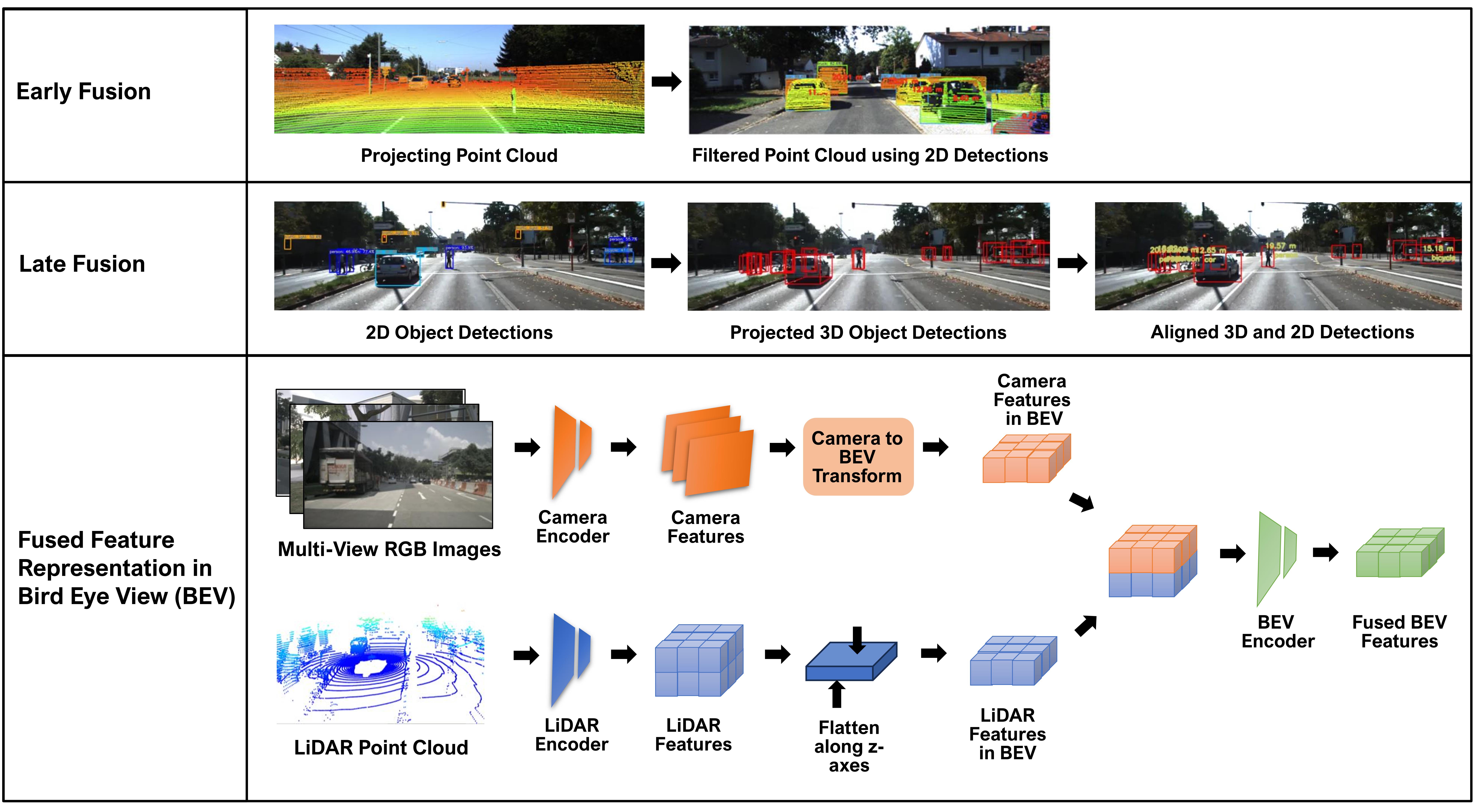}
   \caption{Different fusion approaches for LiDAR and camera data: early fusion, late fusion and BEV fusion.}
   \label{fig:fusion_fig}
\end{figure*}

The latest detection approaches leverage data fusion from cameras and LiDAR sensors to enhance detection and tracking accuracy as shown in Figure \ref{fig:fusion_fig}. For early fusion, a common implementation involves detecting objects in the image, projecting the point cloud onto the 2D image plane, and matching the projected points with the detected objects for depth estimation \citet{Guidolini2018HandlingPI}. Late Fusion operates at the detection stage, combining LiDAR and camera detection outputs, by projecting LiDAR’s 3D bounding boxes onto the 2D image plane and aligning 2D representation of LiDAR detections with the 2D detections from the camera. Another fusion strategy, Augmenting Point Clouds, enhances LiDAR data by projecting point clouds onto images and enriching each point with matching image features. This augmented representation is then fed into a LiDAR-only detection method. \citet{Vora2019PointPaintingSF} proposed PointPainting, which projects LiDAR points onto semantic segmentation outputs from the camera and appends the corresponding class scores to each point. Camera-to-LiDAR projections can result in a loss of rich semantic information. Bird's-eye view (BEV) Fusion mitigates this issue by combining features from different modalities within a unified BEV representation, providing a more comprehensive spatial understanding, but resulting in a prolonged inference time, e.g., as detection inference speed of 190 ms for BEVFusion \citep{liu2022bevfusion}. For a detailed overview, we refer the reader to the multi-sensor fusion framework presented in \citep{10160968}.

The trackers are responsible of maintaining consistent object identities and motion estimates across frames. Kalman Filter-based trackers \citep{nagy2023dfrfastmotdetectionfailureresistant, Aharon2022BoTSORTRA, zhang2022bytetrack}, leverage predicted object states to mitigate detection inaccuracies caused by sensor noise or missed detections. Advanced trackers \citep{nagy2024robmotrobust3dmultiobject} address partial occlusions, ensuring robust trajectory maintenance and minimizing false positives that lead to ghost tracklets. Prediction models rely on the precision of trajectories over the past 10–20 timestamps, making robust tracklets essential for accurate forecasting. Excessive ghost tracklets result in non-existent predictions being passed to the planner, potentially degrading overall system performance.

While the Detect-Track-Predict paradigm is modular and intuitive, its sequential nature causes errors to propagate across modules, with detection inaccuracies leading to unreliable tracklets and degraded predictions. Additionally, the modular independence of its components often results in suboptimal performance due to the lack of end-to-end optimization. These limitations motivated the need for alternative paradigms that integrate perception and prediction more cohesively.

\subsubsection{Joint Learning Models}

Joint learning (perception \& prediction) models \citep{hu2021fiery, 8578474, kim2022pnpnet} are designed to optimize object detection and trajectory prediction under a unified training objective. These models can operate without tracker, capturing temporal dependencies directly from sequential sensor inputs frames. As shown in Figure~\ref{fig:paradigms_fig}, a pool module can be integrated to bridge the gap between detection and prediction. The pool aggregates and encodes object features from detection outputs into a unified representation tailored for trajectory forecasting. It also organizes and streamlines the inputs to the prediction module, dynamically filtering or prioritizing detections based on their relevance to the forecasting task. Training is typically guided by a compound loss over both detection and prediction objectives.

Depending on the design, joint learning can involve different degrees of integration. Some models share intermediate scene features between perception and prediction, enabling contextual reuse across tasks, e.g, FIERY \citep{hu2021fiery} reuses BEV-encoded features for both detection and forecasting. PnPNet \citep{kim2022pnpnet} supports shared gradient flow across modules, allowing the prediction loss to propagate through the detector and encoder. Alternatively, structurally decoupled modules may be co-optimized via a combined loss, as in FaF \citep{8578474}, which performs detection and forecasting using a shared LiDAR backbone but without full gradient sharing between modules. The limitation of these models is that the tight coupling of modules in joint learning blurs task boundaries, making it challenging to debug and attribute responsibility for poor performance. 

\subsubsection{Query-Based Prediction Models}
Query-based prediction models \citep{10205186} unify detection and forecasting by leveraging learnable object queries that are iteratively refined to predict agent trajectories. These queries encode spatial and temporal information and directly attend to scene and map features, removing the need for separate detection and prediction stages. As shown in Figure~\ref{fig:paradigms_fig}, predictions are generated holistically, reducing error propagation across subsystems.

\citet{10203873} propose DenseTNT, which uses anchor-based and anchor-free trajectory queries to refine multi-agent predictions from shared scene encodings. \citet{hu2023planning} present UniAD, a full-stack framework that uses task-specific queries (e.g., TrackFormer, MotionFormer) to handle detection, tracking, forecasting, and planning jointly. \citet{Casas2024DeTraAU} introduce DeTra, where spatial and temporal queries jointly estimate object poses and future motion directly from LiDAR and HD maps. \citet{10204893} propose a scene-centric model where spatio-temporal BEV queries guide occupancy and flow prediction over time.

Query-based methods offer flexibility and holistic scene understanding but may face challenges in real-time deployment in dense scenarios due to the computational overhead of iterative query refinement. Additionally, these methods depend heavily on accurate scene encoding and high-quality input data, such as HD maps and LiDAR, for reliable forecasting.
\section{Survey of Modelling Methods}
\label{sec:survey}

This section describes the refined taxonomy of trajectory prediction approaches. To systematically categorize the wide range of prediction models in the literature, the taxonomy is structured based on modeling methods. It builds upon the categorizations presented in reviews \citep{9756903, BHARILYA2024100733}, and is further extended based on a review of recent works in which prediction models were systematically tagged and grouped. Each category of methods is discussed in a dedicated subsection, highlighting their main characteristics and limitations. For a method-by-method analysis of limitations, we refer the reader to the review \citep{BHARILYA2024100733}.
 
\subsection{Taxonomy}

\begin{figure*}[t!]
\centering
   \includegraphics[width=1\textwidth]{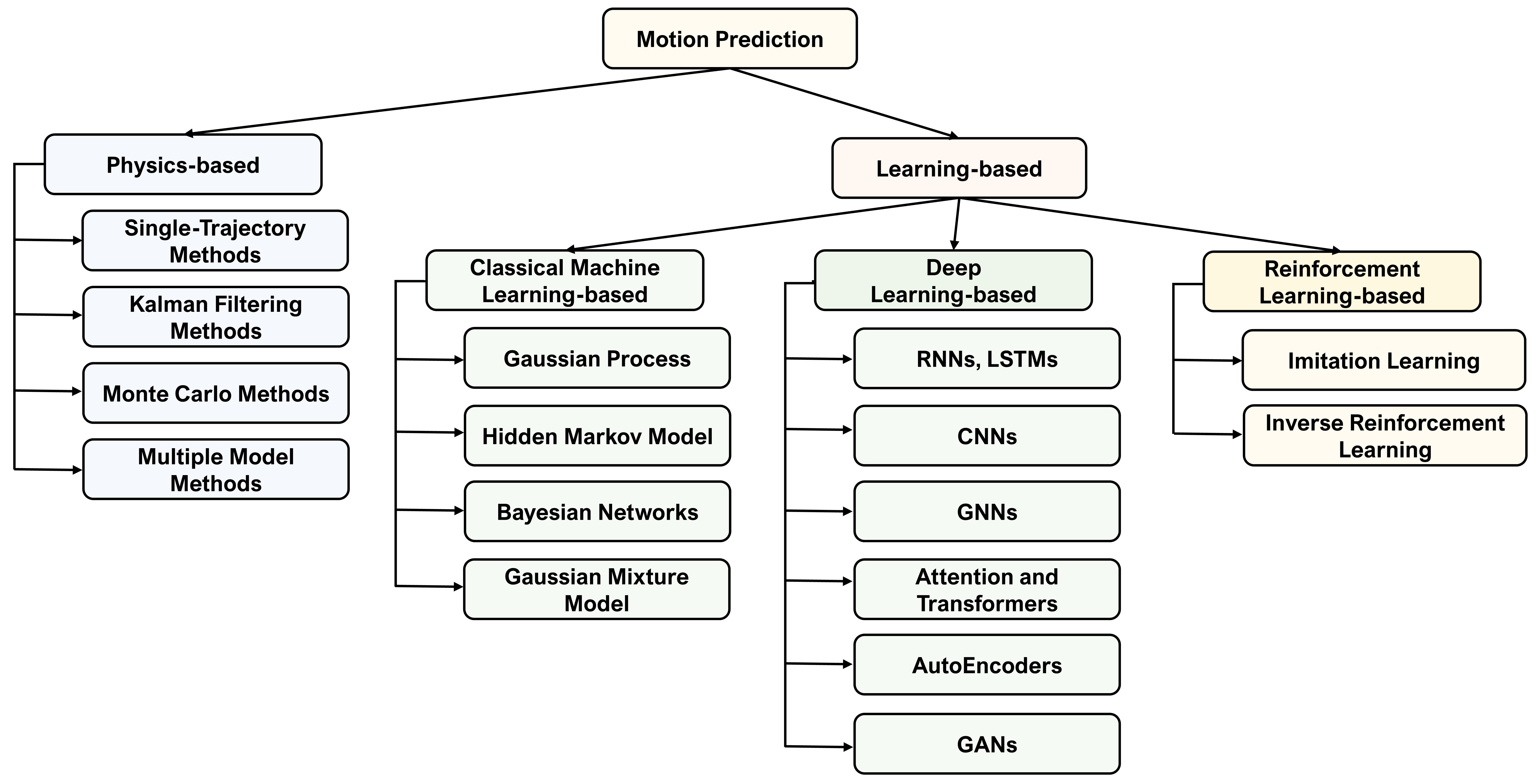}
   \caption{Taxonomy of modeling methods for trajectory prediction, divided into two main branches: physics-based approaches and learning-based approaches. The learning-based category is further subdivided into classical machine learning methods, deep learning architectures and reinforcement learning approaches.}
   \label{fig:taxonomy}
\end{figure*}

In the proposed taxonomy, prediction methods are divided based on the way motion is modeled, i.e., physics-based methods explicitly define motion equations, while learning-based approaches learn motion patterns from observations. Learning-based approaches are further subdivided into classical machine learning, deep learning, and reinforcement learning-based methods (please refer to Figure \ref{fig:taxonomy}). 

\begin{enumerate}
    \item \textbf{\textit{Physics-based methods}} predict motion by simulating a set of equations describing a physics-based model. \textit{Single-Trajectory methods} predict trajectory by directly applying dynamic or kinematic equations to the agent's current state. \textit{Kalman Filtering methods} account for the agent's state uncertainty, which increases during prediction steps and reduces after receiving observations. \textit{Monte Carlo methods} approximate the state distribution by randomly sampling from the input variables and applying a physics model to generate potential future trajectories. \textit{Multiple Model methods} account for motion-mode uncertainty and combine several motion models to better describe the complexity of agent motion behavior.
    
    \item \textbf{\textit{Learning-based methods}}, in contrast to physics-based approaches learn motion dynamics from driving data.
    \begin{enumerate}
        \item \textbf{\textit{Classical Machine Learning (ML) methods}} learn motion dynamics from driving data by fitting function approximators, e.g., Markov Models or Gaussian Processes. In this paper, we consider four main groups of methods: \textit{Gaussian Processes}, \textit{Hidden Markov Models}, \textit{Dynamic Bayesian Networks}, and \textit{Gaussian Mixture Models}.
        \item \textbf{\textit{Deep Learning (DL) methods}} also learn motion dynamics from driving data by leveraging their ability to capture complex and hierarchical patterns. Among DL methods, the following main groups of models are reviewed: \textit{Sequential Models}, \textit{Convolutional Neural Networks}, \textit{Graph Neural Networks}, \textit{Transformers}, \textit{AutoEncoders}, and \textit{Generative Adversarial Networks}.
        \item \textbf{\textit{Reinforcement Learning (RL) methods}} interpret driving as a high-dimensional complex policy. Despite these methods having direct applications for planning, recent works attempt to utilize RL methods for trajectory prediction through learning via a reward system. \textit{Inverse Reinforcement Learning} methods learn a reward function based on expert demonstrations. \textit{Imitation Learning} bypasses the stage of learning reward functions and directly extracts an optimal policy from the data, generating future trajectories similar to expert trajectories.
        \end{enumerate}
    \end{enumerate}

\subsection{Physics-based Methods}
\label{sec:physics-based}

\textbf{\textit{Physics-based}} methods accomplish motion prediction by simulating a set of dynamics equations describing a physics-based model. These methods apply physical laws to estimate steering and acceleration, predicting position, orientation, and speed. To describe vehicle movement, they employ kinematic \citep{9472993} and dynamic models. Commonly used kinematic models include the Constant-Velocity (CV) model, which estimates position and velocity under the assumption of constant velocity; the Constant-Acceleration (CA) model, which estimates position, velocity, and acceleration assuming constant acceleration; and the Constant Turn Rate and Velocity (CTRV) and Constant Turn Rate and Acceleration (CTRA) models, which estimate position and orientation under the assumption of constant yaw rate. Dynamic models account for forces acting on the vehicle, which introduces added complexity when modeling components such as wheels, engines, and gearboxes. These limitations make dynamic models less preferable than kinematic models, with a tendency toward simpler dynamic formulations in practical applications.

The simplest approach to trajectory prediction is to apply the vehicle’s current state to a dynamic model \citep{Lin2000VehicleDA, 1689645, Kaempchen2009SituationAO, 10.1109/TITS.2010.2048314} or a kinematic model \citep{Miller2002AnAP, Lytrivis2008CooperativePP}. This approach assumes perfect state estimation and does not account for uncertainty.
\textbf{\textit{Kalman Filtering (KF) Methods}} address this limitation by modeling the vehicle state as a Gaussian distribution. The state typically includes position $(x, y)$ and velocity $(v_x, v_y)$, and its future estimate is computed using a state transition matrix defined by the underlying motion model. The filter updates the covariance of the state estimation error based on system dynamics, predicts the expected measurement, and corrects the state estimate using the Kalman gain derived from the prediction-error residual.
Variants of Kalman Filters, such as the standard linear Kalman Filter \citep{Ammoun2009RealTT}, the Extended Kalman Filter (EKF) \citep{5164400}, and the Unscented Kalman Filter (UKF) \citep{4357713} --- have been employed for trajectory prediction in both linear and nonlinear systems.

While KF methods assume Gaussian distributions, \textbf{\textit{Monte Carlo Methods}} approximate state distributions without such assumptions. These methods randomly sample input variables (e.g., initial conditions, sensor noise) and apply a physics model to generate potential future trajectories \citep{Rudnyk2018TrajectoryPS}. Several works extend physics-based methods by embedding map information and incorporating road constraints. \citet{6728549} proposed a stochastic filter that uses digital map information to select a set of reasonable trajectories by projecting detected vehicles onto the centerlines of map lanes. \citet{7406377} mapped environments to location maps representing pedestrian locations from occupancy grids of sidewalks and obstacles, converting the prediction problem into a planning task by modelling pedestrian destinations as latent variables. \citet{COSCIA201881} used polar grids centered at predicted agents to construct point-wise circular distributions, combining them into statistically smoothed predictions. Next, \citet{8569434} incorporated pedestrian dynamics, traffic rules, road connectivity, and semantic regions to obtain reachable pedestrian occupancy. \citet{8550300} employed graph representations of road geometry to predict pedestrian motion, assuming motion along graph edges representing sidewalks and crosswalks, with a unicycle model ensuring trajectory adherence to edges.

In real traffic, agent motion behavior is complex and cannot be described by a single model, often involving linear movements, turns, and acceleration/deceleration modes. Since motion modes of other vehicles are not directly observable, \textbf{\textit{Multiple-Model (MM)}} methods address motion uncertainty by considering a set of predefined motion models. A bank of elemental filters is used, each corresponding to one model in the set. At each time step, the filters independently generate state estimates, and their outputs are evaluated and combined typically through weighted fusion to represent complex or switching motion patterns. For further details on handling discrete and continuous-based uncertainties, the review on MM methods \citep{1561886} provides comprehensive insights. The \textbf{\textit{Interacting Multiple Model (IMM)}} estimator \citep{640267, Xie2018VehicleTP} is a hybrid filter with nearly linear computational complexity, offering a balance between complexity and performance. \citet{Kaempchen2004IMMOT} proposed an IMM approach utilizing constant speed, constant acceleration, and acceleration change models to compute vehicle states as weighted state estimates. Other IMM designs accounting for road constraints are presented in \citep{826310, 10.1117/12.391981}. Recent works, such as Interacting Multiple Model Kalman Filter (IMM-KF) \citep{9234702}, employs intention-based models and optimization-based projections for generating non-colliding predictions using a hierarchical approach for multiple vehicles. \citet{7266697} proposed a Unified Traffic Situation Estimation Model using IMM with constant and target velocity models, incorporating predefined environmental geometry to estimate future trajectories.

\begin{figure*}[t!]
\centering
   \includegraphics[width=1\textwidth]{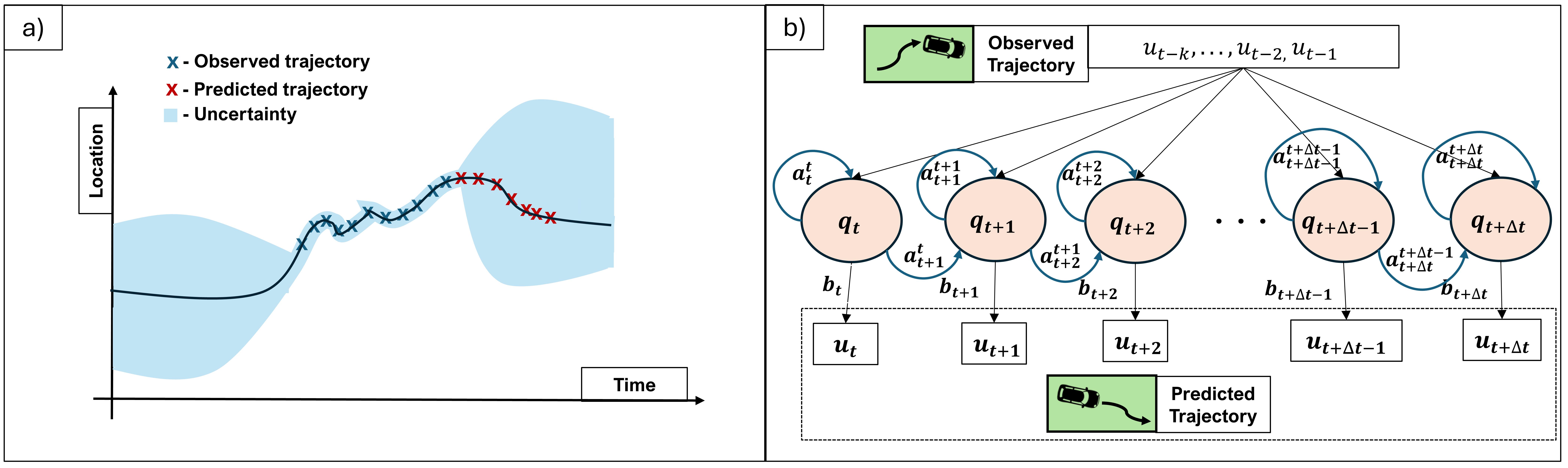}
   \caption{ML approaches for trajectory prediction: (a) - GP modelling observed and predicted trajectories showing uncertainty bounds; (b) - HMM representing latent states and their temporal transitions, with associated observations.}
   \label{fig:ml_overview}
\end{figure*}

\begin{figure*}[t]
\centering
   \includegraphics[width=1\textwidth]{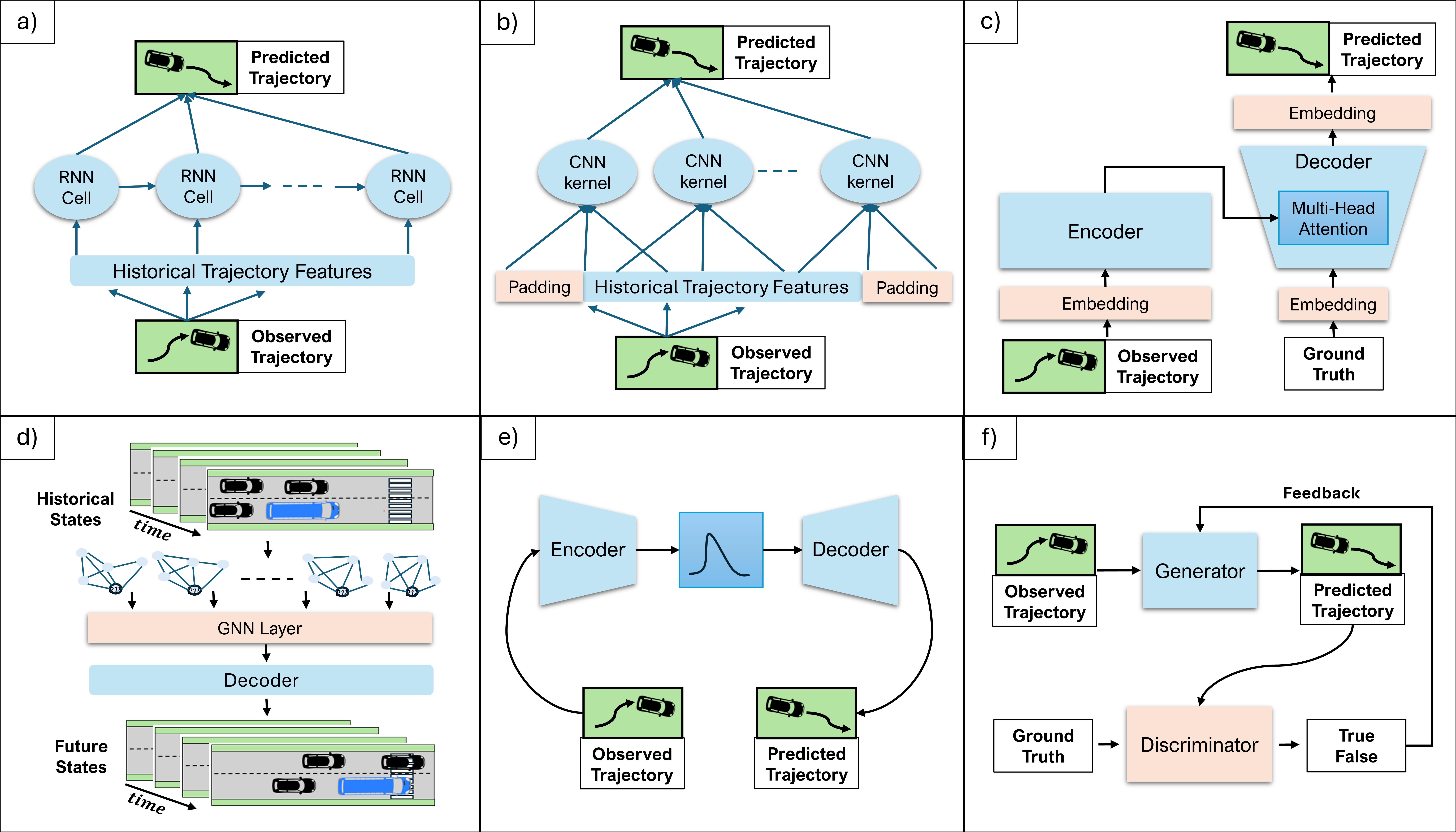}
   \caption{Main DL approaches presented schematically in application to prediction where the input is past trajectory: (a) - RNN/LSTM models, (b) - convolution models, (c) - transformers, (d) - GNNs, (e) - autoencoders and (f) - GANs.}
   \label{fig:dl_overview}
\end{figure*}

\subsubsection{Strengths and Limitations of Physics-based Models}

Physics-based models are efficient for short-term trajectory prediction, particularly in situations requiring rapid adaptation to abrupt or unexpected maneuvers. Their dependence on current or recent vehicle state makes them computationally lightweight, but also limits their ability to incorporate longer motion histories. Constant motion models do not account for uncertainty, whereas Kalman Filter variants introduce Gaussian noise modeling but remain limited in flexibility. Monte Carlo methods offer greater expressiveness for modeling uncertainty without relying on specific distributional assumptions, but they are computationally demanding and require careful parameter tuning and scenario-specific constraints to yield reliable results.

To overcome limitations of physics-based methods and combine their advantages with the power of learning-based approaches, hybrid models \citep{Kim2015BRVOPP, Ma2019WassersteinGL, 10.1007/978-3-031-19830-4_22, GENG2023104272} have been proposed by the community.  \citet{Kim2015BRVOPP} proposed a hybrid method that uses the Reciprocal Velocity Obstacles (RVO) framework to predict pedestrian trajectories in dynamic environments, enhanced with Ensemble Kalman Filtering (EnKF) and Expectation-Maximization (EM) for real-time learning and parameter refinement. \citet{Ma2019WassersteinGL} introduced a deep latent variable model built on a Wasserstein auto-encoder, enabling the automatic satisfaction of kinematic constraints. \citet{10.1007/978-3-031-19830-4_22} presented Neural Social Physics (NSP), a Neural Differential Equation model that integrates an explicit physics-based framework as an inductive bias, paired with deep neural networks for system parameter estimation and stochastic dynamics modelling.

\subsection{Machine Learning-based Methods}
\label{sec:ml}

\textbf{\textit{Machine Learning-based}} methods learn motion dynamics from driving data by fitting function approximators. Figure \ref{fig:ml_overview} provides an overview of two key ML techniques discussed in this section: Gaussian Processes (GPs) and Hidden Markov Models (HMMs). Additionally, the section discusses prediction models based on Gaussian Mixture Models (GMMs), which combine multiple GPs, and Dynamic Bayesian Networks (DBNs), which capture interactions between agents. Each method is discussed in dedicated subsections, elaborating on the theoretical foundations, their application to prediction tasks, and relevant studies available in the literature.

\subsubsection{Gaussian Process}
Gaussian Process (GP)-based prediction models \citep{4359316, 6025208} treat each position at discrete time steps as a random variable governed by a probabilistic prior. To capture the structure of the trajectory and assumptions about smoothness, the GP uses a kernel function to measure the similarity between inputs. This similarity translates into how correlated the outputs (e.g., predicted positions) are, ensuring that similar inputs result in similar outputs. The kernel also defines the covariance matrix, which connects observed inputs and outputs to predictions for unobserved points. During training, the kernel hyperparameters are optimized by maximizing the likelihood of the observed data, ensuring the GP accurately reflects the underlying motion dynamics. Predictions for future positions are made by conditioning the prior distribution on observed data, yielding a posterior distribution with a mean trajectory (future trajectory) and variance (uncertainty) for each future time step.

\citet{QuinteroMnguez2019PedestrianPP} proposed a model based on Balanced Gaussian Process Dynamical Models (B-GPDMs) for pedestrian path and intention prediction, reducing high-dimensional motion data into a latent space to enable smooth trajectory modelling. \citet{Guo2019ModelingMI} utilized a GP velocity field to represent each motion pattern, mapping positions to velocities in a region of interest. The model integrates a Dirichlet Process (DP) prior to avoid assumptions about the number of motion patterns, allowing it to learn them directly from data. Lastly, \citet{Ferguson2015} employed Gaussian Process Mixture Models to detect changes in pedestrians' intent and learn new motion patterns online, enabling the handling of previously unobserved scenarios.

\subsubsection{Hidden Markov Models}

Prediction models based on Hidden Markov Models (HMMs) \citep{Deo2018HowWS, 9334598, 8519506, 4732630, 8500533, Qiao2015ASP} represent trajectories as sequences of hidden states evolving over time. Figure \ref{fig:ml_overview} (b) illustrates HMM-based trajectory prediction, where hidden states ${q_t, q_{t+1}, ..., q_{t+\Delta t}}$ represent latent factors such as intent or motion patterns, linked by transition probabilities $a_{ij}$ that model temporal dependencies. The observed trajectory ${u_{t-k}, ..., u_{t-1}}$ serves as input to infer hidden states using decoding algorithms. These states generate the predicted trajectory ${u_t, u_{t+1}, ..., u_{t+\Delta t}}$ via emission probabilities $b_i(u)$, which define the likelihood of observations given the hidden states. By learning transition and emission probabilities from data, HMMs capture sequential dependencies and enable robust trajectory prediction.

\citet{Rathore2018ASF} proposed a hybrid framework, Traj-clusiVAT, combining trajectory clustering and Markov chain modelling. The clustering algorithm partitions trajectories into representative clusters, and Markov models are trained for prediction. \citet{9776157} proposed an enhanced HMM for behavior prediction, with key improvements including random initial parameter generation to ensure reliability, parameter learning through the Baum-Welch algorithm, and prediction via the Viterbi algorithm combined with Bayesian decision theory. On Continuous HMMs, \citet{Ren2022AMF} proposed a framework for behavior recognition. The CHMM is employed for early detection of lane-changing behavior by analyzing vehicle motion patterns and contextual traffic features.

\subsubsection{Gaussian Mixture Models}
Applying Gaussian Mixture Models (GMMs) for trajectory prediction leverages their ability to capture the multimodal nature of future trajectories. The GMM represents the distribution of possible future paths as a weighted sum of Gaussian components:
\[
p(x) = \sum_{k=1}^K \pi_k \mathcal{N}(x | \mu_k, \Sigma_k),
\]
where \( \pi_k \) is the weight of the \( k \)-th Gaussian component, \( \mu_k \) represents the mean, and \( \Sigma_k \) is the covariance. The mean \( \mu_k \) corresponds to the central point or the most likely position of a trajectory predicted by the \( k \)-th Gaussian component. The covariance \( \Sigma_k \) quantifies the spread and uncertainty of predictions around this mean, capturing both variance along each axis and correlations between them. Together, these parameters enable the model to represent multiple plausible future paths while accounting for prediction uncertainty.

\citet{6232277} applied Chebyshev polynomial decomposition to represent trajectories uniformly and used GMMs to provide multimodal predictions by estimating conditional densities. Combining HMM and GMMs, \citet{Deo2018HowWS} proposed a holistic motion prediction framework that integrates HMM for maneuver recognition, Variational GMM (VGMM) for trajectory prediction, and a Vehicle Interaction Module (VIM) for inter-vehicle interactions. VGMM models the conditional probability of future trajectories using a variational Bayesian approach, improving robustness against overfitting and outliers. In more recent approaches, \citet{Murad} utilize a Transformer architecture to extract sequential dependencies from observed pedestrian trajectories, coupled with a Mixture Density Network (MDN) to estimate GMM parameters.

\begin{table*}[h]
\centering
    \caption{Summary of reviewed DL-based Models relying on RNN, GRU and LSTM (part 1)}
    \label{tab:dl_rnn_lstm_part1}
    \begin{adjustbox}{max width=1.0\textwidth}
    \begin{tabular}{|p{5cm}|c|p{14cm}|}
        \hline
        \textbf{Reference} & \textbf{Name} & \textbf{Description}  \\
        \hline 
        \citet{7780479} (2016) & Social LSTM &  Individual LSTMs model each pedestrian's motion and are connected in a social pooling sharing hidden state information with neighbors.  \\ \hline
        \citet{AlMolegi2016STFRNNST} (2016) & STF-RNN &  Space Time Features-based RNN, extracts spatial data from GPS logs and temporal information from timestamps.   \\ \hline
        \citet{10.5555/3157096.3157270} (2016) & - & Hierarchical policy network (HPN), integrates macro-goals for long-term trajectory planning  and micro-actions using attention mechanisms.  \\ \hline
        \citet{7780942} (2016) & Structural-RNN & Transforms spatio-temporal graphs into a structured RNN-based architecture by employing RNNs for nodes and edges. \\ \hline 
         \citet{Bhattacharyya2017LongTermOP} (2017) &  LSTM-Bayesian & First RNN stream with CNN-encoder for odometry prediction, second RNN stream with Bayesian framework to model uncertainty and condition on odometry. \\ \hline
         \citet{Xue2017BiPredictionPT} (2017) & - &  Uses a bidirectional LSTM to classify trajectories into route classes.   \\ \hline
        \citet{Kim2017ProbabilisticVT} (2017) &  Occupancy-LSTM &  Individual LSTMs for each surrounding vehicle, generating the output as an occupancy grid map.  \\ \hline
        \citet{Varshneya2017HumanTP} (2017) &  SSCN-LSTM & LSTM for short-term dependencies, Spatially Static Context Network (SSCN) for spatial context and polling mechanism for interactions.  \\ \hline
        \citet{Bock2017SelflearningTP} (2017) &  Continuous-LSTM   & Continuous learning with LSTM model, uses Kalman Filter for prediction if new data is not sufficiently covered in dataset.   \\ \hline
        \citet{8317941} (2017) & - &   Uses stacked LSTMs for intention prediction.  \\ \hline
        \citet{10.1109/IVS.2017.7995919} (2017) & - & LSTM for driver intention prediction trained on position, heading, and velocity. \\ \hline
        \citet{Phillips2017GeneralizableIP} (2017) & - &  LSTM for intention prediction at intersection trained on base, history, traffic and rule features. \\ \hline
        \citet{Zyner2018NaturalisticDI} (2018) & - & Combines recurrent model with a mixture density network.   \\ \hline
        \citet{10.1109/IVS.2018.8500493} (2018) & - &  Encodes past trajectories into a context vector, combines it with representations of lateral and longitudinal maneuvers, and decodes into maneuver specific trajectories and probabilities of maneuvers.    \\ \hline
        \citet{8290702} (2018) & - &  A single fully connected layer followed by three recurrent layers.     \\ \hline
        \citet{8578651} (2018) & CIDNN  &  Combines LSTMs for individual motion and a multi-layer perceptron
        for transforming locations into high-dimensional feature space to further predict displacement.   \\ \hline
        \citet{Saleh2018CyclistTP} (2018) & - & Stacked recurrent neural networks for cyclist trajectory prediction.  \\ \hline
        \citet{8354239} (2018) & SS-LSTM &     Uses three LSTMs: to capture individual information, social and scene scale.  \\ \hline
        \citet{Manh2018SceneLSTMAM} (2018) &  Scene-LSTM  &  Two coupled LSTMs trained simultaneously: first group pedestrian LSTMs  modelling movement of agents and second group Scene-LSTMS one per grid-cell.   \\ \hline
        \citet{Hasan2018MXLSTMMT} (2018) &  MX-LSTM  &  Merges tracklets with short sequences of head pose estimations using a joint optimization.    \\ \hline
        \citet{10.1109/ITSC.2018.8569595} (2018)  & -  &  Contains two LSTM blocks: one block to decode driver intentions and second block for predicting.   \\ \hline    
        \citet{Bisagno2018GroupLG} (2018)  &   Group-LSTM  &  Splits pedestrians into groups by clustering similar motion patterns and social-LSTM for prediction.   \\ \hline
        \citet{8545447} (2018) & - &  Social LSTM with context-aware pooling layer for static objects in the neighborhood of a person.  \\ \hline
        \citet{Ivanovic2018TheTP} (2018)  &  Trajectron &  Combines recurrent and graph-based models to capture individual behavior and interactions.  \\ \hline 
        \citet{8461157} (2018) &  Static-LSTM &  Encodes pedestrian using 1D grid; accounts for static obstacles and influences of other pedestrians.    \\    \hline
        \citet{Huynh2019TrajectoryPB} (2019) & - &  Combines Scene-LSTM for capturing traveled paths and Pedestrian-LSTM for individual prediction.   \\ \hline 
        \citet{9010834} (2019) &  STGAT & Combines a graph attention for capturing spatial interactions and LSTM for temporal interactions.  \\ \hline          
    \end{tabular}
    \end{adjustbox}
\end{table*}

\subsubsection{Dynamic Bayesian Networks}

Dynamic Bayesian Networks (DBNs), applied to trajectory prediction \citep{8957246,Gill2019APF}, capture temporal dependencies and interactions between multiple agents. A typical DBN architecture comprises three layers: the behavior layer, representing input data for each agent, such as past trajectories, velocities, interaction features, and scene context; the hidden layer, modeling latent variables that capture unobserved, intermediate states influencing motion and their probabilistic transitions over time; and the observation layer, which outputs the predicted future trajectories, incorporating the uncertainty of the prediction process. During training, the forward-backward algorithm estimates the most likely sequence of hidden states and their influence on future observations.

On maneuver prediction, \citet{6957713} model driving maneuvers as latent states in a DBN and use probabilistic inference to predict maneuvers, which are subsequently used by maneuver-specific models to generate future trajectories. \citet{Li2019ADB} focus on vehicle maneuver prediction in highway scenarios, leveraging multi-dimensional predictive features, including historical vehicle states, road structures, and traffic interactions, to infer maneuver probabilities. Similarly, \citet{8996494} propose a framework to predict vehicle intentions in highway driving scenarios using DBN. The approach incorporates contextual inputs, such as relative speed, safety distance, and lane information, to infer driver intentions like lane-keeping or lane-changing. The DBN's observation layer captures lane deviation and heading angle, which are influenced by the inferred intentions.

Further, focusing on interaction modelling, \citet{8917039} employ DBN to design a Driver Characteristic and Intention Estimation (DCIE) module. This probabilistic framework estimates driver characteristics, e.g., stable or aggressive behavior, and intentions, by integrating lateral position, velocity, and acceleration. Lastly, \citet{Schulz2018InteractionAwarePB} leverage DBN for predicting future trajectories and interactions, comprising three layers: the route intention layer to determine the desired path of each agent based on the road layout and traffic rules; the maneuver intention layer; and the action layer, capturing continuous actions based on current intentions and interactions.

\subsubsection{Strengths and Limitations of ML Methods}
The discussed ML-based methods offer transparent structure enabling straightforward interpretation and debugging, as the models rely on well-defined mathematical components such as transition probabilities and covariance functions. The methods also provide uncertainty estimates and can perform well with limited training data. However, their limited parameterization restricts their ability to model complex, high-dimensional behaviors common in real-world traffic. DBNs, which can model interactions, struggle to scale with the number of agents or observations and require manual design of model structure and features. Unlike deep learning models, they cannot learn hierarchical representations, reducing their adaptability to new environments or interaction patterns.

\subsection{Deep Learning-based Methods}
\label{sec:dl}

\textbf{\textit{Deep Learning-based}} methods excel in learning motion dynamics from driving data spanning different architectures. Figure \ref{fig:dl_overview} illustrates the main DL models reviewed in this section. Each approach is detailed in a separate subsection, with methods summarized in tables that highlight the key features of the architectures. 

\subsubsection{RNNs and LSTMs}

\begin{table*}[t]
\centering
    \caption{Summary of reviewed DL-based Models relying on RNN, GRU and LSTM (part 2)}
    \label{tab:dl_rnn_lstm_part2}
    \begin{adjustbox}{max width=\textwidth}
    \begin{tabular}{|p{5cm}|c|p{14cm}|}
        \hline
        \textbf{Reference} & \textbf{Name} & \textbf{Description}  \\
        \hline 
         \citet{9021955} (2019) &  SNS-LSTM & Combines three factors using pooling mechanisms: past trajectories patterns, interpersonal interactions, and environment's semantics.  \\ \hline
         \citet{8813801} (2019) &  SSeg-LSTM &  Uses SegNet for semantic segmentation and extracting scene features, which are further integrated into SS-LSTM.  \\ \hline
         \citet{Chandra2019ForecastingTA} (2019) & graph-LSTM & Two-stream framework: first stream - LSTM encoder-decoder to predict spatial trajectories, second stream predicts eigenvectors of dynamic geometric graphs (DGGs) to infer agent behaviors.  \\ \hline
        \citet{zhang2019srlstmstaterefinementlstm} (2019) & SR-LSTM & Includes  intentions of neighbors through  message-passing and refines the states of all pedestrians.  \\ \hline
        \citet{8954462} (2019) & TraPHic &  A hybrid network, which combines LSTM and CNN for predictions, focusing on the varied interactions among different road users. \\ \hline
        \citet{8788650} (2019) & - & Focus on forecasting short and long-term future trajectories, employs ensemble mechanism to enhance robustness. \\ \hline
        \citet{Mangalam2019DisentanglingHD} (2019) & - & 
        Predicts by disentangling global (rigid body) using Quasi-Recurrent Neural Networks (QRNNs) and local (joint-specific) motions using  Encoder-Recurrent-Decoder framework.   \\ \hline
        \citet{8672889} (2019) & ST-LSTM & Integrates spatial interactions within LSTM. \\ \hline
        \citet{Ding2019PredictingVB} (2019) & -
         &  Uses LSTM to predict vehicle’s driving policy to guide a low-level optimization process. 
          \\ \hline        
        \citet{10.1609/aaai.v33i01.33016120} (2019) & TrafficPredict & Instance layer for individual movements and interactions and category level to capture similarities in a group motion.  \\ \hline
        \citet{bhujel2019} (2019)  & - &  Combines RNN and attention mechanism to capture human-human and human-scene interactions.  \\ \hline
        \citet{s20061776} (2020) & AT-LSTM & Inferred a set of variables impacting a crossing intention and trained LSTM with attention mechanism modeling these parameters.  \\ \hline
        \citet{Tao2020DynamicAS} (2020) & - & Two components: Context Module (ICM) for temporal dependencies and Social-aware Context Module (SCM) for spatial interactions. \\ \hline
        \citet{Brito2020SocialVRNNOM} (2020) & Socila-VRNN & Feature extraction module for agent states, social interactions, and environmental context, followed by a probabilistic inference module. \\ \hline \citet{lorenzo2020rnnbasedpedestriancrossingprediction} (2020)  & - & Recurrent model with additional features as pedestrians direction orientation and gaze direction.  \\ \hline
        \citet{9294491} (2020) & - & Includes pedestrian intention and behavior to generate predictions.  \\ \hline
        \citet{Bertugli2021-acvrnn} (2021) & AC-VRNN & Integrates GCN and KL-divergence loss with belief maps to capture interaction and context.  \\ \hline
        \citet{9146979} (2020) & RAI & Contains a temporal attention module to mitigate speed variations and spatial pooling to capture social dynamics and intentions.    \\ \hline
        \citet{8933492}  (2020)  & - &  Classifies driving styles using GMM and uses a shared LSTM layer for each driving style.  \\ \hline
        \citet{9416981}  (2021) & -  &  
        Uses a separate LSTM model for intention prediction and integrates intersection prior trajectories built from analyzing traffic flow.  \\ \hline 
        \citet{9857236}  (2022) & - & Has a goal-estimation module using a U-Net architecture to predict scene-aware trajectories.  \\
        \hline  
       \citet{zhang_lstm_2023}  (2023) & STS-LSTM & Spatial-Temporal-Spectral LSTM uses spatial, temporal, and spectral domain features and contains a transferable model for pedestrian trajectory prediction.    
        \\
        \hline  
        \citet{Xu2023UncoveringTM} (2023) & GC-VRNN & Integrates a Multi-Space GNN for spatial feature extraction and RNN with a Temporal Decay (TD) module for temporal dependencies and handling incomplete data. \\
        \hline  
    \end{tabular}
    \end{adjustbox}
\end{table*}

Sequential models such as RNNs and LSTMs assume that motion evolves through temporally causal dependencies. By maintaining hidden states across time steps, they effectively model dynamic behavior and long-term temporal patterns for trajectory prediction. Focusing on model design, earlier studies primarily utilized a single recurrent model to map past trajectories to future paths \citep{8569641, Liu2020SpatiotemporalRR, Kotseruba2020DoTW}. However, more recent works employ multiple groups of recurrent models to address different aspects of motion \citep{8317848, 10.1109/ITSC.2018.8569595, 8672889}, as well as stacked LSTMs and convolutional LSTMs. \citet{8672889} proposed a model, where one group of LSTMs models individual vehicle motions, while another group captures interactions between agents. The architectural approach for stacked LSTMs involves hierarchically organizing recurrent networks into layers, as seen in \citep{8848853, 10.1109/IVS.2018.8500493}. Tables \ref{tab:dl_rnn_lstm_part1} and \ref{tab:dl_rnn_lstm_part2} provide a summary of architectures in the literature that rely on recurrent models.

%%%%%%%%%%%%%%%%%%%%%%%%%%%%%%%%%%%%%%%%%%%%%%%%%%%%%%%
\citet{Varshneya2017HumanTP} combined LSTM with an attention mechanism to simultaneously preserve dynamic context and learn patterns from the spatial coordinates of agents. To capture spatial context, the model utilizes a Spatially Static Context Network (SSCN). \citet{Ding2019OnlineVT} introduced a prediction system with a dual-level design. The first level is a policy anticipation network, which employs an RNN-based encoder to analyze past trajectories and predict drivers' intentions. These predicted intentions are passed to the second level, where a context reasoning module incorporates detailed contextual data. The framework uses non-linear optimization to generate continuous predictions based on structured contextual information. For multimodal predictions, \citet{Zyner2018NaturalisticDI} utilize an encoder-decoder three-layer LSTM to predict parameters of a weighted Gaussian Mixture Model (GMM) to generate trajectories. \citet{Park2018SequencetoSequencePO} also use encoder-decoder LSTMs, focusing on predicting the probability of occupancy on a grid map. A beam search is then employed to select the $k$ most probable future trajectories from the candidates. Another approach is presented in \citep{10.1109/IVS.2018.8500493}, where six distinct decoder LSTMs are used, each corresponding to a specific highway driving maneuver. An encoder LSTM processes past trajectories, and the hidden state of each decoder LSTM is initialized with a combination of the encoder LSTM's last hidden state and a one-hot vector representing the maneuver specific to that decoder. These decoder LSTMs predict parameters for a maneuver-conditioned bi-variate Gaussian distribution of future vehicle locations.

\begin{table*}[t]
    \centering
    \caption{Summary of reviewed DL-based Models relying on CNNs}
    \label{tab:dl_cnn}
    \begin{adjustbox}{max width=\textwidth}
    \begin{tabular}{|p{4cm}|c|p{15cm}|}
        \hline
        \textbf{Reference} & \textbf{Name} & \textbf{Description}  \\
        \hline 
        \citet{7508877} (2016)  & Behavior CNN &  Integrates formation of stationary crowd groups and analyzes their interaction with moving pedestrians to generate predictions. \\ \hline
        \citet{Varshneya2017HumanTP} (2017) & SSCN &  Models static artifacts to improve predictions and uses a soft attention mechanism to learn motion patterns of humans using different navigational modes. \\ \hline
        \citet{Lee2017ConvolutionNN} (2017) & - &  Uses CNN and predictive controller to predict a lane-change intention from frontal camera image.
        \\ \hline
        \citet{8578474} (2018) & - & Uses 3D convolutions across space and time on LiDAR data for predictions.
        \\ \hline
        \citet{pmlr-v87-casas18a} (2018) & - &  One-stage detector and forecaster from 3D point cloud and dynamic environmental maps. \\ \hline
        \citet{8460203} (2018) & RMDN & Destination network uses CNN to predict a mixture of possible destinations, topology network uses FCN to predict planning topology for generating predictions. \\ \hline
        \citet{nikhil2018convolutionalneuralnetworktrajectory} (2019) & CNN-TP & CNN framework with multiple convolution layers run in parallel. \\ \hline
        \citet{Jain2019DiscreteRF} (2019) & DRF-Net & CNN for long-term human motion prediction, by capturing a wide range of potential future movements. \\ \hline
        \citet{10.1109/ICRA.2019.8793868} (2019) & - & Uses CNN to encode contextual surroundings of each agent from raster images. \\ \hline
        \citet{8668475}  (2019) & - &  Two-stream architecture: spatial stream at frame level and temporal stream across frames. Applies TVNet model to generate optical-flow-like features for each frame-pair. \\ \hline
        \citet{9304846} (2020) & - & Uses potential field method to model navigation of each agent and passes potential field representation to neural network as input. \\ \hline
        \citet{9255162} (2020)  & - & Interaction-Aware Trajectory Prediction of Connected Vehicles using CNN-LSTM Networks. \\ \hline
        \citet{9156890} (2020) & TPNet & Employs a two-stage process: first, generates trajectory proposals using polynomial curve fitting; second, refines them, incorporating knowledge from semantic maps and prior physical constraints. \\ \hline
        \citet{9157523} (2020) & CoverNet &  Treats trajectory prediction problem as a classification task over a curated set of possible trajectories. \\ \hline
        \citet{chou2020predictingmotionvulnerableroad} (2020)  & - & Transforms HD maps and agents' surrounding context into BEV and passes to CNN as the input. \\ \hline
        \citet{9341327} (2020)  & - & CNN for raster features and TCN for states and motion history.\\ \hline
        \citet{Marchetti2020MANTRAMA} (2020)  & MANTRA &  Uses LSTM for encoding, Memory Augmented NN to decode future trajectories and CNN for refining. \\ \hline 
        \citet{Gilles2021HOMEHO} (2021) & HOME &  Uses CNN to encode local context from rasterized image of the agent's environment. \\ \hline
        \citet{ye2021tpcntemporalpointcloud}  (2021)& TPCN & Contains dual-representation spatial learning module and dynamic temporal learning modules for spatial and temporal features.  \\ \hline 
        \citet{WANG2021104110}  (2021)& MI-CNN &   
        Uses multi-input: past trajectories, pose, depth maps, and 2D-3D size data to generate predictions. \\ \hline
        \citet{9575958}  (2021) & Social-IWSTCNN & Social Interaction-Weighted Spatio-Temporal CNN with a Social Interaction Extractor for learning social interaction weights and spatial features. \\ \hline
        \citet{zamboni2021pedestriantrajectorypredictionconvolutional} (2022) & Conv2D & \begin{tabular}{@{}c@{}} Uses CNN  with data augmentation. \end{tabular} \\ \hline
        \citet{CHEN2024120455} (2024) & DSTCNN & Deformable Spatial-Temporal CNN models spatial and temporal interactions and uses adaptive kernels.
        \\  
        \hline
    \end{tabular}
    \end{adjustbox}
\end{table*}

\subsubsection{CNNs}

\begin{table*}[t]
    \centering
    \caption{Summary of reviewed DL-based Models relying on Autoencoders}
    \label{tab:dl_autoencoders}
    \begin{adjustbox}{max width=\textwidth}
    \begin{tabular}{|c|l|p{15cm}|}
        \hline 
        \textbf{Type} & \textbf{Reference}  & \textbf{Description}  \\
        \hline 
         \multirow{8}{*}{\begin{tabular}{@{}c@{}} Encoder-\\ Decoder \end{tabular}} & \citet{Park2018SequencetoSequencePO} (2018) & LSTM-based encoder-decoder architecture.  \\ \cline{2-3}
        & \citet{10.1109/IVS.2018.8500493} (2018) & LSTM-based encoder-decoder  with maneuver classification branch. \\  \cline{2-3}
        & \citet{Hong2019RulesOT} (2019) & Encodes scene context using a unified spatial grid representation. \\  \cline{2-3}
        & \citet{Chen_2022} (2022) & End-to-end CNN-LSTM encoder-decoder with an attention mechanism, leveraging semantic segmentation to extract obstacle information. \\  \cline{2-3}
        & \citet{HUI2022126869} (2022) & DNN-based Encoder-Decoder \\  \cline{2-3}
        & \citet{10656745} (2024) &  Masked Autoencoder, leveraging masked reconstruction of input features to adapt to distribution shifts.   \\   \cline{2-3}
        & \citet{Lv2024} (2024) &  Learning Autoencoder Diffuser modelling group relationships for pedestrian trajectory prediction. \\  
        \hline 
        \multirow{21}{*}{\begin{tabular}{@{}c@{}} Variational \\ Autoencoder \end{tabular}} & \citet{Lee2017DESIREDF} (2017) & CVAE-based RNN encoder-decoder for samples generation and ranking the most likely samples. \\ \cline{2-3}
        & \citet{hu2018framework} (2018) & Two modules: module predicting intentions and module for predicting motion within the interacting scene.  \\ \cline{2-3}
        & \citet{8967708} (2019) & CVAE-based interaction module.  \\ \cline{2-3}
        & \citet{Feng2019VehicleTP}  (2019)  &  Uses GRU-based encoder-decoder and integrates intention prediction module.   \\ \cline{2-3}
        & \citet{10.1109/ITSC.2019.8917105}  (2019)  & Samples joint trajectories of interacting vehicles, converts joint distribution to conditional + MPC. \\ \cline{2-3}
        & \citet{Hu2019MultimodalPP}  (2019) & Conditioned on agent's intention and historical scene information. \\ \cline{2-3}
        & \citet{9294482} (2020) &  Uses Wasserstein distance for loss and generates predictions using a GMM. \\ \cline{2-3}
        &  \citet{Ivanovic2020MultimodalDG} (2020)  &  Encodes interaction history and target agent intentions. \\ \cline{2-3}
        & \citet{zhang2020multimodal}  (2020) &  
         Stacked auto-encoders to encode state; LSTM encoder for past trajectory and CNN pooling for interactions.  \\ \cline{2-3}
        & \citet{Cheng2020ExploringDC} (2020)  & Integrates a self-attention with a two-stream encoder to encode observed trajectories and agent interactions. \\ \cline{2-3}
        & \citet{9880379} (2022)  & A cascade of Conditional VAEs: macro-stage predicts high-level goals and micro-stage refines predictions into sequential paths. \\ \cline{2-3}
        & \citet{10.1007/978-3-031-19772-7_30}  (2022) & Uses a backward RNN to capture navigation patterns and attention mechanism for social interaction. \\ \cline{2-3}
        & \citet{stgm_2022}   (2022) & Employs CVAE with a designed modal-wise sampling strategy to improve prediction stability. \\ \cline{2-3}
        & \citet{9880415}  (2022) & Uses graph partitioning to form cliques of interacting agents to reduce latent space complexity. \\ \cline{2-3}
        & \citet{10.1016/j.patcog.2022.109030}  (2023) &  Cascaded CVAE module using past trajectories and earlier predicted locations with a socially-aware regression module for refinement. \\ \cline{2-3}
        & \citet{10.1007/978-981-99-7019-3_2} (2023) &  VAE with a diffusion prior. \\ \cline{2-3}
        & \citet{xu2023contextawaretimewisevaesrealtime} (2023) & Uses a dual-attention mechanism to encode environmental and social contexts, generating map-compliant and socially-aware predictions. \\ \cline{2-3}
       & \citet{Xiang2024SocialCVAEPP} (2024) & Uses a socially explainable energy map to model social interactions.  \\ 
        \hline 
    \end{tabular}
    \end{adjustbox}
\end{table*}

Convolution Neural Networks for trajectory prediction are categorized into two groups in the literature: those using image-based input and Temporal Convolution Networks (TCNs). Image-based inputs can come from various perspectives, including BEV, vehicle-mounted cameras, or other sensor modalities. BEV images or aerial photographs offer a top-down perspective that effectively captures spatial relationships and road structures. CNN-based models \citep{9157523, 10.1109/ICRA.2019.8793868, chou2020predictingmotionvulnerableroad} primarily focus on the feature extraction stage to capture road geometries, traffic agent positions, and environmental context. The second category involves Temporal Convolution Networks (TCNs), which apply convolution layers directly to trajectory sequences to model temporal dependencies. Unlike RNNs, TCNs operate in parallel over the entire sequence, incorporating causal convolutions to preserve temporal order and dilated convolutions to extend the receptive field, capturing long-term dependencies. TCNs are designed to maintain stable gradients, avoid information leakage, and achieve computational efficiency for long sequences. One example is the framework proposed by  \citet{nikhil2018convolutionalneuralnetworktrajectory}, which preserves spatio-temporal consistency through multiple convolution layers while leveraging parallelism for efficiency. Table \ref{tab:dl_cnn} provides a summary of CNN-based architectures for trajectory prediction.

\citet{Marchetti2020MANTRAMA} introduced a multimodal prediction model utilizing a Memory Augmented Neural Network. This model encodes previously observed trajectories as keys to access probable future trajectory encodings from memory, which are then decoded into multimodal predictions conditioned on observed data. The surrounding context is processed by a CNN and passed through a refinement module to enhance the decoded predictions. In a separate study, \citet{Gilles2021HOMEHO} developed a prediction model that takes a rasterized image of the agent's environment as input and generates a heatmap indicating potential future positions with associated probabilities. A standard CNN encodes the local context from a rasterized HD map, while past trajectories are encoded using a 1D convolutional layer followed by a recurrent layer. Interactions are modeled using an attention mechanism. \citet{9341327} designed a model for trajectory prediction from rasterized images, employing a CNN to extract features that capture both static and dynamic elements. These features are merged with current and past states processed through a TCN. The combined data is then mapped to future states through a series of linear layers. Lastly, \citet{ye2021tpcntemporalpointcloud} proposed a Temporal Point Cloud Network (TPCN) for trajectory prediction, featuring two main blocks: a dual-representation spatial learning module and a dynamic temporal learning module. The spatial module consists of two sub-modules: one for learning point-wise features to capture geometric information and interactions, and another for learning voxel features to extract semantic context. These modules are trained jointly.

A series of studies have combined LSTM’s ability to capture temporal dependencies with CNN’s spatial feature extraction to address spatio-temporal patterns in an integrated manner. Convolutional LSTM (ConvLSTM) models have emerged as a unified solution, incorporating convolutional operations directly within recurrent units to enable simultaneous spatial and temporal processing. ConvLSTM-based models \citep{8954462, 8966366, 9196807, song2019, Ridel2019SceneCT, Liang2019TheGO, 9093426, 10.1007/978-3-030-58583-9_28} are designed to handle tasks requiring spatio-temporal coherence. \citet{8954462} proposed TraPHic, which uses convolutional operations for localized spatial interactions and a horizon-based weighting mechanism to rank interactions by proximity. \citet{8966366} utilized ConvLSTM to model spatio-temporal motion patterns and inter-vehicle interactions, processing risk maps based on metrics like Time-to-Collision (TTC). \citet{9196807} introduced a three-module framework comprising interaction learning for surrounding vehicle effects, temporal learning for past movements, and motion learning for future position predictions.

\begin{table*}[h]
    \centering
    \caption{Summary of reviewed DL-based Models relying on GANs}
    \label{tab:dl_gans}
    \begin{adjustbox}{max width=\textwidth}
    \begin{tabular}{|p{4cm}|c|p{15cm}|}
        \hline
        \textbf{Reference} & \textbf{Name} & \textbf{Description}  \\
        \hline 
        \citet{DBLP:journals/corr/abs-1812-07667} (2018)
        & GD-GAN & GAN pipeline for trajectory forecasting and group detection. $G$ generates trajectories conditioned on the local neighborhood context.  \\ \hline
        \citet{Gupta2018SocialGS} (2018) & Social GAN & $G$ consists of LSTM encoder, social pooling and LSTM decoder. $D$ consists of LSTM encoder and MLP for classification. \\ \hline
        \citet{Amirian2019SocialWL}  (2019) & 
        Social Ways & Uses attention social pooling for modelling interactions.  \\ \hline
        \citet{8916927}   (2019) &  Interaction-GAN & Embeds interactions as social context; tested on dataset with inconsistent lane behavior at intersections. \\ \hline      
        \citet{Li2019InteractionawareMT}   (2019) & Prob-GAN  & Both $G$ and $D$ are combination of one recurrent layer and several fully connected layers. \\ \hline
        \citet{8953374}   (2019)  & SoPhie & LSTM-based GAN takes as the input features from attention module to generate realistic future paths complying with learnt constraints.  \\ \hline
        \citet{DBLP:conf/cvpr/ZhaoXMCBZ0W19}   (2019) & MATF & $G$ is conditioned on the static scene context and agents past trajectory.  \\ \hline
        \citet{Li2019ConditionalGN} (2019) & CGNS & Architecture consists of a deep feature extractor, conditional encoder to learn latent space, $G$ to sample trajectory hypotheses from the learnt distribution and recurrent $D$. \\ \hline
        \citet{10.1145/3394171.3413602}   (2020) & AEE-GAN &  Consists of three blocks: Feature Encoder, Visual and Social Attention, and LSTM-based GAN. The attended ecology visual and social embedding is passed to $G$. \\ \hline
        \citet{9243464} (2020)  & Vehicle-GAN & $G$ consists of LSTM-based encoder and decoder, and pooling layer to model interactions. $D$ employs LSTM-based encoder and MLP. \\ \hline
        \citet{Yang2020TPPOAN} (2020)  & TPPO & Includes latent variable predictor encoding environmental context and motion patterns, and Social Attention Pooling module.  \\ \hline
        \citet{Huang2021STIGANMP} (2021) & STI-GAN & Similar architecture to Social GAN. Employs graph attention model instead of pooling layer. \\ \hline
        %%%%%%%%%%%%%%%%%%%%%%%%%%%%%%%%%%%%%%%%%%%
        \citet{9564674} (2021) & DSA-GAN &  Contains a driving style recognition network. $G$ generates trajectories conditioned on driving style. \\ \hline       
        \citet{Dendorfer2021MGGANAM} (2021) & MG-GAN & Multi-Generator GAN, each generator specializes in a distinct trajectory mode + Path Mode Network (PM-Net) to predict a categorical probability distribution over these generators based on the scene context. \\ \hline 
        \citet{9151374} (2022) & TS-GAN &   $G$ consists of encoder and LSTM-based decoder. Encoder includes Social Recurrent module and Social Convolution block.  \\ \hline
        \citet{9366373} (2021) & STL-GAN &   Employs signal temporal logic (STL) to encode rules.  Rule features extracted from STL syntax tree are incorporated into $D$. \\ \hline
        \citet{Fang2022AttenGANPT} (2022) & Atten-GAN & $G$ contains an attention pooling module to allocate influence weights among pedestrians and extract interaction information.
        \\ \hline
        \citet{10.1609/aaai.v37i4.25557} (2023) & Two-stage GAN & $G$ uses LSTM and graph attention networks (GAT) to estimate cost of transitioning between road network nodes and predict next movements, ensuring spatial and temporal coherence.
        \\ \hline
        \citet{10197467} (2024) & SAGAN & Social self-attention mechanism to refine interaction information by focusing on most relevant agents.   
        \\ \hline
    \end{tabular}
    \end{adjustbox}
\end{table*}

\subsubsection{Autoencoders}
\label{sec:auutoencoders}

In trajectory prediction, autoencoders encode past trajectories into compact latent representations, which are then decoded into future positions. These models excel at capturing complex temporal dependencies and can function as standalone predictors or feature extractors integrated into larger architectures. Variational Autoencoders (VAEs) extend this approach by modeling uncertainty, generating a latent distribution instead of fixed representations. By sampling from this distribution, VAEs generate diverse and multimodal future trajectories, modelling the inherent uncertainty in trajectory forecasting tasks. Table \ref{tab:dl_autoencoders} provides a summary of autoencoder architectures available in the literature, grouped by deterministic or stochastic type.

\citet{10.1109/IVS.2018.8500493} introduced an LSTM-based encoder-decoder architecture with a maneuver classification branch. The encoder processes the historical trajectories of the target vehicle and its neighbors, condensing this information into a context vector. This vector, combined with maneuver encodings, is further utilized by the decoder to produce maneuver-specific probabilistic distributions of future trajectories. \citet{Hong2019RulesOT} designed a model to map a world representation to the future trajectories of a single target agent. For the surrounding world context, the authors created a 4D tensor comprising the target agent's state, its surrounding agents' states, dynamic context, and road network. This tensor is encoded into a latent representation, which the decoder uses to model future state distributions at specified time offsets, enhanced by temporal convolutions.

Among VAEs, \citet{8967708} proposed a CVAE-based interaction module where past and future trajectories are encoded to generate a stochastic latent variable \( z \). The ego vehicle’s state and its neighbors’ states are used to infer rule information, integrating encoded states with \( z \) to generate future trajectories. \citet{10.1109/ITSC.2019.8917105} proposed a generic prediction framework to address cases of irrational human behaviour. First, the model samples joint future trajectories of interacting vehicles based on past trajectories. Next, the joint distribution is converted to a conditional distribution, and Model Predictive Control (MPC) is employed to generate the most probable trajectories. Finally, the weights of generated trajectories are re-weighted to align with rational behaviours. Lastly, \citet{zhang2020multimodal} proposed a vehicle-descriptor-based LSTM model to capture interactions between the global context and agents' local position resolutions. The model encodes the agent's multimodal state information using stacked autoencoders and employs an LSTM-based decoder to generate trajectory distributions based on estimated maneuvers.

\begin{table*}[t]
    \centering
    \caption{Summary of reviewed DL-based Models relying on GNN (part 1)}
    \label{tab:dl_gnns}
    \begin{adjustbox}{max width=1.0\textwidth}
    \begin{tabular}{|p{4cm}|c|p{15cm}|}
        \hline
        \textbf{Reference} & \textbf{Name} & \textbf{Description}  \\
        \hline 
        \citet{Lee2019JointIA} (2019)
        & - &  RNN Encoder, GNN Layers to process pairwise features, and to refine node embeddings through edge-type-specific message passing, RNN Decoder. \\ \hline
        \citet{Sun2019StochasticPO} (2019) & Graph-VRNN & Each agent is a node with a VRNN to capture dynamic uncertainties, and hidden states of these VRNNs interact through a graph interaction network. \\ \hline
        \citet{Yang2020ANG} (2020) & - & Uses LSTM for motion encoding, social graph attention with obstacle avoidance for interaction modeling, and a pseudo oracle predictor to handle uncertainties via informative latent variables. \\ \hline
        \citet{Wang2020GraphTCNSI} (2020) & GraphTCN & Combines edge-feature GATs (EFGAT) for spatial interactions and TCNs for temporal dynamics. \\ \hline
        \citet{Casas2020ImplicitLV} (2020) & - & Models scenes as fully connected graphs, using GNN for inter-agent dependencies, with latent variables encoding stochastic scene dynamics. \\ \hline
        \citet{Zhang_Chang_Meng_Xiang_Pan_2020} (2020) & ST-LGSL & Constructs spatiotemporal graphs using a latent graph structure, MLP and kNN; employs diffusion graph convolutions and gated temporal convolutions. \\ \hline
        \citet{Zhao2020GISNetGraphBasedIS} (2020) & GISNet & Encodes each vehicle's past trajectory for interaction-aware prediction. \\ \hline
        \citet{10.5555/3495724.3497384} (2020) & EvolveGraph & Static graph learning for interactions and dynamic graph refinement to model temporal dependencies. \\ \hline
        \citet{10.1007/978-3-030-58523-5_40} (2020) & Trajectron++ &  Uses CNN to encode semantics from environment, LSTMs and attention for agent histories and interactions, and CVAE to generate predictions based on GNN-encoded features.\\ \hline
        \citet{9506209} (2021) &  STR-GGRNN & Uses  Gated Graph Recurrent Neighbourhood Network (GGRNN) to construct dynamic graphs with Non-negative Matrix Factorization (NMF) to adapt neighbourhood structures.\\ \hline
        \citet{Zhang2021TrajectoryPW} (2021) & - & Dual-scale context fusion with layers for geometric and topological features. \\ \hline
        \citet{9561461} (2021) & SpecTGNN & Captures inter-agent correlations and temporal dependencies in frequency and time domains, using spectral graph and temporal gated convolutions. \\ \hline
        \citet{ZHOU2021298} (2021) & AST-GNN & using spatial GNN for interaction modeling and temporal GNN for motion pattern extraction, combined with a time-extrapolator CNN for predictions. \\ \hline
        \citet{Chen2021SCSGAA}(2021)  & SCSG & Uses multi-head attention for social interactions and star graph decoder. \\ \hline
        \citet{Fang2022HeterogeneousTF} (2022) & - & Uses Heterogeneous Risk Graph (HRG) to capture collision risks; and Hierarchical Scene Graph (HSG) for agent-environment constraints. \\ \hline
        \citet{9468360} (2022) & - &  Combines two GNN modules: one for recognizing interactive events and another for predictions, integrated with LSTM for multi-step predictions. \\  \hline 
        \citet{9811632} (2022) & MESRNN & Leverages meta-path features in spatio-temporal graphs for interactions, improves long-term predictions and social compliance in dense crowds. \\ \hline
        \citet{9879068} (2022) & GroupNet & Multiscale hypergraph neural network: for pair-wise interactions across different groups sizes - multiscale hypergraph topology inference; for group-wise interactions - hypergraph neural message passing.  \\ \hline
        \citet{9880042} (2022)& TGNN & Uses a domain-invariant GNN and attention module for features transfer. \\ \hline
        \citet{10.1109/ICRA46639.2022.9812253}(2022) & GOHOME & Represents HD map as graph and uses graph convolutional layers capture spatial and agent-lane interactions. \\ \hline
        \citet{10.1007/978-3-031-20047-2_16} (2022) & - & Assigns pedestrians to behaviour groups, forming intra- and inter-group interaction graphs; uses group pooling operations to capture collective dynamics.\\ \hline
        \citet{ZHU2023109772} (2023) & Tri-HGNN & Triple Policies: extrinsic-level policy uses GNN for interactions, intrinsic-level policy uses GCN to infer intentions, and basic-level policy to integrate features from both policies using TCN.\\  \hline 
        \citet{Westny2023MTPGOGP} (2023) & MTP-GO & Uses Graph-GRU units to encode interactions and neural ODEs to capture dynamic motion constraints. \\ \hline
        \citet{10205349} (2023) & EqMotion & Includes a reasoning module to infer interaction graphs and employs equivariant geometric feature learning. \\ \hline
        \citet{Westny2023EvaluationOD} (2023) & - & Combines MTP-GO with differentially-constrained motion models. \\ \hline
        \citet{MO2024103748} (2024) & - & Uses Heterogeneous Graph Social (HGS) pooling to model vehicles and infrastructure in a unified graph for vehicle-infrastructure interactions.  \\ \hline
        \citet{10656710} (2024) & - &  Uses a collision-aware kernel function to construct adjacency matrices and higher-order graph convolutions to model indirect social influences. \\ \hline
        \end{tabular}
    \end{adjustbox}
\end{table*}

\begin{table*}[t]
    \centering
    \caption{Summary of reviewed DL-based Models relying on GNN (part 2, GCNs)}
    \label{tab:dl_gcns}
    \begin{adjustbox}{max width=1.0\textwidth}
    \begin{tabular}{|p{4cm}|c|p{15cm}|}
        \hline
        \textbf{Reference} & \textbf{Name} & \textbf{Description}  \\
        \hline 
         \citet{8917228} (2019) & GRIP & Alternates between temporal convolutional layers for motion feature extraction and graph operation layers for feature propagation. \\ \hline
         \citet{Li2019GRIPEG} (2019) & GRIP++ & Refines interaction modeling with dynamic edge weights and incorporating scene context. \\ \hline
         \citet{Liu2020SpatiotemporalRR} (2020) & - & Constructs pedestrian-centric spatiotemporal graphs, leveraging GCNs to aggregate contextual and temporal features, with GRU-based temporal connections integrating node-level features over time.\\ \hline
        \citet{liang2020learning} (2020) & LaneGCN & Four blocks: ActorNet, encoding past trajectories, MapNet, encoding lane graph features; FusionNet, modeling interactions; and the Prediction Header, generating multimodal predictions. \\ \hline
        \citet{9156583} (2020)
        & Social-STGCNN & Constructs social behaviour graphs with GCNs for group-based interactions, recursive updates, and an LSTM decoder for trajectory prediction.  \\ \hline
        \citet{9156898} (2020) & - & Uses GCNs to model group social interactions recursively, combining with BiLSTM-encoded individual features, uses LSTM-based decoder.  \\ \hline 
        \citet{Xue2020SceneGS} (2020) & - &  Models interactions using a star graph combined with scene features extracted by a DeepLabv3+ encoder, employs a gating mechanism, a VAE for multimodality, and LSTM encoders/decoders. \\ \hline
        \citet{10.1109/ICRA48506.2021.9560908} (2021) & AVGCN &  Integrates learned attention weights from gaze data and visual field constraints to infer neighbor importance. \\ \hline
        \citet{9577379}  (2021) & SGCN & Uses sparse graphs with adaptive adjacency matrices, capturing spatial \& temporal dynamics via self-attention and asymmetric convolutions.  \\ \hline
        \citet{Bae2021DisentangledMG}  (2021) & - & Constructs multi-relational weighted graphs and employs disentangled multi-scale aggregation and global temporal aggregation. \\ \hline
        \citet{9501325}  (2021) & STUGCN & Usese kernel-based adjacency matrices to capture social interactions and Time-Extrapolator CNN to decode predictions from these embeddings.  \\ \hline
        \citet{Zhou2021GrouptronDM}  (2021) & GroupTron & A multi-scale graph neural network - models interactions at individual with LSTMs, group with STGCNs, and scene levels.  \\ \hline 
        \citet{Lv2021SSAGCNSS}  (2021) & SSAGCN & Employs social soft attention mechanism for interactions and a sequential scene attention mechanism; combined features are processed by GCNs and TCNs to generate predictions. \\ \hline 
        \citet{Sheng2021GraphBasedSC}  (2021) & - & Models spatial interactions with a GNN applied to proximity-based weighted graphs and temporal features using a CNN-based Temporal Dependency Extractor (TDE). \\ \hline
        \citet{Schmidt2022CRATPredVT}  (2022) & CRAT-Pred &  Combines crystal graph convolutional neural networks (CGCNNs) and multi-head self-attention to model social interactions among vehicles. \\ \hline 
        \citet{10.1155/2022/4192367} (2022)  & MDST-DGCN & Motion encoder for individual motion features, multilevel dynamic spatiotemporal directed graph encoder (MDST-DGEN) for multilevel social interaction modelling and fusion, motion decoder for generating predictions. \\ \hline 
        \citet{10.1109/TITS.2022.3142248}  (2022) & - & Employs directed graph topologies (view, direction, and rate graphs) to model asymmetric interactions between agents. \\ \hline 
        \citet{10.1109/TITS.2022.3173944} (2023) & - & Combines individual trajectories via a TCN, scene context using a 2D CNN with temporal attention, and interaction patterns with a Spatio-Temporal Dynamic GCN.  \\ \hline 
        \citet{Lian2023} (2023)  & PTP-STGCN & Uses a spatial GCN with a crowd interaction attention function to capture spatial interactions and a temporal transformer to extract motion features.\\ \hline 
        \citet{Xu2023MVHGNMA} (2023)  & MVHGN & Combines multi-view logical correlations and adaptive spatial topology; logical-physical features are extracted using graph convolution modules and regional clustering. \\ \hline 
        \citet{10.1007/s11227-023-05850-8} (2023)  & STIGCN & Uses spatial and temporal graphs to model correlation between social interactions and pedestrian movement.  \\ \hline 
        \citet{10309163} (2024)  & - & Combines directed graphs with multi-head self-attention and graph convolutions to model spatial interactions.\\ \hline
        \citet{Du_2024} (2024) & SFEM-GCN & Combines semantic, position, and velocity-based social force graphs into a mixed spatio-temporal graph.  \\ \hline
        \citet{Liao2024MFTrajMB} (2024)  & MFTraj & Four blocks: behaviour-aware module with dynamic geometric graphs, position-aware with LSTMs, interaction-aware module with adaptive GCN, and a residual decoder.
 \\ \hline
    \end{tabular}
    \end{adjustbox}
\end{table*}

\begin{table*}[t]
    \centering
    \caption{Summary of reviewed DL-based Models relying on GNN (part 3, GATs)}
    \label{tab:dl_gats}
    \begin{adjustbox}{max width=\textwidth}
    \begin{tabular}{|p{4cm}|c|p{15cm}|}
        \hline
        \textbf{Reference} & \textbf{Name} & \textbf{Description}  \\
        \hline 
        \citet{9010834} (2019) & STGAT &  Uses graph attention mechanism for spatial interactions and an LSTM to encode temporal correlations.  \\ \hline
         \citet{e294141389194a54a05536938fcdd509} (2019) &
        Social-BiGAT &  Utilizes GAT for interactions, MLP and LSTM to encode pedestrians' past trajectories, CNN as scene-level feature extractor and  Bicycle-GAN to generate predictions.  \\ \hline
        \citet{Zheng2019GMANAG} (2020) & GMAN & Employs an encoder-decoder architecture with Spatial-Temporal Attention blocks and gated fusion to capture dynamic spatio-temporal correlations. \\ \hline
        \citet{Li2020SocialWaGDATIT} (2020)  & Social-WaGDAT & Uses spatio-temporal graphs with double-attention mechanisms for interactions, and context- and a kinematic constraint-aware decoder. \\ \hline
        \citet{Eiffert2020ProbabilisticCG} (2020)  & - &  Uses RNN-based encoder-decoder with a Mixture Density Network (MDN) and Graph Vehicle-Pedestrian Attention Network (GVAT) to model heterogeneous interactions. \\ \hline
        \citet{Carrasco2021SCOUTSA} (2021) & SCOUT & Uses three attention mechanism: self-attention: for temporal dependencies, cross-attention interactions for Models interactions attending relevant features from neighbouring agents and contextual attention for scene context.  \\ \hline
        \citet{9412114} (2021) & DAG-Net & Incorporates both past interactions and future goals using two attentive graph neural networks. \\ \hline 
        \citet{mo_HEAT_2024} (2022) & HEAT & Combines agent-specific dynamics encoders, a directed edge-featured graph for interaction modeling, and an adaptive map selector for road context. \\ \hline 
        \citet{10.1016/j.robot.2021.103931}  (2022) & PTPGC &  Uses a graph attention network to capture spatial interactions, and applies TCN to encode individual dynamics. \\ \hline 
        \citet{9491972}  (2022) & STG-DAT & Uses dual-attention mechanism to capture topological and temporal relationships and a kinematic constraint layer in the decoder.\\ \hline
        \citet{10.1016/j.neucom.2022.03.051}  (2022) & EvoSTGAT & Models recursive interactions across contiguous time points using a dynamic attention mechanism, evolving social influences frame by frame. \\ \hline 
        \citet{Zhou2022GCHGATPT}  (2022) & GCHGAT &  Model intragroup, outgroup, and intergroup interactions, combining a vanilla generative adversarial network for rough predictions with a state-refinement module to incorporate interactions. \\ \hline 
        \citet{CHENG2023163} (2023) & GATraj & Uses attention mechanisms for spatial-temporal dynamics, GCN for interactions, and a Laplacian mixture decoder for generating predictions. \\ \hline
        \citet{10.5555/3540261.3542337} (2024) & GRIN & Encodes inter-agent latent codes to model social relations and intra-agent latent codes to represent individual intentions, leveraging these in a graph attention network for continuous pair-wise interaction learning. \\ \hline 
        \citet{9723649} (2023) & AI-TP &  Integrates GAT for interactions. \\ \hline 
    \end{tabular}
    \end{adjustbox}
\end{table*}

\subsubsection{GANs}
\label{sec:gans}

Generative Adversarial Networks (GANs) also found their application in trajectory prediction, leveraging their ability to model complex, multimodal data distributions. In this context, the generator (\( G \)) predicts plausible future trajectories conditioned on past trajectories, while the discriminator (\( D \)) evaluates these predictions against real trajectory data to guide the generator's learning. This adversarial training setup enables GANs to generate diverse and realistic trajectories that account for the inherent uncertainty in future motion. A key advantage of GAN-based models is their capacity to address the multimodal nature of trajectory prediction, where multiple plausible futures exist for a given past trajectory. This is achieved by conditioning \( G \) on contextual information such as social interactions, environmental constraints, or road semantics, allowing it to adapt predictions to dynamic and complex scenarios. Table \ref{tab:dl_gans} provides a summary of GAN architectures in the literature, detailing the design of \( G \) and \( D \).

\citet{Gupta2018SocialGS} presented the Social GAN model, which combines sequence-to-sequence predictions with a pooling mechanism for the generator and a recurrent discriminator. The generator takes as input the past trajectories of all agents in the scene and employs an LSTM encoder to capture motion history. After encoding, the social pooling mechanism aggregates the encoded trajectories to capture interactions between agents. The generated pooled vectors reflect the influence of nearby agents on an individual's trajectory. These pooled vectors are then combined with the encoded past trajectories and passed to the decoder, while the discriminator classifies plausible trajectories. \citet{Amirian2019SocialWL} proposed the Social Ways model, which aims to generate more diverse samples and avoid mode collapse. To model interactions, the model uses LSTM to encode the influence of other agents and applies an attention weighting process to generate interaction weights. \citet{8953374} focused on designing a model that generates trajectories plausible under physical and social constraints. The model consists of three modules: a feature extractor with CNN and LSTM encoders, a physical and social attention module, and a GAN network to generate trajectories. While the physical attention module learns spatial constraints, the social attention layer captures interactions between agents. Finally, \citet{DBLP:conf/cvpr/ZhaoXMCBZ0W19} introduced a multi-agent tensor fusion model, which employs a conditional generator for predictions conditioned on the static scene context and agents' past trajectories. Both the generator and discriminator utilize CNN and LSTM-based encoders and an LSTM-based decoder.

\subsubsection{GNNs}

Graph Neural Networks (GNNs) are increasingly applied in trajectory prediction for their ability to model complex spatial and temporal interactions. GNNs operate on graphs where nodes represent traffic agents, and edges encode pairwise relationships, such as proximity or relative motion. GNNs are scene-centered models, and there are two approaches for constructing graphs: static and dynamic graphs, depending on how node and edge updates are handled over time. In dynamics graphs approach, a new graph \( G_t \) is constructed at each time step from sensor data \citep{10.1007/978-3-030-58523-5_40,9156583,Gao2020VectorNetEH}. Since the number of agents varies across frames, the graph structure and adjacency matrix change dynamically, allowing for a variable number of nodes per time step. If an agent enters or leaves the scene, the graph updates accordingly without requiring a fixed structure. Static graphs approach maintain a fixed number of nodes throughout the model’s operation, as seen in models like \citep{Lee2017DESIREDF,Amirian2019SocialWL}. If fewer agents are present in a frame than the fixed number, dummy nodes are introduced to maintain a consistent structure. Conversely, if the number of agents increases in a later frame, new agents are assigned previously unoccupied dummy node slots. Usually, valid and dummy nodes are handled through masking to exclude inactive nodes from message passing and feature updates. Implementation-wise, many PyTorch-based implementations of GNNs operate on static graphs with a fixed adjacency matrix size for computational efficiency. For dynamic graphs, to process a sequence of \( \{G_{t-T}, ..., G_t\} \) and learn temporal and spatial dependencies, Temporal Graph Networks (TGNs) can be employed, available in PyTorch Geometric.

Message passing can employ Graph Convolutional Networks (GCNs), which uniformly aggregate neighbor information, or Graph Attention Networks (GATs), which dynamically compute attention weights to prioritize more influential interactions. However, GCNs and GATs only capture spatial relationships from a single snapshot and do not inherently model temporal dependencies. To aggregate historical trajectory information and learn temporal dependencies, two approaches can be used when static graphs coupled with GCN/GATs are employed. The first option is to store past trajectories inside node embeddings, allowing the network to process temporal dependencies in a sequence-to-sequence manner. The second approach is to use an external temporal mechanism, such as recurrent networks (LSTMs, GRUs) or memory modules, to handle historical information before passing the processed features into the GNN. 

Alternatively, dynamic graphs coupled with Temporal Graph Networks (TGNs) provide more flexibility by naturally capturing temporal dependencies across a sequence of graphs. However, TGN relies on consistent object IDs across frames to correctly track agent-specific memory updates. If object IDs are inconsistent, past information cannot be reliably retrieved, leading to incorrect temporal associations. In such cases, alternative techniques like memory-based retrieval or node re-identification should be applied. The final GNN layer outputs refined node embeddings, which are then decoded into future trajectories. We summarize GNN-based methods across three categories: GCN-based models in Table \ref{tab:dl_gcns}, GAT-based networks in Table \ref{tab:dl_gats} and the rest of GNN-based models are shown in Table \ref{tab:dl_gnns}.

Focusing on interaction modelling, \citet{Wang2020GraphTCNSI} proposed the GraphTCN architecture, which combines edge-feature GATs to capture spatial information and TCNs for temporal dynamics. EFGAT models spatial interactions using relative spatial relations as edge features, dynamically learning adaptive adjacency matrices via multi-head graph attention layers. TCNs handle temporal dynamics with gated 1D convolutions, enabling efficient parallel processing for long-term dependencies. \citet{10.5555/3495724.3497384} introduced EvolveGraph, which leverages a latent interaction graph parametrized by learnable edge embeddings. This framework employs a two-stage training process: static graph learning initializes stable interactions, while dynamic graph refinement models temporal dependencies. A recurrent decoder generates multi-modal predictions based on the evolving graph structure, with training guided by supervised trajectory loss and relational consistency loss. Trajectron++ \citep{10.1007/978-3-030-58523-5_40}, a SOTA graph-structured recurrent model for trajectory prediction, represents the environment as a spatiotemporal graph, where nodes represent agents and edges capture inter-agent interactions. Semantic information from the environment, encoded via a CNN, is integrated into the graph as node or edge features. The GNN processes these enriched features, encoding agent histories and interactions using LSTMs and attention mechanisms. Finally, a CVAE framework generates diverse and multimodal predictions based on the GNN-encoded features.

Continuing on SOTA models, the GOHOME framework \citep{10.1109/ICRA46639.2022.9812253} adopts a graph-based approach for trajectory prediction, representing the HD map as a graph with lane segments as nodes and connectivity as edges. Graph convolutional layers capture spatial and agent-lane interactions, while cross-attention integrates past trajectories and map features. Lane-level heatmaps identify high-probability regions for future trajectories. Among the latest models, MTP-GO is a graph-based probabilistic model designed by \citet{Westny2023MTPGOGP} and EqMotion is a prediction model with invariant interaction reasoning proposed by \citet{10205349}. MTP-GO incorporates graph-gated recurrent units (Graph-GRU) to encode inter-agent interactions over time and employs neural ODEs to model dynamic motion constraints. Specifically, neural ODEs solve differential equations parametrized by neural networks to represent motion constraints. EqMotion ensures motion equivariance, guaranteeing predicted trajectories transform consistently under translation and rotation, and interaction invariance, maintaining stable inter-agent relationships. It achieves this through an invariant reasoning module for interaction graph inference and equivariant geometric feature learning to model temporal and spatial dependencies.

\begin{table*}[t]
    \centering
    \caption{Summary of reviewed DL-based Models relying on Attention Mechanism}
    \label{tab:att}
    \begin{adjustbox}{max width=1.0\textwidth}
    \begin{tabular}{|p{4cm}|c|p{15cm}|}
        \hline
        \textbf{Reference} & \textbf{Name} & \textbf{Description}  \\
        \hline 
        \citet{Mercat2019MultiHeadAF} (2019) &  - & Multi-head self-attention mechanism applied before and after encoding the input, where multiple heads specialize in distinct interaction patterns. \\ \hline
        \citet{Zhu2019ProbabilisticTP}  (2019)  & ARNP & Leverages global latent variables to encode uncertainty and capture fine-grained local dynamics. \\ \hline
        \citet{Pan2019LaneAttentionPV}  (2019)  & - &  Attention to aggregate information from lanes surrounding vehicle. \\ \hline
        \citet{Bhat2020TrajformerTP} (2020) & Trajformer & Operates within a transformer-based backbone, combining multi-headed self-attention to encode social and spatial dependencies among agents.  \\ \hline
        \citet{9341034}  (2020) & - & Goal-oriented lane attention mechanism, leveraging dot-product attention to predict the probability of each lane a vehicle is heading toward. \\ \hline
        \citet{Kamra2020MultiagentTP} (2021) \textbf{ }& - & Uses a pairwise attention mechanism with fuzzy decisions, focusing on sender-receiver interactions.\\ \hline
        \citet{9084255} (2021)  & - & Multi-head attention mechanism captures spatio-temporal dependencies, and uses multiple heads to model different-order interactions. \\ \hline
        \citet{Liang2021NetTrajAN} (2021)  & NetTraj & Local graph attention captures dynamic spatial dependencies with road networks and sliding temporal attention for historical states. \\ \hline
        \citet{9710777} (2021) & RAIN & Hybrid attention mechanism combining RL-based hard attention to select key information and soft graph attention to rank selected elements. \\ \hline
        %%%%%%%%%%%%%%%%%%%%%%%
        \citet{9577353} (2021)  & Introvert & Uses 3D spatio-temporal attention mechanism to capture human-human and human-space interaction, focusing on relevant scene elements. \\ \hline
        \citet{9576054} (2021)  & - & Uses multi-head attention, each head specializing in capturing distinct interactions and predictive modes.\\ \hline
        \citet{Sekhon2021SCANAS} (2021)  & SCAN & Interleaved spatial and temporal attention mechanism. \\ \hline
        \citet{9347678} (2021)  & Tra2Tra  & Global social spatial-temporal attention mechanism for interactions. \\ \hline
        \citet{Nayakanti2022WayformerMF} (2022)  & Wayformer & Employs multi-axis attention and optimizations like factorized attention and latent query attention to manage spatio-temporal and multimodal data. \\ \hline
        \citet{9349962} (2022)  & STA-LSTM & Uses spatial-temporal attention mechanisms. \\ \hline
        \citet{9732437} (2022)  & - & Integrates attention mechanism for temporal and spatial dependencies. \\ \hline
        \citet{PENG2022258} (2022)  & SRAI-LSTM & Social Relation Attention-based mechanism focusing on interactions within walking groups, aggregating motion features from neighbouring pedestrians. \\ \hline
        \citet{9373939} (2022)  & - &  Unified spatial-temporal attention mechanism to select key information from surrounding individuals in both spatial and temporal domains. \\ \hline
        \citet{ZHANG2022103829} (2022)  & MTPT & Uses a modified Swin Transformer, attention to identify the most critical spatial and contextual factors. \\ \hline
        \citet{Da2022PathAwareGA} (2022)  & PAGA & Path-Aware Graph Attention to capture interactions between lanes and agents by learning attention between vertices in a heterogeneous graph. \\ \hline
        \citet{Cao2022LeveragingSA} (2022)  & - & Attention with total variation temporal smoothness prior for interactions. \\ \hline
        \citet{Guo2022VehicleTP} (2022)  & - & Dual-attention mechanism integrating temporal and spatial attention. \\ \hline
        \citet{Duan2022ComplementaryAG} (2022)  & CAGN & Normal and inverse attention to capture frequent and rare patterns. \\ \hline
        \citet{10.1609/aaai.v37i3.25389} (2023)   & - & Uses self-attention for temporal dependency, sparse self-attention for spatial interactions, and cross-attention for spatio-temporal correlations. \\ \hline
        \citet{tang2024hpnet} (2024)  & HPNet & Triple Factorized Attention module: spatial agent attention for inter-agent interactions, historical prediction attention, and self-attention across different trajectory modes. \\ \hline
        \citet{10.1016/j.patcog.2023.109997}  (2024) & MFAN & Mixing feature attention mechanism to determine the importance of different interaction features.\\ \hline
         \citet{Sun2022InteractionMW}  (2024)  & - & Multiplex attention for interactions, by constructing layers in a latent graph, each layer for distinct interaction type. \\ \hline
    \end{tabular}
    \end{adjustbox}
\end{table*}

\begin{table*}[t]
    \centering
    \caption{Summary of reviewed DL-based Models relying on Transformers (part 1)}
    \label{tab:transformers_1}
    \begin{adjustbox}{max width=\textwidth}
    \begin{tabular}{|p{4cm}|c|p{15cm}|}
        \hline
        \textbf{Reference} & \textbf{Name} & \textbf{Description}  \\
        \hline 
        \citet{yu2020spatiotemporalgraphtransformernetworks} (2020) & STAR & Combines temporal Transformers for modelling individual pedestrian dynamics and a graph-based spatial Transformer to capture crowd interactions, with alternating layers to model spatio-temporal dependencies. \\ \hline
        \citet{giuliari2020transformernetworkstrajectoryforecasting} (2021) &  - & Uses encoder-decoder Transformer, encoding past trajectories with positional encodings and self-attention. \\ \hline
        \citet{9710708} (2021) & AgentFormer &  Agent-aware Transformer modelling temporal and social dimensions, with attention mechanism for intra-agent and inter-agent interactions. \\ \hline
        \citet{9575242} (2021) & - & Transformer architecture augmented with positional, heading information. \\ \hline
        \citet{9636241} (2021) & - & Incorporates bounding boxes, speeds, orientations, and road semantics, weighted by an attention module, for trajectory and intention prediction.  \\ \hline
        \citet{ijcai2021p174} (2021) & MTN & Models long-range temporal dependencies and uses co-attention mechanism for observed trajectories and ego-vehicle speed.  \\ \hline
        \citet{9577819} (2021) & mmTransformer  & Combines motion extractor for past trajectories, map aggregator for road topology, and a social constructor to model agent interactions. \\ \hline
        \citet{Huang2021MultimodalMP} (2021) & - & Employs multi-head attention for interactions and multi-modal attention mechanism to capture diverse agent-map relationships.\\ \hline
        \citet{Girgis2021LatentVS} (2021) & - & Uses interleaved temporal and social multi-head self-attention.\\ \hline
        \citet{Ngiam2021SceneTA} (2021) & - &  Uses self-attention layers for inter-agent interactions and integrates road graph information using cross-attention mechanisms. \\ \hline 
        \citet{pmlr-v157-chen21a} (2021) & S2TNet & Self-attention sub-layer for interactions and temporal Transformer decoder. \\ \hline
        \citet{Saleh2020PedestrianTP} (2020) & - &  Context-Augmented Transformer fusing positional information, agent interactions, and scene semantics. \\ \hline
        \citet{Zhou2022GASTTHT} (2022) & GA-STT & Combines spatial and temporal Transformers, using cross-attention to fuse spatial-temporal embeddings and capture both individual and group-level motion features.\\ \hline
        \citet{10.1007/978-3-031-20047-2_14}  (2022) & Social-SSL & Introduces three pretext tasks: interaction type prediction, closeness prediction, and masked cross-sequence to sequence prediction for encoding. \\ \hline 
        \citet{10.1007/978-3-031-19772-7_26}  (2022)  & - & Uses current-frame observations as query to suppress cascading errors and integrates spatio-temporal information in the encoder via self-attention and in the decoder via flipped cross-attention. \\ \hline
        \citet{Hu2022HolisticTA}  (2022) & - & Holistic Transformer with sparse multi-head attention for noise reduction, feature selection sparse attention for leveraging prior knowledge, and multi-head attention for utilizing posteriori knowledge. \\ \hline
        \citet{Li2022GraphbasedST}  (2022)  & - & Graph-Based Spatial Transformer with Memory Replay. \\ \hline
        \citet{9878832}  (2022)  & HiVT & Two-stage Transformer: first stage focuses on local context extraction and second stage integrates a global interaction module.  \\ \hline
        \citet{9927798}  (2022)  & - &  Graph-Based Transformer with a memory mechanism to enhance temporal continuity and employing HuberLoss for improved robustness. \\ \hline
        \citet{9812226} (2022)  & - &  Uses cross-modal transformers for cross-relation features between modality pairs, fused with a modality attention module. \\ \hline
        \citet{Achaji2022PreTRSN} (2022) & PreTR & Uses a factorized spatio-temporal attention module to extract features and non-autoregressive parallel decoding mechanism to mitigate exposure bias.  \\ \hline 
        \citet{zhang_2022}  (2022) & - &  Graph-Based Transformer uses GAT to model agent-agent and agent-infrastructure interactions, combining CNNs for spatial features. \\ \hline
        \citet{Amirloo2022LatentFormerMT}  (2022) & LatentFormer & Uses hierarchical attention to model interactions among agents and Vision Transformer to capture local and global scene context. \\ \hline
    \end{tabular}
    \end{adjustbox}
\end{table*}

Among GCN models, the GRIP architecture \citep{8917228} is designed for interaction-aware trajectory prediction and consists of three main components. First, the Input Preprocessing Model converts observed trajectories into a 3D tensor, followed by constructing a spatial graph where edges encode proximity-based interactions. Then, the GCN alternates between temporal convolutional layers to extract motion features and graph operation layers to propagate features across nodes using a spatial adjacency matrix and normalized graph operations. Its successor, GRIP++ \citep{Li2019GRIPEG}, enhances the original GRIP by refining interaction modelling with dynamic edge weights and incorporating scene context. GRIP++ introduces advanced GNN architectures with improved attention mechanisms, enforces temporal consistency, and optimizes computational efficiency. \citet{liang2020learning} proposed LaneGCN, which constructs a lane graph from vectorized HD maps to preserve topology and connectivity. The architecture includes four modules: ActorNet, encoding past trajectories of traffic agents; MapNet, encoding lane graph features using LaneGCN; FusionNet, modelling interactions between agents and lanes; and the Prediction Header, generating multimodal trajectory predictions with confidence scores. By integrating structured map features and dynamic actor-map interactions, LaneGCN effectively addresses motion forecasting in structured driving environments. Lastly, \citet{Liao2024MFTrajMB} proposed a map-free, behaviour-driven trajectory prediction model - MFTraj. The architecture consists of four components: the behaviour-aware module leveraging dynamic geometric graphs; the position-aware module emphasizing relative positions and spatial dynamics using LSTMs; the interaction-aware module with adaptive GCNs to model spatio-temporal dependencies; and a residual decoder that processes these features into predictions. 

%%%%%%%%%%%%%%%%%%%%%%%%%%%%%%%%%%%%%%%%%%%%%%%%%%%%

Among GAT Networks, Social-BiGAT, designed by \citet{e294141389194a54a05536938fcdd509}, is a generative model for pedestrian trajectory forecasting. It employs GAT to capture social interactions and a Bicycle-GAN architecture to generate multimodal trajectories. The model integrates an MLP to embed each pedestrian's relative displacements, an LSTM to encode temporal pedestrian movements, and a CNN to extract scene-level features from a BEV representation. To address the mode collapse issue in multimodal trajectory prediction, \citet{CHENG2023163} proposed GATraj, an attention-based graph model. It utilizes attention mechanisms for modelling spatial-temporal dynamics, a graph convolutional network for capturing agent interactions, and a Laplacian mixture decoder to ensure diverse and reliable multimodal predictions.

\subsubsection{Attention and Transformers}

\begin{table*}[t]
    \centering
    \caption{Summary of reviewed DL-based Models relying on Transformers (part 2)}
    \label{tab:transformers_2}
    \begin{adjustbox}{max width=\textwidth}
    \begin{tabular}{|p{4.0cm}|c|p{15cm}|}
        \hline
        \textbf{Reference} & \textbf{Name} & \textbf{Description}  \\
        \hline 
        \citet{Chen2023VNAGTVN} (2023) & VNAGT &  Combines a variational non-autoregressive graph transformer with a unified graph attention module to capture social and temporal interactions. \\ \hline 
        \citet{Mozaffari2023MultimodalMA}  (2023) & - & Decoder integrates maneuver-specific information via cross-attention to predict multimodal trajectories and maneuver sequences. \\ \hline
        \citet{Feng_2023}  (2023) & MacFormer & Two key modules: the Coupled Map integrates map constraints and Reference Extractor to identify and select the most relevant map features.\\ \hline
        \citet{10377193}  (2023) & GameFormer &  Uses self-attention mechanisms to extract spatial and temporal context and decoder refines interactions based on level-k game theory principles. \\ \hline
        \citet{10378344}  (2023)& - & Mode-level encoding and social-level decoding.   \\ \hline
        \citet{10398503}  (2024) & MTR++ & Employs a symmetric scene context encoding and a mutually-guided intention querying module to enhance interaction-aware predictions. \\ \hline
        \citet{10504962}  (2024)
        & -  & Uses GCNs to aggregate social interactions and refine adjacency matrices for spatio-temporal graphs and Transformers for temporal dependencies.  \\ \hline
        \citet{Shi2022MotionTW}  (2024) & MTR & Motion Transformer uses motion query pairs to combine global intention localization and local movement refinement. \\ \hline
    \end{tabular}
    \end{adjustbox}
\end{table*}

Attention mechanisms enhance prediction models by allowing them to focus on the most relevant spatial and temporal features. Spatial attention mechanisms assign importance weights to neighboring agents computed at each timestamp, based on features such as relative distance, velocity, and heading, which influence the target agent’s trajectory. Temporal attention processes sequences of historical states, assigning weights to time steps that have the most significant impact on future predictions. This is often integrated with recurrent architectures, where the hidden states from prior time steps are passed through an attention module to refine the context vector for trajectory prediction. Advanced implementations combine spatial and temporal attention in multi-head architectures, enabling the model to process multiple attention patterns in parallel, effectively handling heterogeneous agents in dense environments. Table \ref{tab:att} provides a summary of reviewed models relying on attention mechanisms.

Transformers extend attention mechanisms by unifying spatial and temporal dependencies across agents and time steps in a global framework. In trajectory prediction, they learn from past trajectories by employing self-attention layers within an encoder-decoder architecture. The encoder processes input trajectories combined with positional encodings to preserve temporal order while capturing long-range dependencies across time steps. Self-attention enables the model to filter out less relevant variations and identify meaningful motion patterns. The decoder generates future trajectories by leveraging cross-attention, which refines predictions based on learned spatio-temporal dependencies, ensuring consistency with past motion patterns. Tables \ref{tab:transformers_1}, \ref{tab:transformers_2} provide a summary of Transformer-based prediction models.

Among SOTA Transformers, \citet{9577819} proposed mmTransformer, a multimodal trajectory prediction framework utilizing stacked Transformers to hierarchically process information. It integrates a motion extractor for past trajectories, a map aggregator for road topology, and a social constructor to model agent interactions. The model employs a region-based training strategy, grouping trajectory proposals into spatial regions to optimize predictions for each region. \citet{Ngiam2021SceneTA} introduced the Scene Transformer, a factorized Transformer architecture leveraging self-attention layers to model inter-agent interactions. It incorporates road graph information via cross-attention mechanisms, enabling agents to account for both static and dynamic road features. \citet{pmlr-v157-chen21a} proposed a Spatio-Temporal Transformer for heterogeneous traffic prediction. The model integrates a spatial self-attention sub-layer to capture agent-agent interactions and combines these with a temporal Transformer decoder to refine predictions autoregressively, jointly modelling spatio-temporal dependencies within the encoder. \citet{9878832} presented the Hierarchical Vector Transformer (HiVT), a two-stage framework for multi-agent motion prediction. The first stage focuses on local context extraction, aggregating spatial and temporal features in localized agent-centric regions, while the second stage captures long-range dependencies across the scene through a global interaction module. This hierarchical approach ensures computational efficiency. Finally, a series of works \citep{yu2020spatiotemporalgraphtransformernetworks, Li2022GraphbasedST, zhang_2022} explore graph Transformers, which apply Transformer-based attention mechanisms to graph representations. Here, nodes denote agents and edges represent interactions. These models capture spatial dependencies through graph-based attention and integrate temporal dependencies using sequential attention mechanisms or positional encodings, enabling the simultaneous capture of local and global context.

\subsubsection{Strengths and Limitations of DL Methods}

Deep learning-based trajectory prediction models exhibit varying strengths and limitations depending on their architectural design. \textit{LSTM-based models} effectively capture temporal dependencies and can be trained in an autoregressive mode, similar to inference mode. They converge reliably but lack spatial reasoning and struggle with long-range temporal dependencies. \textit{CNN-based models} efficiently extract spatial features from image-based inputs such as BEV or rasterized maps and enable fast inference. In \textit{Autoencoders}, their variational forms support multimodal prediction by modeling uncertainty, but their performance depends heavily on the latent space structure and they may produce unrealistic trajectories if not properly trained. \textit{GAN-based models} can learn complex trajectory distributions and generate diverse outputs, yet they require large training datasets, can be challenging to stabilize, and may not generalize well under distribution shifts when run in out-of-distribution scenarios. \textit{GNN-based models} excel in dense interaction scenarios by explicitly modeling agent-agent relationships, but they add unnecessary complexity in sparse environments such as highway driving, where interactions are minimal. In this case, rather than model limitations, it becomes a matter of appropriate design choice. Dynamic graph-based GNNs depend on consistent object IDs for temporal tracking and can suffer from tracking errors. \textit{Transformers}, due to their ability to capture long-term dependencies, are expected to excel in long-horizon prediction. However, they are highly sensitive to training stability, require velocity-based inputs (e.g., $v_x, v_y$) for convergence (which is beneficial for all DL models), and demand extensive tuning and warm-up for optimal performance.

\subsection{Reinforcement Learning-based Methods}
\label{sec:rl}

Reinforcement Learning-based methods generate predictions by interpreting agent behaviour as a high-dimensional policy optimization problem. RL methods learn through a reward system to generate trajectories that maximize long-term objectives. This section reviews two main categories of methods: Inverse Reinforcement Learning and Imitation Learning. 

\subsubsection{Inverse Reinforcement Learning} 
Inverse Reinforcement Learning (IRL) methods aim to learn a reward function from expert demonstrations by observing trajectories and inferring the underlying structure that drives the demonstrated behaviours. Markov Decision Process (MDP) is a common framework used in IRL for trajectory prediction \citep{7795546, Choi2020RegularisingNN, Shen2018TransferablePM}. To obtain an estimate of the reward function, expert demonstrations are first analyzed to provide an initial estimate, capturing the implicit motivations behind the observed behaviour. Using the MDP framework, this reward function is iteratively refined by ensuring that the expert’s observed actions align with the optimal policy derived from the reward function. Once the reward function is fully inferred, it can be used to model decision-making and predict socially-aware trajectories in dynamic environments. For estimating the reward function, Maximum Margin IRL methods \citep{Choi2020RegularisingNN, NIPS2012_559cb990} aim to minimize the feature expectation difference between expert demonstrations and predictions. \citet{Choi2020RegularisingNN} employ Maximum Margin IRL to estimate the reward function while enforcing a margin-based separation between expert and non-expert trajectories. By alternating between reward optimization and trajectory prediction, this approach ensures realistic, context-aware predictions that align with observed behaviors. Another approach, Maximum Entropy IRL methods \citep{7139642, 7795546} aim to estimate the reward function by maximizing the entropy of the policy distribution under the inferred reward, ensuring robustness in cases with suboptimal or ambiguous expert trajectories. \citet{7795546} proposed a model that infers a cost function as a linear combination of static and dynamic features from expert demonstrations. Their approach further incorporates Maximum Entropy IRL to resolve ambiguities in cost estimation and integrates a heuristic policy model for risk-aware decision-making.

On hierarchical inverse reinforcement learning, \citet{Sun2018ProbabilisticPO} proposed a two-layer model: the first layer handles discrete decisions, such as yielding or merging, which partition the trajectory space into distinct classes; and the second layer focuses on continuous decisions, such as velocity adjustments and smoothness optimization, within each class. This framework learns probabilistic models for both layers from human driving demonstrations using Maximum Entropy IRL. Lastly, on the integration of IRL and Transformers, \citet{10.1109/TITS.2023.3285891} proposed a framework leveraging a spatial-temporal Transformer network to generate trajectory candidates. IRL is then applied to infer a reward function from expert-like behavior, evaluating these trajectories to select the optimal one.

\subsubsection{Imitation Learning} 
Imitation Learning models \citep{9156400} bypass the stage of learning explicit reward functions and instead extract an optimal policy directly from the data, generating future trajectories similar to expert demonstrations. \citet{Rhinehart2018r2p2AR} introduced the Re-parametrized Pushforward Policy (R2P2) model, which combines imitation learning and probabilistic modelling. While the model implicitly learns a policy mapping states to future trajectories through a neural network, it models uncertainty and multimodality in future trajectories using a latent variable framework. A latent variable is sampled from a fixed distribution to capture variability, and the re-parametrized pushforward mechanism transforms these latent variables into trajectory predictions via a neural network. \citet{10.5555/3600270.3601775} propose a model that leverages a probabilistic latent dynamics framework to generate future states and actions based on offline expert data, employing BEV representations to enhance spatial reasoning. Generative Adversarial Imitation Learning (GAIL) combines imitation learning with adversarial training, where a generator produces trajectories and a discriminator differentiates between expert and generated trajectories to optimize the policy. \citet{10.1109/IVS.2017.7995721} extend GAIL by incorporating recurrent policies using GRUs, enabling the model to capture temporal dependencies in driver behavior. \citet{8953498} enhances GAIL with a latent decision-making process, where latent decisions inferred from historical trajectories guide multimodal path generation. The generator-discriminator framework is further improved by modelling these latent decisions as distributions and optimizing mutual information for greater diversity and realism in predictions.  \citet{Choi2020TrajGAILGU} integrate GAIL with a Partially Observable Markov Decision Process (POMDP), allowing the generator to account for both observable past states and unobservable latent states when generating trajectories, while the discriminator ensures alignment with real data. 

\subsubsection{Strengths and Limitations of RL Methods} 

RL-based methods learn directly from expert driving demonstrations, enabling the model to replicate realistic motion patterns observed in human behavior. Imitation Learning and generative RL frameworks via stochastic policies or latent variable models naturally achieve multimodal trajectory prediction. Additionally, POMDP-based extensions allow forecasting under partial observability, where agents must reason about latent intent or unobserved factors. As for limitations, RL methods typically require extensive expert data and are computationally demanding to train. Although in IRL, the reward function is learned from demonstrations, learning a function which captures not just safety constraints but also social dynamics and traffic agents intent ambiguity is challenging to formalize. Additionally, the learned reward and policy representations can be hard to interpret when debugging is needed.
\section{Incorporating Driving Knowledge}
\label{sec:incorporating_knowledge}

Formalizing and incorporating driving knowledge into trajectory forecasting frameworks \citep{9341199, 9740533, Wang2020VehicleTP, XU2022375} tends to enhance performance and improve the explainability of prediction models. Human drivers accumulate such knowledge through learning and experience. Similarly, forecasting models can benefit from a combination of injected driving knowledge and patterns learned from data. The intuition is that integrating external constraints narrows the space of plausible future trajectories. For example, a vehicle's possible paths are shaped by road topology, lane directions, and adherence to traffic rules, e.g., traffic lights and right-of-way priorities at intersections and roundabouts. Vehicle motion is also constrained by physical factors such as turning radius and braking distance. Pedestrian motion, in turn, is governed by traffic rules like crossing at designated crosswalks or waiting for traffic lights at intersections. These behaviors can be formalized as structured knowledge to support trajectory forecasting. Beyond such conventional rules, less formal but still possible sources of driving knowledge include behavioral priors and region-specific driving styles. For instance, drivers may take turns from adjacent lanes to avoid congestion, or pedestrians may rush to cross even after the light has turned red, attempting to beat the incoming traffic flow due to long waiting times

Capturing these variations can enhance the performance of prediction models and improve generalizability in case the formalized knowledge is modular and can be adapted to the region of deployment. This section explores different types of driving knowledge and the ways in which this knowledge is incorporated. For a more detailed discussion, we refer the reader to the study \citep{10100881}.

\subsection{Types of Driving Knowledge}

We categorize driving knowledge into three groups: (i) map-related information that encodes the static environment, such as road topology, pedestrian crossings, stop lines, etc.; (ii) agent-related information about the kinematics of vehicles, specific driving styles and pedestrians mobility constraints due to age, disability, or behavioral factors; and (iii) traffic-related knowledge, which captures information about traffic rules and interaction patterns between road users.

\subsubsection{Map-related knowledge}
HD maps provide information about road geometry, traffic lights, stop lines, pedestrian crosswalks, and traffic signs. Section \ref{sec:input_modalities} discusses rasterizing map data into grid images and encoding them using CNNs. While effective for extracting spatial features, this method often results in information loss, as the structured and geometric details of the map are rasterized, reducing precision. Recent advancements use GNNs to encode HD maps, preserving structural and semantic details while remaining scalable.

One spotlight example is VectorNet, proposed by \citet{Gao2020VectorNetEH}, which represents HD map elements and agent trajectories in a vectorized format. Each map element is broken into polylines composed of individual vectors, forming local graphs where nodes represent vectors and edges capture spatial or sequential relationships. These local graphs are aggregated into a global graph to encode lane-to-lane or agent-to-lane interactions using a hierarchical GNN. This approach enables interaction-aware feature encoding for downstream tasks. Next, \citet{liang2020learning} introduced LaneGCN, which leverages HD maps by constructing a lane graph where each lane segment is a node, and edges denote connections like predecessors, successors, and neighboring lanes. The model dynamically encodes the scene by capturing the topology and dependencies of the lane graph. By integrating actor-map interactions through a fusion network, it combines real-time traffic information with map-based context for prediction.  \citet{Deo2021MultimodalTP} proposed a complementary approach, which focuses on conditioning trajectory predictions on specific paths within the lane graph. The proposed method uses discrete policy rollouts to select plausible routes, reducing computational complexity by targeting relevant map regions and decoupling lateral variability (e.g., route choices) from longitudinal variability (e.g., speed, acceleration).

\subsubsection{Agent-related knowledge}
Accounting for a vehicle’s kinematics is a common feature in prediction models \citep{Xie2018VehicleTP, 9815528}, as it determines turning radius, braking distance, and acceleration capacity —-- all factors which influence the vehicle’s future motion. In terms of driving style, analyzing a target vehicle’s historical trajectories allows the model to associate its behavior with known driving patterns. This behavioral prior can then be used to improve prediction, e.g., an aggressive driver may be more likely to select risky maneuvers such as overtaking with insufficient gaps or forcing trailing vehicles to brake abruptly.

\subsubsection{Traffic-related knowledge}
Traffic-related knowledge includes additional driving information beyond agent- and map-related categories. For instance, incorporating explicit traffic rules, such as speed limits and right-of-way regulations into prediction models can help generate trajectories that better comply with expected behaviors in regulated environments. \citet{9636143} discuss the impact of such rules on intersection interactions, where vehicles may be required to stop, yield, or proceed depending on traffic signals or signs. In roundabouts, \citet{8813796} highlight how right-of-way rules constrain permissible behaviors and support the generation of conflict-free trajectories for all participants. \citet{9304764} observe that vehicle acceleration patterns often reflect compliance with traffic regulations, such as slowing near crosswalks or maintaining appropriate speeds in restricted zones. 

Another source is interaction knowledge helping participants coexist in dynamic environments. In the study \citep{9636143}, pairwise interactions at intersections are categorized into discrete types such as ignoring, going, or yielding, based on the behavior of the target vehicle.  \citet{9811567} extend this approach by introducing follow and yield interactions, requiring models to explicitly predict these labels to better capture traffic dynamics. For more complex scenarios, such as roundabouts, \citet{8813796} define interaction types using reference paths of interacting vehicles, enabling prediction of conflict-free trajectories. \citet{9304764} link interactions to acceleration patterns, proposing a sub-network that predicts vehicle acceleration to infer interaction-driven trajectories. Additionally, social norms and informal yielding practices can also be of use, e.g., a driver signaling a pedestrian to pass. These implicit forms of interaction knowledge can improve prediction accuracy in heterogeneous dense traffic scenarios.

\begin{table*}[t]
    \centering
    \caption{Summary of reviewed Models leveraging Driving Knowledge}
    \label{tab:dl_driving_knowledge}
    \begin{adjustbox}{max width=\textwidth}
    \begin{tabular}{|p{4cm}|c|p{14cm}|}
        \hline
        \textbf{Reference} & \textbf{Type of Knowledge} & \textbf{Description}  \\
        \hline 
        \citet{Lee2017DESIREDF} (2017) & map, traffic-related & For map knowledge, utilizes static semantic information from maps, such as road layouts, drivable areas, and road boundaries; for traffic knowledge: uses predefined cost function evaluating trajectories based on collision avoidance, road compliance, and interaction dynamics.\\ \hline
        \citet{Xie2018VehicleTP} (2018) & agent-related &  Leverages vehicle kinematic parameters such  acceleration and velocity. \\ \hline
        \citet{8967708} (2019) & traffic-related & Integrates traffic rules as Signal Temporal Logic (STL) and filters out invalid trajectories. \\ \hline
        \citet{Greer2020TrajectoryPI} (2020) & traffic-related & Uses a designed auxiliary loss function, which penalizes trajectories deviating from the expected lane heading. \\ \hline
        \citet{s20174703} (2020) & map-related & Incorporates curvilinear coordinates, lane boundaries, and feasible maneuvers from HD maps. \\ \hline
        \citet{9197560} (2020) & agent-related & Embeds a kinematic model of two-axle vehicles, constraining turning radius, acceleration, and heading to align predictions. \\ \hline
        \citet{9196738} (2020) & map, agent-related &  Uses a vector map to define lane geometry and encodes the geometric relationship between a vehicle's state and surrounding lanes. Incorporates vehicle-motion constraints. \\ \hline
        \citet{Deo2021MultimodalTP} (2021) & map-related & Captures geometric and semantic details like lane centerlines, stop lines, crosswalks, and traffic flow directions. \\ \hline
        \citet{Bahari2021InjectingKI} (2021) & map, agent-related & For map-knowledge, extracts from HD map lane geometry, centerlines, and boundaries; for agent-relates knowledge, integrates kinematic constraints through Model Predictive Control (MPC), such as turning radius and speed limits. \\ \hline
        \citet{9878679} (2022) & map-related &  Uses lane segments extracted from HD map, encoding geometric and semantic information, as proposals for predictions.  \\ \hline 
        \citet{9815528} (2022) & agent-related & Incorporating vehicle dynamics constraints through Model Predictive Control (MPC). \\ \hline
    \end{tabular}
    \end{adjustbox}
\end{table*}

\subsection{Injecting Driving Knowledge}

Table \ref{tab:dl_driving_knowledge} provides a brief summary of learning-based models that incorporate driving knowledge, including the type of knowledge and the way it is integrated.

For agent-related knowledge, a set of works focus on integrating kinematic parameters. \citet{Xie2018VehicleTP} proposed the Interactive Multiple Model (IMM) method leveraging vehicle kinematic parameters such as acceleration and velocity. \citet{9197560} embedded a kinematic model of two-axle vehicles into the prediction pipeline, constraining outputs like turning radius, acceleration, and heading to align with vehicle dynamics. Additionally, works \citep{Bahari2021InjectingKI, 9815528} incorporated vehicle dynamics and kinematics constraints through Model Predictive Control (MPC), ensuring that predicted trajectories comply with speed limits and acceleration.

In terms of traffic-related knowledge, DESIRE \citep{Lee2017DESIREDF} incorporates this knowledge through a predefined cost function within its Inverse Optimal Control (IOC)-based ranking module. The cost function evaluates trajectory predictions based on criteria such as collision avoidance, road compliance, and interaction dynamics, assigning lower costs to socially and contextually plausible trajectories to ensure alignment with expected traffic norms. \citet{Greer2020TrajectoryPI} proposed the YawLoss framework, which integrates traffic-related knowledge by enforcing vehicle motion constraints aligned with lane directionality. This is achieved through a designed auxiliary loss function (YawLoss), which penalizes trajectories deviating from the expected lane heading ensuring lane-following behavior. \citet{8967708} integrates traffic-related knowledge by representing predefined traffic rules as Signal Temporal Logic (STL) formulas. The model jointly predicts future trajectories and the degree of satisfaction (robustness slackness) for these rules, filtering out invalid trajectory predictions. 

Map-related knowledge represents the largest class of models integrating driving knowledge \citet{Bahari2021InjectingKI, Lee2017DESIREDF}. \citet{9878679} proposed a lane-based prediction model that represents lane segments as vectorized elements, encoding geometric and semantic information and enabling interaction modelling between agents and lane segments. \citet{s20174703} proposed a road-aware prediction model that incorporates curvilinear coordinates, lane boundaries, and feasible maneuvers from HD maps, using this information as prior knowledge to constrain the predictions.  Next, \citet{9196738} introduced the LAMP-Net architecture for lane-aware trajectory prediction, by encoding the geometric relationship between a vehicle's state and surrounding lanes. Lastly, DESIRE \citep{Lee2017DESIREDF} integrates map-related knowledge through the Scene Context Fusion (SCF) layer, which utilizes static semantic information from maps, such as road layouts, drivable areas, and road boundaries, to provide contextual guidance for trajectory generation.

\section{Uncertainty Estimation and Mitigation}
\label{sec:uncertainty}

\begin{figure*}[h!]
\centering
   \includegraphics[width=1.0\textwidth]{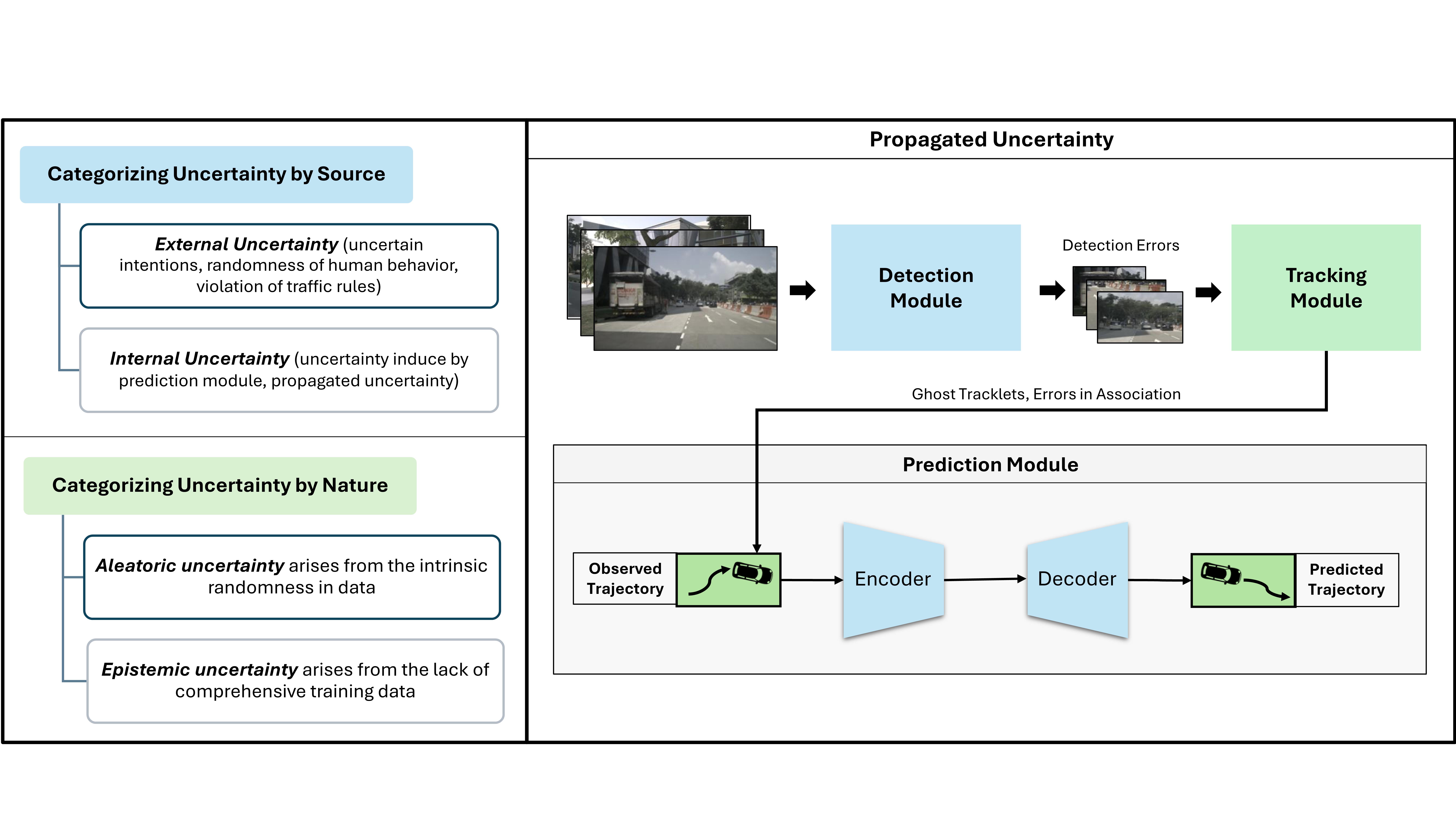}
   \caption{Types of Uncertainty: the left side shows the categorization followed in the section and the right part demonstrates the flow of the prediction model and propagation of perception and tracking modules' errors in detect-track-predict paradigm.}
   \label{fig:uncertainty}
\end{figure*}

Accounting for different sources of uncertainty and designing uncertainty-aware prediction models is an active research area across both academia and industry. For robust autonomous prediction, uncertainty must be explicitly modeled and systematically addressed at each stage of the autonomy stack. This section provides a high-level overview to build intuition on what constitutes uncertainty-aware modeling and why such awareness is critical. It covers sources of uncertainty (please refer to Figure \ref{fig:uncertainty}), techniques for both quantifying and mitigating uncertainty. However, we strongly recommend readers interested in a deeper treatment of uncertainty across the full autonomy pipeline to refer to the  reviews by \citet{10122777, wang2025deployablegeneralizablemotionprediction}.

\subsection{Sources of Uncertainty}

External uncertainty arises from the environment and includes factors beyond the model’s control. These encompass unobservable agent intentions, occlusions that limit visibility, environmental variability such as adverse lighting or weather conditions, and unexpected objects or rare events, including road accidents. For example, violations of traffic rules, such as crossing solid lanes or running red lights, can severely impact prediction performance by expanding the set of plausible future trajectories and degrading model reliability. Another challenge arises from rare events, like animals crossing the road are often underrepresented in training data.

Internal uncertainty arises from limitations within the prediction system itself. It reflects factors such as poor generalization to unseen scenarios, biases in the training data, and insufficient model capacity to capture the complexity of real-world driving behavior. A common contributor to internal uncertainty is propagated error, which originate from upstream modules that affect downstream predictions. For example, perception systems may produce imprecise outputs, such as shifted or shrunken bounding boxes, or entirely miss detections. When prediction is tightly coupled with perception, as in joint learning setups, such errors directly degrade forecasting performance. When connected through a tracker, perception errors may be partially compensated by estimations, but this can introduce new issues such as ID switches or misassociations, resulting into perturbed past trajectories. Consequently, any system that relies on outputs from earlier modules is inherently susceptible to accumulating uncertainty throughout the pipeline.

By nature, uncertainty can be categorized into aleatoric and epistemic. Aleatoric uncertainty arises from intrinsic randomness in the data. In the context of autonomous driving, it is often linked to sensor characteristics, such as measurement noise, limited sensing range, calibration errors, or low resolution. Aleatoric uncertainty reflects the inherent difficulty in fully capturing the environment. For instance, predicting whether a pedestrian will cross at a crosswalk or continue walking along the sidewalk involves uncertainty due to the multimodal nature of human behavior. Aleatoric uncertainty can be further categorized as heteroscedastic (varying across samples, e.g., lighting conditions affecting image quality) or homoscedastic (uniform across samples, e.g., consistently miscalibrated sensors). Epistemic uncertainty, on the other hand, arises from limited knowledge of the environment, often due to insufficient or non-diverse training data. It can be reduced through improved data collection, better modeling techniques, and increased exposure to diverse scenarios, e.g., cross-dataset domain adaptation. Many motion prediction models assume that training, testing, and deployment data share the same distribution, but this assumption rarely holds in real-world applications. The resulting domain gap can degrade performance, especially when encountering rare or previously unseen scenarios. Addressing epistemic uncertainty is therefore essential for improving model generalization and ensuring reliable operation in diverse, real-world settings.

\begin{table*}[h]
    \centering
    \caption{Summary of reviewed Prediction Models quantifying Uncertainty}
    \label{tab:uncertainty_qt}
    \begin{adjustbox}{max width=\textwidth}
    \begin{tabular}{|p{4cm}|c|p{15cm}|}
        \hline
        \textbf{Reference} & \textbf{Type of Uncertainty} & \textbf{Description}  \\
        \hline 
        \citet{Bhattacharyya2017LongTermOP} (2017) & epistemic, aleatoric & Captures aleatoric uncertainty by modelling the distribution of bounding box predictions and epistemic through posterior distributions over model parameters.\\ \hline 
        \citet{8794282} (2019) & epistemic, aleatoric & Uses a mixture-of-experts architecture, with confidence estimations determining the best predictor for specific scenarios. \\ \hline
        \citet{9093332} (2020) & epistemic, aleatoric & Quantifies aleatoric uncertainty by modelling displacement errors and epistemic uncertainty through dropout-based Bayesian analysis. \\ \hline
        \citet{Casas2020ImplicitLV} (2020) & epistemic, aleatoric & Captures scene dynamics, actor interactions, and unobservable variables in a distributed latent space using GNNs, leveraging the Evidence Lower Bound (ELBO) framework to balance reconstruction accuracy and latent variable expressivity. \\ \hline
        \citet{pustynnikov2021estimating} (2021) & epistemic, aleatoric & Quantifies uncertainty using a spectral-normalized Gaussian process (SNGP), which estimates posterior variance to measure scene uncertainty and detect out-of-distribution scenarios. \\ \hline
        \citet{Ivanovic2021HeterogeneousAgentTF} (2021) & epistemic, aleatoric & Quantifies uncertainty by incorporating class probabilities from perception systems into a CVAE framework, where these probabilities influence the output distributions of predictions. \\ \hline
        \citet{9815528} (2022) & epistemic, aleatoric & Incorporates quantified prediction uncertainty into MPC framework with a potential field that accounts for vehicle dynamics, road constraints, and prediction uncertainty. \\ \hline
        \citet{10173747} (2023) & interaction-driven & Quantifies individual uncertainty and interaction-driven collaborative uncertainty using multivariate Gaussian and Laplace distributions. \\ \hline
        \citet{Liu2023UncertaintyAwarePT} (2023) & aleatoric & Maps trajectories into bivariate Gaussian distributions, separating intrinsic uncertainty, and employs diffusion processes for trajectory sampling. \\ \hline
        \citet{10657827} (2024) & map uncertainty & Extends HD map estimation models to output Laplace-distributed location uncertainties for map elements and confidences for road elements, capturing spatial ambiguity. \\ \hline
    \end{tabular}
    \end{adjustbox}
\end{table*}

\subsection{Mitigating Uncertainty}
The reliability of data-driven trajectory prediction models heavily depends on the representativeness of the training data. When deployed in out-of-distribution (OOD) scenarios, where traffic conditions deviate significantly from those seen during training, model performance often degrades due to distributional shifts. One challenge lies in determining whether the training data sufficiently captures the variability of real-world environments. To this end, \citet{10.1145/3550270} propose a structured testing framework, advocating that scenario coverage should span key factors influencing autonomous decision-making, with statistical tools used to handle simulation uncertainties. Beyond dataset diversity, models should also be equipped with mechanisms to detect OOD inputs at runtime. OOD detection methods include label-based techniques that monitor prediction error, Bayesian approaches that estimate epistemic uncertainty by modeling distributions over outputs or parameters \citep{10.5555/3045390.3045502, lakshminarayanan2017simple}, and generative techniques that learn the input data distribution and flag deviations \citep{lee2018training}.

To address uncertainty sourced from incomplete inputs, \citet{Liao2024PhysicsInformedTP} proposed a physics-informed model employing a Wavelet Reconstruction network to recover missing historical data and a Kinematic Bicycle model to enforce physically plausible predictions. A Risk-Aware module further integrates real-time risk factors using probabilistic attention, enabling adaptation to underrepresented scenarios.  \citet{Bhattacharyya2018BayesianPO} mitigated aleatoric and epistemic uncertainty by employing synthetic likelihoods, which relax the requirement to explain every data point, enabling the model to generate diverse hypotheses for uncertain and multi-modal cases. Next, \citet{10657146} introduced the Out-of-Sight Prediction framework (OOSTraj), which mitigates uncertainty in predictions for out-of-sight agents by utilizing a mobile denoising encoder and a camera parameters estimator to address sensor noise and map sensor-based to visual trajectories. The framework employs a Transformer-based decoder to predict trajectories even in the absence of direct visual observations.

\citet{10657124} proposed a causal disentanglement approach to mitigate uncertainty caused by distribution shifts. Using a Transformer-GNN with disentangled attention blocks, the model separates causal and spurious factors in driving scenes, simulates potential shifts, and incorporates an invariance objective to focus on causal features while suppressing spurious ones for enhanced robustness. Next, \citet{8578474} addressed cascaded uncertainty by designing an end-to-end framework, ``Fast and Furious" (FaF), which integrates 3D detection, tracking, and motion forecasting into a single model. By jointly optimizing these tasks, FaF implicitly mitigates cascading errors across the detection and prediction pipeline.

\subsection{Quantifying Uncertainty}
Quantifying prediction uncertainty and incorporating it into downstream decision-making can reduce driving risks by enabling more cautious and informed planning. Instead of relying on a single deterministic forecast, many models generate multi-modal predictions for each agent \citep{10.1109/IVS.2018.8500493, 8813796, Gupta2018SocialGS, 10.1109/ICRA.2019.8793868, Malla2020SocialSTAGESM}. This helps capture uncertainty stemming from the inherent ambiguity in agents’ future intentions, since a single past trajectory can lead to multiple plausible futures. To represent these possibilities, some methods estimate probability distributions over future trajectories. Common approaches include modeling outcomes using parametric forms such as Gaussian distributions \citep{Vemula2017SocialAM, 7780479, Choi2019DROGONAC, 9578053} or Gaussian Mixture Models (GMMs) \citep{10.1109/ICRA.2019.8793868, 8953435}, which allow capturing both the mean prediction and its associated uncertainty. Table \ref{tab:uncertainty_qt} summarizes deep learning-based models for quantifying uncertainty.

\citet{Bhattacharyya2017LongTermOP} proposed a Bayesian LSTM framework for uncertainty quantification, capturing aleatoric uncertainty by modelling pedestrian bounding box distributions and epistemic uncertainty through posterior distributions over model parameters. Final predictions are represented as probabilistic distributions, providing a measure of reliability. \citet{8794282} introduced a variational neural network that evaluates the value of information from each input and integrates with other prediction models to create a composite expert system, generating confidence estimates across different time horizons to effectively manage uncertainty. Similarly,  \citet{9093332} quantified aleatoric uncertainty by modelling displacement errors and epistemic uncertainty through dropout-based Bayesian analysis, combining prediction variances from multiple stochastic forward passes to derive final confidence estimates. Building on these models, \citet{LI2024104659} designed UQnet, which employs a 2D histogram-based deep learning model with deep ensemble techniques, using entropy-based metrics to quantify both types of uncertainty. Lastly, \citet{Gilles2022UncertaintyEF} quantified uncertainty by calculating the variance of predicted spatial probability distributions from heatmap outputs, providing confidence measures for trajectory predictions and dynamically adjusting the sampling radius based on estimated uncertainty.
 
\section{Planning-conditioned Trajectory Prediction}
\label{sec:planning-driven}

Planning-conditioned trajectory prediction methods rely on the assumption that agents act rationally and their future trajectories are guided by a motion planner. Models \citep{Song2021LearningTP, 9009551, 7487505, song2020pip} effectively adopt this paradigm by introducing planning-prediction-coupled frameworks, where predictions are conditioned on observed motion plans or intentions. These methods go beyond solely relying on past trajectory, contextual, and interaction features; by modelling the notion that agents follow potentially planned trajectories or have specific destinations to reach. 
 
To review planning-conditioned approaches, we follow two categorizations. In the first category, the approaches are divided into \textbf{\textit{model-based}} planning, \textbf{\textit{learning-based}} planning, and \textbf{\textit{combined}} methods, which are reviewed in Section \ref{sec:plan-based-approaches}. In the second category, it focuses on anchor-based prediction methods, where trajectories are generated based on predefined anchors or guidance points derived from the environment or learned from data. These anchors typically represent potential future goals, paths, or key waypoints. Anchor-based methods can be further divided into \textbf{\textit{goal anchor-based}} prediction, where tentative goals for the agents are estimated, and \textbf{\textit{trajectory-anchor-based}} prediction, where possible trajectories are directly constructed based on environmental and physical constraints. This categorization is further explored in Section \ref{sec:planning-anchor-based}.

\subsection{Planning Methods for Trajectory Prediction}
\label{sec:plan-based-approaches}

Motion planning literature has extensively developed methods for generating safe and efficient trajectories for ego vehicles. This has inspired the application of planning algorithms to predict the trajectories of other agents, under the assumption that their future paths align with specific objectives. Accordingly, planning-conditioned trajectory prediction methods adopt a similar categorization to motion planning: \textit{model-based}, \textit{learning-based}, and \textit{combined} approaches \citep{10122127}. Model-based approaches rely on explicit rules, physical constraints, and optimization techniques. In contrast, learning-based approaches infer plans or intentions directly from observed data. Combined methods aim to integrate the strengths of both paradigms. Reviewed works using these approaches for trajectory prediction are summarized in Table~\ref{tab:planing_based}.

\begin{table*}[h]
    \centering
    \caption{Summary of Model-based, Learning-based, and Combined planning methods for Trajectory Prediction}
    \label{tab:planing_based}
    \begin{adjustbox}{max width=\textwidth}
    \begin{tabular}{|p{3.5cm}|p{4cm}|p{15cm}|}
        \hline 
        \textbf{Type} & \textbf{Reference} & \textbf{Description}  \\
        \hline 
         \multirow{4}{*}{\begin{tabular}{@{}c@{}} Model-based \end{tabular}} & \citet{6126296} (2011) &   Optimizes a handcrafted cost function to plan paths for the observed agents.  \\ \cline{2-3}
         & \citet{yan2014modeling} (2014) & Introduces a probabilistic motion model  towards specific goals in the environment. \\ \cline{2-3}
        & \citet{8317848} (2017) & Uses  A* algorithm  with the potential cost function for trajectory prediction.   \\  \cline{2-3}
        & \citet{9197560} (2020) & Accounts for planning kinematic constraints while predicting trajectories.     \\  \hline 
        \multirow{6}{*}{\begin{tabular}{@{}c@{}} Learning-based \end{tabular}} & \citet{7838128} (2016) & Uses IRL for pedestrian trajectories by modelling their interactions with the environment, such as avoiding collisions or heading toward crosswalks. \\ \cline{2-3}
        & \citet{Choi2020TrajGAILGU} (2020) & GAIL for matching long-term  distributions over states and actions. \\ \cline{2-3}
        & \citet{10203873} (2023) & Uses reusable scene encoding and a query-based  decoder with adaptive refinement to handle multimodal future behaviour. \\ \cline{2-3}
        & \citet{10313963} (2024)  & Uses GNN for trajectory predictions conditioned on guided-goals. \\ \hline
         \multirow{3}{*}{\begin{tabular}{@{}c@{}} Combined Approaches \end{tabular}} & \citet{Song2021LearningTP} (2021) & PRIME combines model-based generation with learning-based evaluation to ensure accuracy, feasibility, and multimodality in trajectory prediction. \\ \cline{2-3}
         & \citet{chen2024interactive} (2024) & Combines model predictive control with learned prediction model to optimize joint behavior.
         \\ \hline 
    \end{tabular}
    \end{adjustbox}
\end{table*}

\subsubsection{Model-based planning approaches}
Model-based planning approaches integrate physical motion principles, behavioral models, control objectives, and road structure information to generate interpretable and feasible future trajectories \citep{helbing1995social, houenou2013vehicle}. Core components include: dynamics models representing how agents' state evolves over time based on kinematic equations or physical motion laws \citep{polack2017kinematic}; cost functions, which define optimization objectives such as minimizing energy use, travel time, or collision risks \citep{6126296}; and planners used to solve the resulting optimization problems under physical and contextual constraints. 

\subsubsection{Learning-based planning approaches}
Learning-based planning for trajectory prediction employs data-driven models to generate trajectories for target agents \citep{Song2021LearningTP}. These methods often employ reinforcement learning to infer planning strategies directly from data. By training on large datasets, they capture both high-level decision-making (e.g., navigating junctions or changing lanes) and low-level control behaviors (e.g., reacting to obstacles). By learning implicit behavioral patterns and modeling multi-agent interactions, these approaches generate predictions in dynamic and complex traffic scenarios \citep{10.1109/ICRA.2019.8793868,Mercat2019MultiHeadAF,Lee2017DESIREDF}.

\citet{7838128} use inverse reinforcement learning (IRL) to predict pedestrian trajectories by inferring cost functions from observed behavior. The learned cost captures decision-making patterns such as avoiding collisions or moving toward crosswalks, and is used in a planner to generate plausible future paths. Further, the study by \citet{10.1109/IVS.2018.8500493} uses LSTM to predict maneuver classes (e.g., lane changes, braking) for target agents, and generates corresponding trajectories by assigning probabilities to each maneuver. Next, model  \citep{ziebart2009planning} learns a cost function from observed pedestrian trajectories to predict future positions. The ego vehicle plans its trajectory by jointly considering predicted pedestrian motion and its own objectives. The intention-interaction graph method proposed by \citet{li2022intention} jointly models pedestrians' intentions (goals) and their social interactions. The graph represents each pedestrian and their intended destination as nodes, enabling the model to predict trajectories by inferring goals based on observed behaviors and interactions.

In summary, learning-based planning for trajectory prediction leverages large-scale, real-world data to uncover both strategic and reactive driving patterns. By using reinforcement and inverse reinforcement learning, these approaches infer cost functions and planning policies directly from observed behavior, enabling the generation of realistic future paths for both vehicles and pedestrians
 \subsection{Anchor-based Trajectory Prediction}
 \label{sec:planning-anchor-based}
 
\begin{figure*}[t]
\centering
\includegraphics[width=1.0\textwidth]{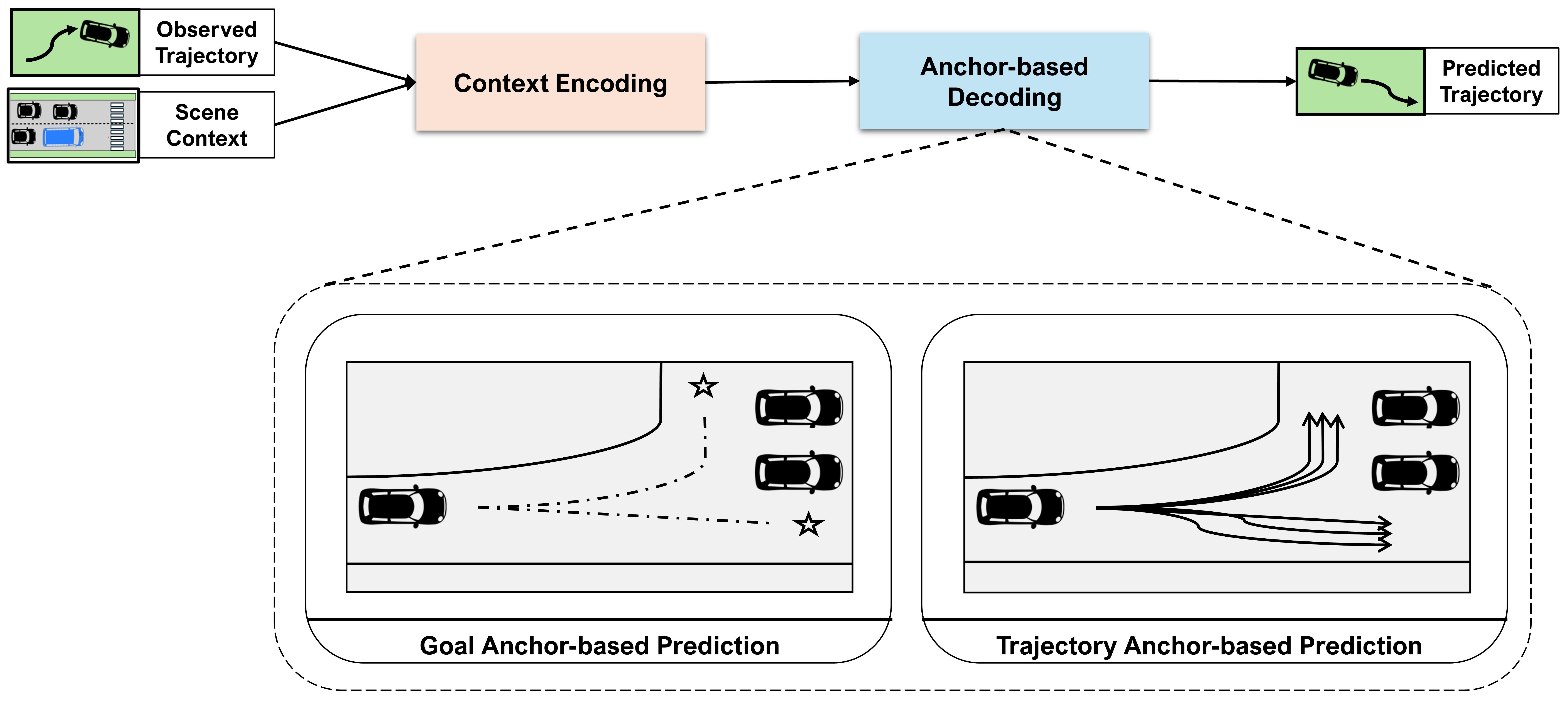}
   \caption{A general pipeline for anchor-based prediction models: (left) - goal anchor-based and (right) - trajectory anchor-based.}
   \label{fig:planning_prediction}
\end{figure*}

Anchor-based trajectory prediction methods assume that future trajectories are conditioned on planning objectives, such as goal destinations or potential paths. These methods use predefined spatial or trajectory ``anchors" to guide the prediction process. Figure \ref{fig:planning_prediction} illustrates the general pipeline for the coupled planning and prediction framework commonly found in the literature. The pipeline begins by encoding past trajectories and scene context into a context vector. This context vector is then utilized in the anchor-based decoding process to predict possible future trajectories. Anchor-based decoding can be guided by two types of anchors: \textbf{\textit{goal anchors}} or \textbf{\textit{trajectory anchors}}. Further we discuss each category and provide a summary of reviewed anchor-based prediction methods in Table \ref{tab:anchor-based}.                       

\begin{table*}[h]
    \centering
    \caption{Summary of Anchor-based Trajectory Prediction methods}
    \label{tab:anchor-based}
    \begin{adjustbox}{max width=\textwidth}
    \begin{tabular}{|c|p{4.5cm}|p{15cm}|}
        \hline 
        \textbf{Type} & \textbf{Reference}  & \textbf{Description}  \\
        \hline 
         \multirow{24}{*}{\begin{tabular}{@{}c@{}} Goal \\ Anchor \\ Based \end{tabular}} & \citet{Zhang2020MapAdaptiveGT} (2020) & Selects potential goal points based on observed trajectories and contextual features and ranks candidate goals based on learned scoring function.   \\  \cline{2-3}
         & \citet{Tran2020GoaldrivenLT} (2020) & Dual-channel framework: a goal channel predicts potential destinations, and a trajectory channel generates detailed paths conditioned on these goals. \\ \cline{2-3}
         & \citet{mangalam2020not} (2020) & Generates goals by sampling plausible future endpoints directly from the scene context, leveraging map features, and observed trajectories. \\  \cline{2-3}
         & \citet{9709992} (2021) & Hierarchically models goals, waypoints, and paths, generating scene-compliant multimodal predictions. \\  \cline{2-3}
         & \citet{9710322}  (2021) &  Leverages Dynamic Time Warping (DTW) to retrieve pseudo goal positions from an expert repository of previously observed trajectories.   \\ \cline{2-3}
         & \citet{9345445}  (2021) &  Uses a recognition network to encode the relationship between past trajectories and target goals, and a prior network for sampling. \\ \cline{2-3}
        & \citet{wang2022stepwise} (2022) & Generates coarse stepwise goals iteratively, incorporates them using an attention-based aggregator. \\  \cline{2-3}
        & \citet{9857236}  (2022)   & U-Net-based architecture to predict future goals as probability distributions based on scene semantics.  \\  \cline{2-3}
        & \citet{wang2023ganetgoalareanetwork}  (2023)  &  Uses a "GoICrop" module to aggregate map features around potential goal areas, incorporating spatial interactions between agents and map.   \\  \cline{2-3}
        & \citet{Dong2023SparseIC} (2023) & Learns goal candidates by identifying representative points from future trajectories, using a learned mask conditioned on observed trajectories to guide selection. \\  \cline{2-3}
        & \citet{Bae2023ASO} (2023) & Predicts a sequence of control points to divide pedestrian’s trajectory into three sections, capturing intermediate decisions. \\  \cline{2-3} 
        & \citet{wang2023prophnet} (2023) & Generates goal anchors via self-attention mechanism on the unified agent-centric scene representation, embedding contextual information. \\  \cline{2-3}
%%%%%%%%%%%%%%%%%%%%%%%%%%%%%%%%%%%%%%%%%%%%%%%
        & \citet{10313963} (2024)  & Goals are predicted for each agent using a goal-guided state refinement (SSR) module. \\  \hline    
%%%%%%%%%%%%%%%%%%%%%%%%%%%%%%%%%%%%%%%%%%%%%%%%%%%%%%%%%%%
        \multirow{6}{*}{\begin{tabular}{@{}c@{}} Trajectory \\ Anchor \\ Based \end{tabular}} &  \citet{Song2021LearningTP} (2021) & Samples dynamic reference paths and imposes explicit kinematic and environmental constraints to produce feasibility-guaranteed trajectories. \\ \cline{2-3}  
        & \citet{10.1109/ICRA46639.2022.9812107} (2022) &    Learns context-aware latent anchor embeddings that adapt dynamically to the scene, guiding diverse and accurate trajectory predictions. \\ \cline{2-3}
        & \citet{zhou2023query} (2023) & Generates adaptive trajectory anchors via a proposal module, refined with context-specific biases through an anchor-based decoder. \\ \hline
    \end{tabular}
    \end{adjustbox}
\end{table*}

\subsubsection{Goal Anchor-Based Prediction Methods}

Goal anchor-based methods \citep{5354147,8460203,mangalam2020not,zhao2020tnt} focus on predicting a set of feasible goal points that serve as anchors for trajectory generation \citep{wang2023prophnet}. These methods predict potential goals based on the current context and agent behavior \citep{wang2023prophnet, zhou2023query}. Goals are often estimated based on static environmental features, including road topology, lane structures, or sidewalks. In road networks, they can be defined as lane endpoints, intersections, or turn exits, providing a structured basis for trajectory prediction. Once the goals are identified, trajectories are generated by conditioning on these anchors, often through a path-planning or motion-generation module. An advantage of goal anchor-based methods is their natural alignment with goal-driven tasks, such as reaching specific destinations. 

BiTraP \citep{9345445} employs a CVAE with a bi-directional LSTM decoder conditioned on multiple goal hypotheses, improving long-horizon forecasts by mitigating compounding error. SGNet \citep{wang2022stepwise} model estimates goals at multiple temporal scales via a stepwise goal estimator integrated into encoder-decoder architecture. PECNet \citep{mangalam2020not} samples distant endpoints in latent space and uses non-local social pooling and a truncation trick to generate socially compliant long-term trajectories.

\subsubsection{Trajectory Anchor-Based Prediction}

Trajectory anchor-based methods rely on predefined or learned trajectory templates that represent potential future paths \citep{10.1109/ICRA46639.2022.9812107, Chai2019MultiPathMP, yang2023long}. These templates serve as coarse motion representations and are selected or predicted as bases for future trajectories. The anchors can either be directly used or refined further during prediction. Common examples include straight, curved, or turning path templates. TPNet model \citep{9156890} generates feasible future trajectories using polynomial curves. Inspired by the sampling-based paradigm in vehicle motion planning \citep{5547988,werling2010optimal}, PRIME combines planning-based approaches with learning-based prediction by directly searching for reachable paths and generating potential trajectory anchors. This hybrid approach handles complex agent-map interactions while satisfying environmental and kinematic constraints \citep{Song2021LearningTP}, leveraging the strengths of both model-based and learning-based methods for optimal performance in complex scenarios. Next, to enforce constraints during trajectory generation, \citet{li2024planning} proposed the planning-inspired hierarchical (PiH) framework, which employs a hierarchical lateral and longitudinal decomposition to generate path and goal intentions for agents.

\subsection{Summary Discussion and Future Directions}

Planning-conditioned trajectory prediction integrates motion planning concepts into forecasting. Model-based methods offer interpretability and constraint satisfaction but lack adaptability, while learning-based methods handle complex behaviors yet often ignore physical limits. Anchor-based strategies guide predictions toward semantic goals but face scalability challenges in dense scenes. Future work includes combining model-based and learning-based approaches (e.g., PRIME \citep{Song2021LearningTP}), incorporating safety constraints via differentiable control, and exploring multi-agent planning through intent modeling and graph-based reasoning.

An important direction in planning-conditioned trajectory prediction is the anchor generation module, which typically analyzes scene context and/or integrates spatial information to infer potential destinations or paths. However, scene context analysis can be computationally expensive and should be evaluated rigorously in terms of inference time. While map-based anchor generation is often more efficient, it must still be tested for scalability as the number of agents increases, since anchors are generated per agent. The anchor generation module is part of the prediction pipeline and should be evaluated online, rather than relying on pre-generated anchors. Thus, design of efficient and scalable anchor‑generation schemes that can  leverage attention or clustering to learn compact sets and dynamically prune unlikely candidates is a direction for improvement.  
\section{Vision-Language Models for Prediction}
\label{sec:vlm}

\begin{figure*}[h!]
\centering
   \includegraphics[width=1.0\textwidth]{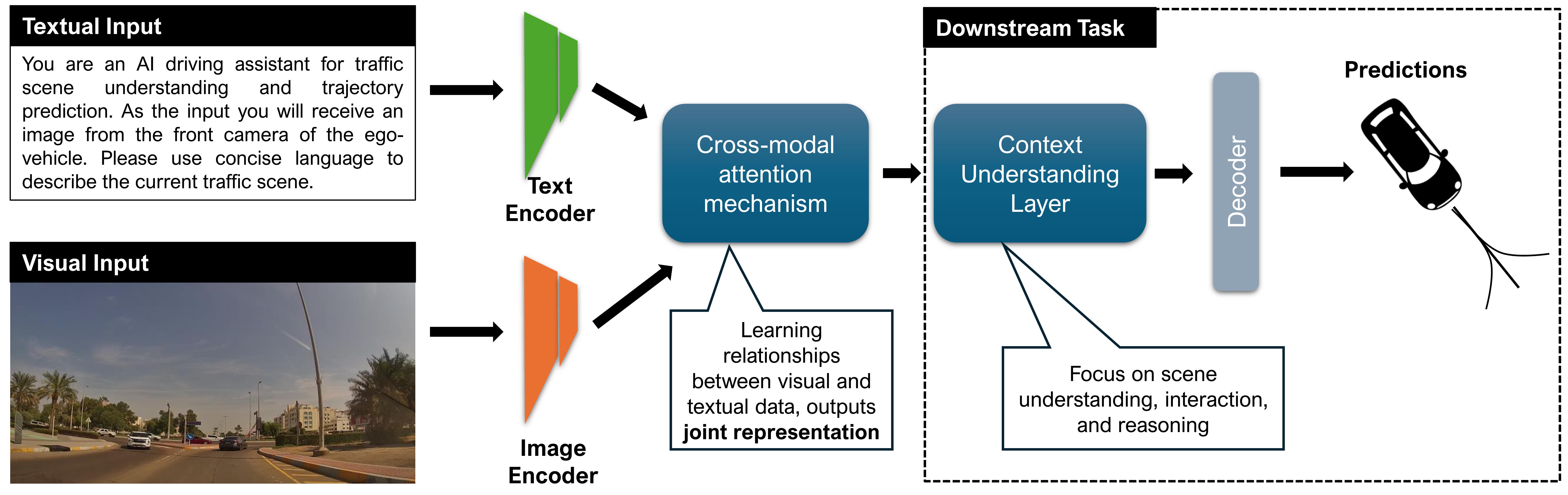}
   \caption{VLM-based trajectory prediction model: encoders for visual and textual inputs, a cross-modal attention mechanism to create a joint representation and context understanding layer tailored for a specific downstream task.}
   \label{fig:vlm}
\end{figure*}

Vision-Language Models (VLMs) \citep{10204586, Liu2023UnsupervisedMP, wu2023referring, wu2023language, peng2023openscene, 10204547, Najibi2023Unsupervised3P, Romero2023ZeldaVA,  jain2022ground, DBLP:journals/corr/abs-2310-02324} are seen as promising technology that could enable vehicles and systems to deeply understand their surroundings, marking a significant advancement in the era of intelligent and explainable transportation. In the context of autonomous driving, VLMs can be roughly categorized into: Multimodal-to-Text (M2T) \citep{Xu2023DriveGPT4IE, qian2024nuscenesqa, fu2023drive, choudhary2023talk2bev} models, which take an image and text as the input and generate text as the output; Multimodal-to-Vision (M2V) \citep{hu2023gaia1, wang2023drivedreamer} models, where the input is an image and text and the output is an image; and Vision-to-Text (V2T) \citep{jin2023adapt, 9306804} models, which take an image as the input and generate text as the output. This section mainly focuses on M2T models in the context of perception and prediction tasks. Readers are encouraged to look into the surveys \citep{zhou2023vision, 10495592} for further details on other VLM configurations and downstream tasks. Figure \ref{fig:vlm} shows the general flow of a VLM-based prediction model. The model processes two main inputs: visual and textual data. While the image encoder processes visual inputs, e.g., single-camera images, multi-camera feeds, or LiDAR-range image, extracting spatial and object-level features; the text encoder interprets contextual information, which can include detailed instructions, scene description, specific object inquiries, navigation commands, etc. The encoded features are then passed to a cross-modal attention mechanism \citep{Cheng2023LanguageGuided3O, 8967929, keysan2023text}, which captures the relationships between visual cues and textual instructions to generate joint representations. This representation is further processed by an autonomous driving task-specific network, e.g., as for the prediction task, by a context understanding layer and decoder. The generated future trajectories consider both visual observations and semantic guidance. 

This section reviews VLM-based models designed for trajectory prediction and traffic scene understanding. Given the limited number of prediction models in the literature, we have included a survey of traffic scene understanding models, considering their potential integration into prediction pipelines. The section also provides a list of available language-enhanced autonomous driving datasets and discusses the challenges associated with integrating VLMs into the autonomous driving stack.

\subsection{Trajectory Prediction}

Table \ref{tab:vlm_tp} summarizes reviewed VLM and LLM-based models for trajectory prediction. \citet{yi2024} proposed a two-stage evaluation framework. In the first stage, a VLM model analyzes images from six camera views and a BEV visualization with labeled agents to rank the prediction difficulty. To improve the model's response accuracy, the VLM is provided with several examples of rankings constructed based on displacement errors from the ground truth. In the second stage, the VLM's output is used to create a subset of training data based on the ranking, resulting in a highly representative and challenging training dataset for training conventional prediction models. \citet{seokha2024} introduced VisionTrap that leverages human gazes and gestures, vehicle turn signals, and road conditions to enhance motion prediction accuracy. The model comprises four main components: a per-agent state encoder that generates agent-specific context features, a visual semantic encoder that encodes images combined with road segments, a text-driven guidance block that guides learning and understanding of detailed visual semantics, and a trajectory decoder. The study includes an inference-time analysis highlighting the model’s suitability for real-time deployment inference constraints. 

\begin{table*}[h]
    \centering
    \caption{Summary of reviewed VLM and LLM-based models for trajectory prediction}
    \label{tab:vlm_tp}
    \begin{adjustbox}{max width=\textwidth}
    \begin{tabular}{|l|c|c|p{14cm}|}
        \hline
        \textbf{Reference} & \textbf{Name} & \textbf{Dataset} & \textbf{Description}  \\
        \hline 
        \citet{keysan2023text} (2023) & - & NuScenes & The prompt describes the target agent and its environment. \\ \hline
        \citet{yi2024} (2024) & - & NuScenes & Uses VLM to create a smaller and representative subset of training data to improve the training efficiency of conventional models. \\ \hline
        \citet{seokha2024} (2024) & VisionTrap & nuScenes-Text &  Utilizes human gazes and gestures, vehicle turn signals, and road conditions to enhance motion prediction accuracy. \\ \hline
        \citet{mingxing2024} (2024) & LC-LLM & highD &  Uses LLM as an explainable lane change prediction model with refined Chain-of-Thought (CoT) reasoning. \\ \hline
        \citet{xiaoji2024} (2024) & TC-Map LLM & WOMD &  Generates Transportation Context Maps as visual prompts and corresponding text prompts to extract situational insights and pass them to a prediction pipeline. \\ \hline
    \end{tabular}
    \end{adjustbox}
\end{table*}

\citet{keysan2023text} expanded the previously proposed CoverNet model \citep{9157523} by incorporating a text-based scene representation. The prompt accepted by the model contains structured information about the target agent and its environment, constructed from polyline and Bézier curve lane representations. \citet{mingxing2024} introduced an explainable lane change prediction model (LC-LLM). The study reformulates the problem of lane change prediction as a language modeling task, processing driving scenario data as natural language prompts. To enhance transparency and prediction reliability, the model’s Chain-of-Thought (CoT) reasoning is refined to include explanatory elements in the prompts during the inference phase. \citet{xiaoji2024} leveraged LLMs to improve the global understanding of traffic context for motion prediction. The study transforms complex traffic scenarios and agents’ past trajectories into visual prompts (Transportation Context Maps) along with corresponding textual prompts. Transportation Context Maps, combined with text prompts, enable the model to generate situational insights, extracting details such as agent intentions and scenario types. The extracted information is then encoded and integrated into a conventional motion prediction pipeline.

\begin{table*}[t]
    \centering
    \caption{Summary of reviewed VLM and LLM-based models for scene understanding}
    \label{tab:vlm_su}
    \begin{adjustbox}{max width=\textwidth}
    \begin{tabular}{|l|c|p{2.5cm}|p{12cm}|}
        \hline
        \textbf{Reference} & \textbf{Name} & \centering \textbf{Dataset} & \textbf{Description}  \\
        \hline 
        \citet{ziang2024} (2024) & VLM-Auto & \centering Custom simulated dataset & Uses Qwen-VL model to analyze visual inputs for identifying environmental conditions and generating control parameters to guide driving behaviour. \\ \hline
        \citet{akshay2024} (2024) & EM-VLM4AD & \centering DriveLM & A lightweight multi-frame model for VQA tasks in autonomous driving. \\ \hline
        \citet{malsha2024} (2024) & - & \centering Waymo PREPER CITY & Evaluates hallucinations in pedestrian detection performed by GPT-4V and LLaVA models. \\ \hline
        \citet{mengjingcheng2024} (2024) & NexusAD & \centering CODA-LM & A framework based on InternVL2-26B for understanding driving corner cases and generate perception and driving recommendations.  \\ \hline
        \citet{yunsong2024} (2025) & ELM & \centering nuScenes Ego4D  & Embodied language model (ELM) to enhance scene comprehension using space-aware pre-training and time-aware token selection.   \\ \hline %313
        \citet{xiaoyu2024} (2024) & DriveVLM & \centering - &  VLM for planning, comprising modules for scene description scene analysis, and hierarchical planning.  \\ \hline
        \citet{junzhou2024} (2024) & - & \centering 100 images from BDD100k & Utilizes the CogVLM model to process sensor data and employs a Chain-of-Thought (CoT) approach to structure reasoning across scene understanding, prediction, and decision-making. \\ \hline % 
    \end{tabular}
    \end{adjustbox}
\end{table*}

\begin{table*}[h]
\caption{Language-Enhanced Autonomous Driving Datasets: types of sensors (Camera, LiDAR), source, year of release and tasks for which datasets are annotated (Single Object Referring (SOR), Action Explanation (AE), Visual Question Answering (VQA), Decision Making (DM), Image Captioning (IC), Visual Spatial Reasoning (VSR), Multi-object Tracking (MOT)).}
\centering
\footnotesize % Increase the font size of the table content
\renewcommand{\arraystretch}{1.2} % Adjusts the space between rows
\begin{adjustbox}{max width=1.0\textwidth}
\begin{tabular}{|p{4cm}|p{5cm}|c|c|c|>{\centering\arraybackslash}p{3cm}|}
\hline
\textbf{Dataset} & \textbf{Source Dataset}  & \textbf{Camera data} & \textbf{LiDAR data} & \textbf{Video} & \textbf{ Tasks}  \\
\hline
Talk2Car \citep{talk2car2020} & nuScenes & \checkmark & - & \checkmark &  SOR \\ \hline
BDD-OIA \citep{bdd-oia2020} & BDD100K  & \checkmark & - & \checkmark  &   AE \\ \hline
NuPrompt \citep{nuprompt2023} & nuScenes & \checkmark  & - & \checkmark &   MOT \\ \hline
Refer-KITTI \citep{refer-kitti2023} & KITTI & \checkmark & - & \checkmark &  MOT \\ \hline
DRAMA \citep{drama2023} & - & \checkmark  & - & \checkmark  & IC, VQA \\ \hline
Talk2BEV \citep{talk2bev2023} & nuScenes & \checkmark & - & \checkmark & VQA \\ \hline
Reason2Drive \citep{reason2drive2023} & nuScenes, Waymo, ONCE & \checkmark & - & \checkmark & VQA \\ \hline
Rank2Tell \citep{rank2tell2023} & -  & \checkmark  & \checkmark & \checkmark & VSR \\ \hline
NuScenes-QA \citep{nuscenes-qa2023} & nuScenes & \checkmark & \checkmark & \checkmark & VQA \\ 
\hline
\end{tabular}
\end{adjustbox}
\label{tab:vlm_datasets}
\end{table*}

\subsection{Traffic Scene Understanding}

Table \ref{tab:vlm_su} summarizes reviewed VLM and LLM-based models for scene understanding. \citet{choudhary2023talk2bev} introduced Talk2BEV --- a VLM designed to handle diverse tasks, including predicting traffic participants' intentions, visual and spatial reasoning, and decision-making based on visual information. \citet{Romero2023ZeldaVA} argue that using VLMs to query images via natural language addresses issues of expressivity and the need for multiple single-purpose query models, but it often returns redundant and low-quality results. To improve this, they introduced Zelda, a system that adds discriminator and synonym terms to refine user queries into natural language prompts, enhancing accuracy by filtering out low-quality frames. \citet{mengjingcheng2024} designed a framework for understanding driving corner cases, built upon the foundation model InternVL2 \citep{chen2024internvl} and consisting of four phases. For visual perception, the framework uses DINO \citep{9709990} for 2D detection and DepthAnything \citep{depth_anything_v1} for depth estimation. This detection information, along with the original image, is processed by a scene-aware retrieval module to generate contextually relevant information. The Driving Prompt Optimization module then uses this information to guide the model, suppressing hallucinations and focusing on critical scene aspects, ultimately producing responses for perception and driving recommendations. \citet{yunsong2024} introduced the Embodied Language Model (ELM) to enhance traditional VLMs' scene comprehension, supporting tasks such as object localization, detailed scene description, prediction, and event memorization. ELM incorporates space-aware pre-training for robust spatial localization and time-aware token selection to effectively process long-term video data and retrieve relevant past events. Finally, \citet{xiaoyu2024} proposed the DriveVLM system for planning in complex environments, with modules for scene description, scene analysis, and hierarchical planning. The scene description module generates linguistic representations of driving conditions and passes these to the scene analysis module, which examines objects' motion states and behaviours that may influence vehicle actions.

\subsection{Benchmark Datasets}

To advance research in VLM-based solutions for autonomous driving, the community is increasingly releasing language-enhanced datasets. Table \ref{tab:vlm_datasets} provides a summary of existing datasets, specifying the source datasets they are built upon and the tasks for which they are suited, such as object tracking, action reasoning, spatial understanding, and question answering in autonomous driving contexts.

Talk2Car dataset \citep{talk2car2020}, designed based on nuScenes dataset, focuses on task of single object referring by pairing natural language commands with specific objects in driving scenes. Similarly, the BDD-OIA dataset \citep{bdd-oia2020}, based on BDD100K, contain  annotations enabling models to interpret the reasoning behind driving actions. Next, the NuPrompt dataset \citep{nuprompt2023} expanded upon nuScenes, enabling models to track multiple objects based on descriptive language prompts. NuScenes-QA \citep{nuscenes-qa2023} introduced question-answer pairs, challenging models to interpret and respond to questions about driving environments. Talk2BEV \citep{talk2bev2023}, another dataset built upon nuScenes, supports VQA with an emphasis on BEV maps, which help models to understand spatial layouts from an overhead perspective. For a broader variety of environments, Reason2Drive \citep{reason2drive2023} integrates data from three datasets nuScenes, Waymo, and ONCE, supporting diverse question-answer sets across varied driving conditions. 

\subsection{Summary and Discussion}

VLMs offer significant potential to enhance trajectory prediction models by incorporating common knowledge and contextual understanding essential for nuanced interpretations of surroundings and environmental factors. This section briefly reviewed foundations of VLM-based models for trajectory prediction, outlined recent advancements in forecasting and scene comprehension models, and provided an overview of current language-embedded datasets for autonomous driving. Despite the notable capabilities that VLMs introduce, seamlessly integrating these models into autonomous driving systems still demands further, in-depth exploration.

VLMs are large-scale models, and the inference time of the VLM-based prediction pipeline is a primary concern when considering integration and computational demands. From an application standpoint, this may necessitate upgrading the vehicle’s onboard compute hardware if inference is conducted locally. Model quantization \citep{nagel2021whitepaperneuralnetwork} is a technique employed to reduce model size, offering faster inference at the cost of some accuracy. Other techniques, such as pruning redundant weights \citep{blalock2020stateneuralnetworkpruning} and knowledge distillation \citep{hinton2015distillingknowledgeneuralnetwork}, serve similar purposes. Further experimentation is needed to explore the trade-off between reasoning capability, model size, and inference speed. Hardware-dependent factors, such as parallelism and processor type (TPU, GPU, or CPU), also play a role in optimizing inference time. If not hosted locally, cloud solutions are a possibility. A set of available VLMs \citep{yao2024minicpmvgpt4vlevelmllm,touvron2023llamaopenefficientfoundation} offer cloud hosting capabilities, implying that an AV would need to request information via an Application Programming Interface (API). This reliance can lead to latency or may require a shift in the type of hardware used onboard, favouring a communication-efficient setup over a computation-heavy station to minimize delays and ensure data transmission with minimal losses.

Another challenge is prompt engineering. It refers to the process of crafting input prompts that guide the model toward specific behaviors or outputs, especially by aligning visual and textual inputs. Providing relevant context within the prompt significantly affects the model's response. For example, including scene-specific cues such as ``heavy traffic ahead" or ``intersection with pedestrians" can influence the model's understanding and downstream reasoning. Prompts can also include instructions on output length or format, helping ensure concise and relevant generation. In some cases, few-shot prompting is used instead, where an example of the expected output is embedded within the prompt. Additionally, prompting strategies such as Chain-of-Thought (CoT) prompting guide the model to break down its reasoning into intermediate steps, enhancing both interpretability and output quality. Efficient prompt design not only improves semantic alignment but also reduces token usage, contributing to lower inference time.

Lastly, the non-deterministic nature of some VLMs \citep{10.5555/3600270.3601993, Li2022BLIPBL}, where the same image-text input may yield different outputs across runs, presents integration challenges in safety-critical scenarios. Strategies such as adjusting the temperature parameter to control output variability, applying majority voting, or averaging across the top-$k$ outputs can help manage this variability. Striking the right balance between consistency and flexibility is essential: enforcing deterministic outputs enhances system stability, while preserving generative flexibility allows the model to better handle edge cases and out-of-distribution scenarios. A rigid deterministic setup may limit adaptability, especially in complex environments. Consequently, the choice between generative and deterministic use of VLMs must be carefully evaluated in the context of trajectory prediction. Although VLMs show promise for enhancing situational understanding and predictive reasoning, their optimal configuration within the autonomous driving stack remains an on-going research investigation.
\section{Collaboration between AVs for Prediction}
\label{sec:collaboration}

Despite advancements in multimodal sensing, each sensor still has fundamental limitations, such as restricted range and the risk of miscalibration. Moreover, even for human drivers, occlusions are a common cause of accidents; for autonomous vehicles (AVs), they present an even greater challenge, as AVs lack human intuition and their perception of the environment relies entirely on sensors. When the limitations of an individual sensor suite are reached, collaboration between vehicles can compensate by aggregating information from multiple viewpoints. The current SOTA work in this area is primarily focused on collaborative perception, where sensor data from multiple vehicles is fused to construct a holistic view of the environment. This fused representation then supports downstream tasks such as object detection \citep{chen2023co, Li2021LearningDC}, tracking \citep{9676458}, trajectory prediction \citep{10.1007/978-3-030-58536-5_36}, planning \citep{Cui2022CoopernautED, glaser2023communicationcritical}, and task-invariant modules \citep{li2022multirobot}.

A major research focus has been the design of efficient fusion architectures to enhance the accuracy of a down-stream task. Based on the stage at which information is integrated, existing models are commonly classified into early, intermediate, and late fusion. Each mode imposes different bandwidth requirements and communication overhead. Communication redundancy has thus become a central research direction, with methods ranging from sparse intermediate fusion to selective sharing of complementary information. These strategies aim to reduce bandwidth consumption while preserving critical context that complements ego-vehicle observations. Other lines of work address bandwidth optimization \citep{9197364, Li2021LearningDC, Hu2022Where2commCC, Chen_2023_ICCV}, mitigation of communication delays \citep{10.1007/978-3-031-19824-3_19, yu2023flowbased}, and handling pose misalignment \citep{pmlr-v155-vadivelu21a, 10.1145/3581783.3611880, Zhang_2024_CVPR}. Additionally, due to different sensor configurations across vehicles, recent models aim to learn in a unified fusion space \citep{10377310, xiang2023div2x, ma2023macp, xia2025plentypolymorphicfeatureinterpreter}, addressing both heterogeneous sensor suites and varying perception architectures.

This section reviews existing collaborative models, surveys available multi-vehicle datasets, and discusses current limitations along with future research directions. It is important to note that the categorization here is based on fusion strategies, rather than the collaboration type (e.g., vehicle-to-vehicle or vehicle-to-infrastructure). For a broader overview of inter-vehicle collaboration challenges, we refer readers to recent comprehensive surveys \citep{hu2024collaborative, huang2023v2x, liu2023vehicletoeverything, Han2023CollaborativePI, bai2022survey}.

\subsection{Type of Fusion}

\begin{figure*}[h!] \centering \includegraphics[width=1.0\textwidth]{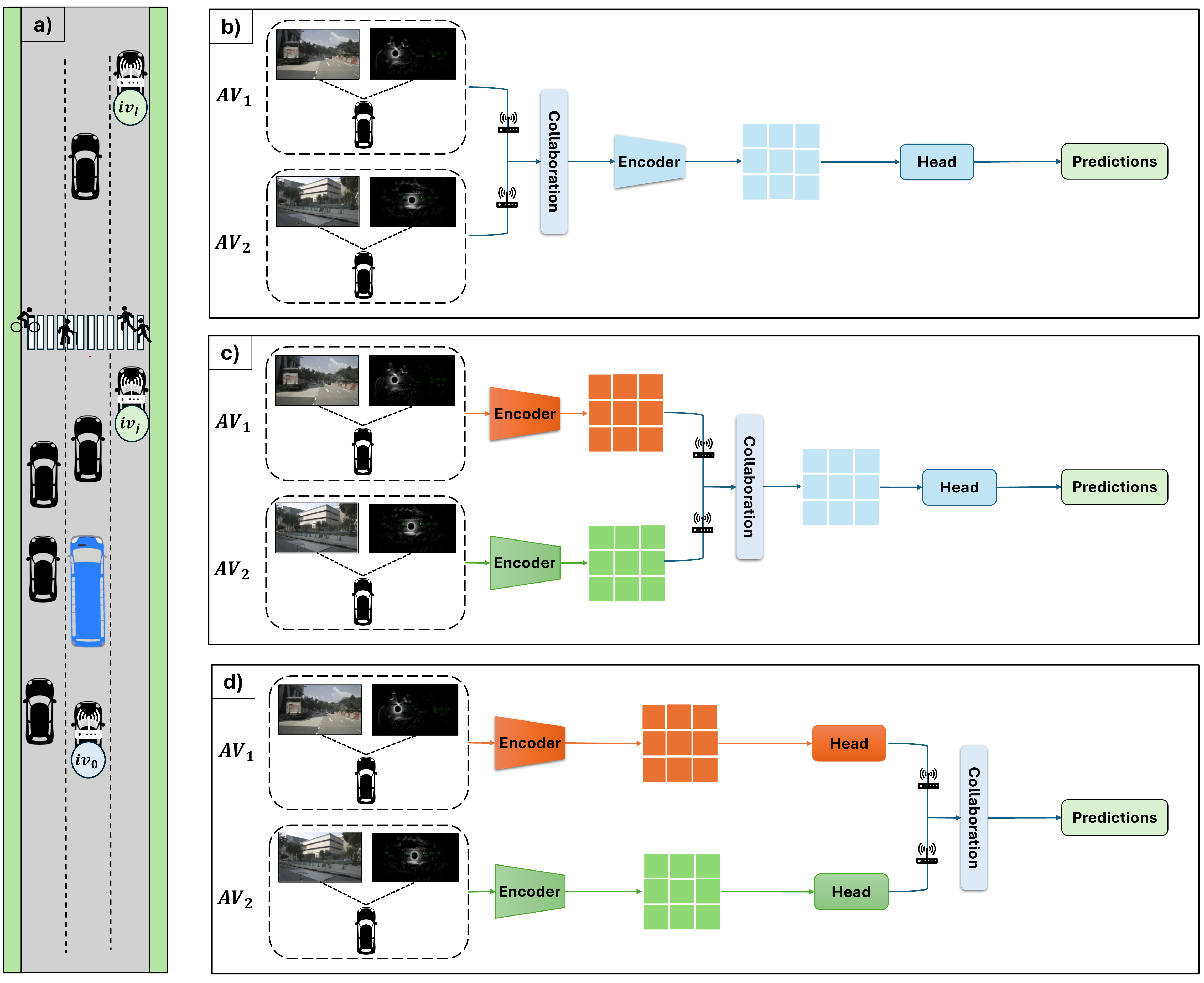} \caption{Collaboration between AVs: (a) - scenario for collaborative perception, vehicle $iv_j$ can pass information about vulnerable traffic agents, such as pedestrians at a crossing, to vehicle $iv_0$. Additionally, both vehicles can refine each other’s predictions by sharing and complementing observations perceived from different angles. (b, c, d) schematically show different collaboration schemes: (b) - early collaboration, (c) - intermediate collaboration, and (d) - late collaboration.} \label{fig:col_fusion} \end{figure*}

Collaborative between vehicles can be achieved by combining data and features at different stages. Currently, the literature distinguishes three fusion modes: early, intermediate, and late fusion (please refer to Figure \ref{fig:col_fusion}).

In \textbf{\textit{early fusion}}, raw data is combined at the beginning of the perception pipeline, allowing the ego vehicle to utilize complementary sensor information to mitigate occlusions. \citet{chen2023co} introduced an unsupervised 3D representation learning approach, CO3, which uses point clouds from both vehicle and infrastructure side views to generate contrasting and semantically aligned representations. Since transmitting raw data demands considerable bandwidth, \citet{Li2021LearningDC} introduced a distilled collaboration graph (DiscoGraph), leveraging a teacher-student training methodology. The teacher model integrates early collaboration with comprehensive inputs, while the student model focuses on partial inputs at intermediate stages of collaboration. Building on the combination of early and intermediate fusion, \citet{li2025v2xdgdomaingeneralizationvehicletoeverything} investigated domain generalization in collaborative perception and introduced Cooperative Mixup Augmentation for Generalization (CMAG) to enhance model generalization by emulating unseen collaborative scenarios.

In \textbf{\textit{intermediate fusion}} \citep{huang2025v2xrcooperativelidar4dradar,9636761}, deep semantic features are shared, which are further integrated by the ego-vehicle for the downstream task. Compared to early fusion, intermediate fusion involves more information loss and requires optimal feature selection and fusion strategies. \citet{10.1007/978-3-031-19842-7_7} proposed V2X-ViT, a vehicle-to-infrastructure cooperative perception system based on a Vision Transformer model. This model employs heterogeneous multi-agent self-attention and multi-scale window self-attention to integrate data, addressing asynchronous data exchange and positioning inaccuracies. \citet{9676458} introduced a GNN-based framework for multi-robot perception, evaluated under challenging conditions like camera occlusion and heavy noise. \citet{Xu2022CoBEVTCB} developed CoBEVT, a simple yet high-performing framework for collaborative BEV map creation. Its fused axial attention module (FAX) integrates camera features from various viewpoints and agents, capturing both dense local and broad spatial relationships. \citet{wei2023asynchronyrobust} introduced CoBEVFlow, a robust framework for collaborative perception using a BEV flow approach to manage non-uniform collaboration messages and transmit perceptual data without generating new features. \citet{yang2023spatiotemporal} proposed SCOPE, a framework designed to unify spatio-temporal awareness across agents for effective information exchange. Wang et al. \citet{10377874} introduced CORE, a bandwidth-efficient framework for multi-agent cooperative perception that combines a compression mechanism, attention-based integration of messages, and a reconstruction module. Lastly, \citet{10030450} developed an adaptive spatial-wise feature fusion model with a trainable feature selection module for efficient information aggregation.

In \textbf{\textit{late fusion}}, the decisions are generated independently by each vehicle and further shared. The ego-vehicle aggregates these outputs to refine its own predictions. While this mode is bandwidth-efficient, it is susceptible to noise and incomplete information in the individual predictions. CoAlign \citep{lu2023robust} proposes a hybrid collaboration framework that combines intermediate and late fusion, mitigating pose estimation inaccuracies by constructing an agent-object pose graph for improved alignment and integrating features at multiple spatial resolutions. \citet{yu2022dairv2x} introduced the DAIR-V2X dataset along with the benchmark late fusion model VIC3D, which compensates for temporal asynchrony by estimating object motion from consecutive frames and projecting states to the vehicle's current time before fusing the adjusted predicted boxes.

\subsection{Communication Redundancy} 
Mitigating communication redundancy has been explored in two main directions: (1) determining which information should be shared by surrounding agents to complement the ego vehicle’s observations \citep{9197364, 10378239, 10.1145/3581783.3611699, yang2023howcomm, xu2025cosdhcommunicationefficientcollaborativeperception}, and (2) designing efficient compression mechanisms to minimize message sizes while preserving essential information \citep{Hu2022Where2commCC, Chen_2023_ICCV}. The first line of research typically follows a standard pipeline: the ego agent generates and broadcasts a request message, surrounding agents evaluate matching scores or generate feature maps based on both their local observations and the received request, and then transmit the selected responses. The resulting messages, when aggregated by the ego vehicle, offer complementary information that enhances perception. In the second category, \citet{Hu2022Where2commCC} introduced a spatial confidence map that enables surrounding agents to transmit only sparse, high-confidence features, thus improving efficiency while maintaining perception quality. \citet{Chen_2023_ICCV} proposed TransIFF, a framework that filters out low-confidence predictions to reduce message size and includes a Cross-Domain Adaptation module for aligning features, as well as a Feature Magnet (FM) module to dynamically fuse instance-level features. \citet{hu2024communicationefficientcollaborativeperceptioninformation} presented a two-stage communication reduction method: first, applying a codebook-based message representation for compression, and second, using an information-filling strategy to prioritize the most informative content for transmission. Along the same lines, \citet{ding2025point} proposed a compact object-level representation that preserves both semantic and structural information through a learnable sampling strategy.

\subsection{Transmission Delays} 
Transmission latency is another challenge that arises when vehicles collaborate. Even under the assumption that all collaborators share the same parameters, such as FPS and broadcasting frequencies, delays are inevitable, since no information share happens instantly. The delay issue further complicates data alignment, which is already affected by inaccuracies in localization systems and transformation processes. \citet{10.1007/978-3-031-19824-3_19} proposed a latency-aware collaborative perception framework with a synchronization component called SyncNet, which uses attention mechanisms and temporal adjustment strategies to synchronize data into a unified timeframe. \citet{yu2023flowbased} introduced the Feature Flow Net (FFNet), which leverages temporal consistency from consecutive frames captured by infrastructure cameras to derive a feature flow. This flow serves as a predictive function to describe feature changes over time, generating features aligned with the current timeframe of ego-vehicles. \citet{xu2025codyntrustrobustasynchronouscollaborative} proposed a cooperative framework in which BEV features are extracted with uncertainty quantification and aligned by warping them using motion displacement to match the ego-agent’s timestamp.

\subsection{Unified Fusion Space} 
The literature explores challenges arising from variations in sensor suites, model architectures, and the integration of collaborative components within existing systems. One category of studies proposes collaboration through a unified fusion space \citep{10377310, xiang2023div2x, ma2023macp, xia2025plentypolymorphicfeatureinterpreter}, addressing variations in sensor suites and prediction module architectures.  \citet{xu2023bridging} addressed discrepancies in neural network architectures across collaborating agents by proposing a trainable feature resizer to harmonize feature dimensions and a sparse cross-domain transformer for domain adaptation. In the related study \citep{xu2023modelagnostic}, the authors introduced a confidence calibrator to adjust for biases in prediction confidence scores across agents. \citet{10377310} tackled the challenge of heterogeneous sensor modalities with a model for hetero-modal cooperative perception, enabling collaboration among agents using different sensor suites. By removing the assumption of uniform sensor setups, the model supports new, diverse agent types in collaborative perception. \citet{Lu2024AnEF} proposed the HEAL framework, which constructs a unified feature space using a pyramid fusion network and aligns new agent types to this space via backward alignment. \citet{xiang2023div2x} addressed discrepancies in LiDAR data with a model comprising three components: a Domain-Mixing Instance Augmentation module to ensure uniform data representation, a Progressive Domain-Invariant Distillation module for domain-neutral feature adaptation, and a Domain-Adaptive Fusion module with calibration-aware attention mechanisms. Finally, \citet{ma2023macp} introduced the MACP framework, equipping single-agent models with cooperative capabilities by freezing the existing prediction model and integrating trainable lightweight modules.

\subsection{Benchmark Datasets}
Collaborative perception is advancing significantly with the release of new datasets that replicate complex urban environments and diverse driving scenarios, supporting the development of algorithms across multiple applications. Table \ref{tab:collab_datasets} outlines the details of available simulated and real-world datasets. This summary includes datasets annotated not only for motion forecasting but also for tracking, which can be adapted for use in prediction tasks.

\begin{table*}[h]
\caption{Large-scale Collaborative Perception Datasets: types of sensors (Camera, LiDAR), number of agents interacting, types of communication (V2V - vehicle to vehilce, V2I - vehicle to infrastructure), number of frames and tasks for which the dataset are annotated (Object Detection (OD), Object Tracking (OT), Semantic Segmentation (SS), Trajectory Prediction (TP), Motion Prediction (MP), Domain Adaptation (DA)).}
\centering
\footnotesize
\renewcommand{\arraystretch}{1.2} % Adjusts the space between rows
\begin{adjustbox}{max width=1.0\textwidth}
\begin{tabular}{|p{3cm}|p{3cm}|c|c|c|c|c|c|p{3cm}|}
\hline
\textbf{Dataset} & \textbf{Source} & \textbf{Frame} & \textbf{V2V} & \textbf{V2I} & \textbf{Camera} & \textbf{LiDAR} & \textbf{Agents} & \textbf{Tasks} \\ \hline
V2V-Sim \citep{10.1007/978-3-030-58536-5_36} & Simulation & 51K & $\checkmark$ & - & - & $\checkmark$ & 1-7 & OD, MP\\
\hline 
V2X-Sim \citep{d2} & Simulation & 10K & $\checkmark$ & $\checkmark$ & $\checkmark$ & $\checkmark$ & 1-5 & OD, SS, OT \\
\hline 
OPV2V \citep{d3} & Simulation & 11K & $\checkmark$ & - & $\checkmark$ & $\checkmark$  & 1-7 & OD, OT, MP\\
\hline 
LUCOOP \citep{10186693} & Real & 54K & $\checkmark$ &  - & - & $\checkmark$  & 3 & OD, OT \\ \hline
V2V4Real \citep{d7} & Real & 20K & $\checkmark$ &  - & $\checkmark$ & $\checkmark$  & 2 & OD, OT, DA \\ \hline
V2X-Seq \citep{d8} & Real & 15K & - & $\checkmark$ & $\checkmark$ & $\checkmark$  & 2 & OD, OT, TP \\ \hline
DeepAccident \citep{d9} & Simulation & 57K & $\checkmark$ & $\checkmark$ & $\checkmark$ & $\checkmark$ & 1-5 &  OD, OT, SS, MP, DA\\ \hline 
TumTraf-V2X \citep{zimmer2024tumtraf} & Real & 7.5K & - & $\checkmark$ & $\checkmark$ & $\checkmark$ & 2 & OD, OT \\
\hline
\end{tabular}
\end{adjustbox}
\label{tab:collab_datasets}
\end{table*}

\begin{table*}[h]
    \centering
    \caption{SOTA Collaborative Prediction Models.}
    \footnotesize
    \begin{adjustbox}{max width=1.0\textwidth}
    \begin{tabular}{|p{5cm}|c|c|c|c|c|}
    \hline
        \textbf{Model} & \textbf{Venue} & \textbf{Dataset}  & \textbf{Task}  & \textbf{Input} & \textbf{Results} \\ \hline
        V2VNet \citet{10.1007/978-3-030-58536-5_36} (2020) & ECCV &  V2V-Sim & \makecell{Detection, \\ Forecasting} & Point Cloud & \makecell{-} \\ \hline
        V2VNet \citet{pmlr-v155-vadivelu21a} (2021) & CoRL &  V2V-Sim & \makecell{Detection, \\Forecasting} & Point Cloud & \makecell{-} \\ \hline
        V2X-Graph \citet{ruan2023learning} (2023) & - & \makecell{DAIR-V2X-Traj}  & Forecasting & Vector Map & \makecell{-} \\ \hline
         V2X-Graph \citet{ruan2023learning} (2023) & CVPR & \makecell{V2X-Seq}  & Forecasting & Point Cloud & minADE : 1.17,  minFDE : 2.03, MR : 0.27  \\ \hline
        V2XFormer \citet{d9} (2025) & AAAI & DeepAccident  & \makecell{Detection, \\ Motion Prediction} & RGB Image & mIOU: 56.2, VPQ: 44 \\ \hline
        CMP \citet{wu2024cmp} (2024) & - & OPV2V  & Motion Prediction & Point Cloud & MinADE\_6@5s: 1.74, MinFDE\_6@5s: 4.05 \\ \hline
    \end{tabular}
    \end{adjustbox}
    \label{tab:sota_prediction}
\end{table*}

In simulated datasets, V2X-Sim 1.0 \citep{10.1007/978-3-030-58536-5_36} is a V2V dataset created using CARLA and SUMO, focusing on dense traffic scenarios and using 32-channel LiDAR for 3D perception. V2X-Sim 2.0 \citep{d2} extends this dataset by adding V2I elements and enhancing vehicle sensor setups. OPV2V \citep{d3} broadens the scope with varied road types and detailed vehicle annotations, supporting collaborative tasks such as object detection, tracking, and motion forecasting. DeepAccident \citep{d9} focuses on diverse collision scenarios in autonomous driving, with multi-modal data from vehicles and infrastructure, realistic accident simulations, and support for both perception and prediction tasks.

Among real-world datasets, LUCOOP \citep{10186693} is a multi-vehicle dataset aimed at urban navigation, object detection, and connected autonomous vehicle research. It includes sensor data from multiple vehicles over a 4 km route, designed to simulate cooperative traffic using both static and dynamic sensors in an urban setting. The V2V4Real dataset \citep{d7}, designed for V2V perception, covers various driving scenarios with annotated 3D bounding boxes and high-definition maps, collected over extensive urban and highway routes in Columbus, Ohio. This dataset includes five object classes and supports detection, tracking, and behavior prediction. V2X-Seq \citep{d8} explores the impact of infrastructure on perception and trajectory forecasting with 15,000 frames from vehicle and infrastructure perspectives across diverse urban scenarios. Lastly, TUMTraf V2X \citep{zimmer2024tumtraf} captures complex traffic scenarios in day and night conditions, with nine sensors providing 360-degree views at intersections to address cooperative perception challenges like pose estimation errors, latency, and synchronization, making it highly applicable for real-world use cases.

\subsection{Collaborative Trajectory Prediction Models}
Table \ref{tab:sota_prediction} summarizes SOTA collaborative prediction models detailing the type of fusion and dataset on which the models were evaluated. V2VNet \citep{10.1007/978-3-030-58536-5_36} employs intermediate fusion, using a CNN block to process raw sensor inputs into a compressible intermediate representation. A cross-vehicle integration mechanism then fuses these intermediate outputs for collaborative prediction. In BEV-centered fusion, V2XFormer \citep{d9} utilizes a Swin Transformer to generate BEV features, which are spatially aligned to the ego-vehicle’s coordinate system. These aligned features are further fused using CoBEVT \citep{Xu2022CoBEVTCB} to produce aggregated representations that are subsequently decoded into motion predictions. Leveraging graph neural networks (GNNs), V2X-Graph \citep{ruan2023learning} constructs a graph from observed trajectories and vector maps, representing agents’ motion features and map elements as nodes, with spatial-temporal relationships as edges. It employs three interpretable sub-graphs: the Agent-Lane sub-graph (ALG) for map-based interactions, the Motion Fusion sub-graph (MFG) for motion features, and the Cooperative Interaction sub-graph (CIG) for inter-agent relationships. These sub-graphs aggregate heterogeneous features across views, which are then decoded into future trajectories. Next, a hybrid fusion prediction model, the CMP model \citep{wu2024cmp}, enhances trajectory prediction through two stages of fusion. In the perception stage, it performs intermediate fusion by using the FuseBEVT module \citep{Xu2022CoBEVTCB} to improve object detection. In the prediction stage, it applies late fusion, refining predictions through a multi-head attention mechanism. Lastly, Co-MTP \citep{zhang2025comtpcooperativetrajectoryprediction} constructs a heterogeneous graph by integrating past and future trajectories from both ego and infrastructure perspectives. It applies a graph transformer for intermediate fusion in the history domain to enhance past features, and leverages future trajectories to model interactions that may influence ego planning.

\subsection{Summary and Discussion}

Collaboration between vehicles can help overcome key limitations in individual perception systems, such as restricted sensing range, occlusions in dense traffic, and sensor degradation. This section reviewed different fusion strategies, highlighted SOTA collaborative prediction models, and summarized available multi-vehicle datasets.

Currently, collaboration is mainly explored at the perception level, which poses significant challenges due to high bandwidth requirements. While constructing a unified global view and decoding it downstream is a well-justified approach, shifting fusion to task-specific stages, such as trajectory prediction, may offer more flexibility and compliance with existing communication protocols requirements. Prediction models typically rely on structured, low-dimensional inputs (e.g., past trajectories, HD maps, and interaction features), rather then dense feature maps shared in collaborative perception. By moving fusion to the prediction stage, collaboration can become both semantically more relevant to the trajectory forecasting task and communication-efficient. This direction remains underexplored.

Next, most existing frameworks assume a fully connected network where the ego vehicle receives data from all collaborators. However, practical deployments must account for varying communication topologies, transmission frequencies, and delays. In trajectory prediction, even a half second delay may render shared data obsolete. Thus, addressing time sensitivity and synchronization is critical. Another limitation is the lack of inference time analysis in current models. Prediction must be delivered within tight time windows to be actionable for planning. Transmitting and fusing high-dimensional features may hinder latency requirements. Additionally, verifying the trustworthiness of shared information is crucial for accuracy. A vehicle with degraded sensors can unintentionally introduce errors into the system. Fusion mechanisms must incorporate verification stages to assess the reliability of incoming data and prevent the propagation of low-quality inputs.

Finally, the development of standardized testing platforms is essential for prototyping and benchmarking collaborative algorithms. OpenCDA \citep{xu2023opencdaopensourceecosystemcooperative} is one such open-source ecosystem, offering tools for cooperative detection and automation testing. Expanding such platforms to cover V2I collaboration, communication robustness, and scenario diversity will be important for advancing the field.
\section{Discussion}
\label{sec:discussion}

In recent years, both academic and industrial research communities have made significant progress in advancing the field of trajectory prediction. To summarize these advancements, this section revisits the research questions posed at the beginning of the survey, elaborates on the remaining open challenges, and outlines potential future research directions.

\subsection{On Research Question 1}
\label{subsec:q1}

\textit{What are the existing solutions for trajectory prediction?}

Section \ref{sec:survey} showed that trajectory prediction realm has evolved through several distinct modeling phases, starting from physics-based methods and advancing to deep learning hybrid architectures. Existing solutions span classical machine learning, recurrent networks, generative models, graph-based architectures, attention mechanisms, and hybrid methods. Early deep learning approaches relied on recurrent models to capture temporal dependencies in sequential data. The next phase focused more on dynamic motion patterns, often using hierarchical or multi-layer structures. As the field matured, modeling complex patterns and agent interactions led to the emergence of attention mechanisms and graph-based methods.

Current SOTA models commonly integrate multiple architectures to handle the inherent complexity of trajectory prediction, e.g., using CNNs for scene understanding, attention for feature weighting, and encoder–decoder structures for trajectory generation. Examples of such combinations include graph-enhanced Transformers and CNN–LSTM pipelines. In parallel, hybrid approaches that fuse physics-based modeling with deep learning are increasingly explored, integrating domain knowledge with data-driven flexibility. Models that embed kinematic constraints within neural networks enhance prediction accuracy while preserving physical plausibility.

\subsection{On Research Question 2}
\label{subsec:q2}

\textit{Is the trajectory prediction problem solved, or is there still room for improvement?}

Despite the extensive range of methods and increasingly refined solutions, the rate of improvement in prediction accuracy appears to be diminishing. As shown in Section \ref{subsec:datasets}, benchmark leaderboards for some SOTA datasets reveal only marginal performance gains between top-performing models. For example, on the Waymo Motion Forecasting dataset, the difference in minADE between the top two solutions is just 0.01 (MTR\_v3: 0.55 vs. ModeSeq: 0.56). In real-world applications, extreme precision at the millimeter level does not necessarily guarantee the effectiveness of prediction modules in all scenarios. Human drivers, though less precise, navigate complex environments successfully. While current models demonstrate high averaged accuracy on dataset scenarios, failures in specific edge cases may lead to severe consequences.

Thus, from a deployment readiness perspective, several gaps remain. Most studies lack qualitative evaluations that reveal where models fail, since metrics are typically averaged, making it difficult to extract insights about edge cases. If prediction instability occurs at certain timesteps, introducing a consistency mechanism becomes a relevant research direction. For underrepresented patterns and scenarios, more studies on cross-dataset evaluations are needed, i.e., training on one dataset and testing on another, to assess performance in out-of-distribution scenarios, which remains largely unexplored.

\subsection{On Research Question 3}
\label{subsec:q3}

\textit{What are the emerging research trends in trajectory forecasting realm?}

Recent research trends in the prediction domain focus on enhancing the robustness and generalization of prediction models. As discussed in Section \ref{sec:uncertainty}, quantifying and mitigating uncertainty at every stage of autonomous stack is an active research area. One approach to mitigate aleatoric uncertainty is through collaborative perception, presented in Section \ref{sec:collaboration}, which compensates for individual perception system limitations through constructing collaborative holistic view of the environment. Planning-conditioned trajectory prediction methods, presented in Section \ref{sec:planning-driven}, represent a promising direction by incorporating motion planning principles into prediction frameworks. By modeling planned trajectories and potential destinations, planning-conditioned approaches provide interpretable and constraint-compliant predictions, particularly in interactive and complex scenarios.

Additionally, Vision-Language Models (VLMs), as discussed in Section \ref{sec:vlm}, offer common knowledge and contextual understanding necessary for navigating complex and uncertain environments. While utilizing VLMs is a highly emerging and active research area, and holds promise as anchors in uncertain conditions, their real-world application in autonomous driving in terms of latency and predictability is still under investigation. Practical challenges, such as hosting these large models on the cloud, can introduce latency issues that may compromise real-time predictions. The integration of VLMs into the autonomous driving stack requires further exploration to determine their effectiveness and deployment configurations.

These trends reflect a shift in trajectory prediction research towards developing robust, context-aware, and planning-informed methods for operation in dynamic environments.

\subsection{Future Directions}
\label{subsec:future_directions}

This section outlines the existing research gaps and proposes future directions for further investigation, based on the comprehensive review and discussion presented.

\begin{enumerate}[I)]

\item \textbf{Real-time Inference}. 

Accurate trajectory prediction in complex traffic scenarios requires semantic understanding of the environment and interpretation of agent behavior. Incorporating advanced reasoning layers to achieve this increases inference time. \citet{lin2018architectural} emphasize that autonomous vehicle systems must process and react to traffic conditions faster than human drivers to ensure safety. While the fastest human reaction times range from 100–150 ms \citep{newell1985prospects, thorpe1996speed}, the safe response time for autonomous systems should not exceed 100 ms. This sets the upper limit and defines the latency constraints for the entire perception pipeline, including prediction systems. Yet, much of the current research prioritizes accuracy and lacks inference-time analysis, creating a gap between theoretical advances and practical deployment. Real-world integration calls for design of predictors that balance precision and speed, supplied with inference-time testing. \\

\item \textbf{Hardware Compatibility.} 

Prediction model selection should be guided by the hardware constraints of the deployment platform and the operational requirements of the system. In many scenarios, a lighter and faster model, like architectures that can be parallelized or hardware-accelerated, may offer lower latency and be more practical than a slower, higher-accuracy alternative. For example, parallelizable models with optimized embeddings, as shown in \citet{giuliari2020transformer, Murad}, have demonstrated improved inference speed without compromising performance. While such models may rank lower on benchmark leaderboards, their ability to provide rapid updates allows for more frequent error correction and faster system responses. However, current research primarily focuses on maximizing accuracy, often overlooking evaluation on edge devices or analysis of deployment feasibility under varying resource constraints. This gap hinders the transition from high-performance prototypes to real deployable systems. \\

\item \textbf{Systematic Evaluation and Benchmarking.}

A major gap in motion prediction research is the lack of standardized, cross-dataset evaluation protocols. Most models are assessed using ADE and FDE on isolated datasets, each with its own format, API, and leaderboard, making fair comparison difficult. However, trajectory prediction is fundamentally about learning motion patterns, and a model trained on one dataset should ideally transfer to others under a unified evaluation framework. While some benchmarks \citep{10.1145/3447548.3467236, li2023scenarionet} support large-scale cross-dataset training, such efforts remain limited in scope. The main limitation is a unified parametrization of traffic scenarios \citep{li2023scenarionet} that captures interpretable factors such as road geometry, interaction density, and rare maneuvers. Without this, it is difficult to isolate failure cases, evaluate the generalization capacity of models, or identify where contextual inputs like maps or multi-agent cues are truly beneficial. A unified framework for scenario characterization, motion pattern clustering, and rare behavior sampling is needed to support robust, transferable, and reproducible model evaluation. \\

\item \textbf{Integration with Perception.}

For integration of perception and prediction, the review briefly discussed both the detect-track-predict and joint learning paradigms. The most common paradigm, detect-track-predict, propagates detection and tracking errors into the prediction stage. However, most prediction models assume clean, error-free trajectories during inference, which is far from real deployment conditions. A key research gap lies in evaluating model robustness under noisy perception and developing training strategies that allow models to generalize to imperfect inputs. Additionally, prediction-informed perception \citep{li2022modar, heo2023p2d}, where future trajectories are used to guide perception, and prediction-informed tracking \citep{10204316}, where forecasts improve association under occlusion, also require further exploration. These approaches represent early steps toward more integrated perception–prediction pipelines but remain underexplored. \\

\item \textbf{Prediction for Multiple Horizons}.

Most trajectory prediction models focus on single horizon predictions. Short-term predictions (typically up to 1, 1.5 or 2 seconds) --- are essential for handling abrupt events such as sudden braking, sharp lane changes, or obstacle emergence. Errors in this window can escalate quickly, so precise, low-latency forecasts are critical for accounting for random behaviors and maintaining vehicle safety and stability. In contrast, long-horizon predictions ($>$ 2 seconds) guide strategic decisions like merging or route planning but suffer from growing uncertainty over time. Recent approaches like DEMO \citep{WANG2025102924} adopt a multi-horizon structure that fuses dynamics-informed short-term forecasting with interaction-aware long-term reasoning. These multi-scale predictions can be jointly consumed by planning modules for real-time refinement. \\

\item \textbf{Mitigating Epistemic Uncertainty}. 

Handling out-of-distribution scenarios remains an active research area. One promising direction involves a dual-process strategy: using predefined, optimized algorithms in familiar scenarios while leveraging the generalized reasoning capabilities of Vision-Language Models (VLMs) in unfamiliar or uncertain situations. Another approach focuses on simulation-to-reality (Sim2Real) adaptation. Sim2Real techniques involve generating synthetic scenes that mimic real-world conditions, enabling models to learn in controllable environments where varying levels of diversity can be introduced before deployment. Additionally, continual learning during trial deployments or pre-deployment phases can help tailor models to local conditions. A small dataset specific to the target city may be collected to fine-tune the model while preserving previously learned knowledge. A set of works in the literature explore this direction \citep{YANG2022110022, Mei2024ContinuouslyLA, Chen2023TrajMAEMA} and continues to be an active area of research. A detailed overview on this topic is provided in \citep{wang2025deployablegeneralizablemotionprediction}; we refer the reader to the generalization section before pursuing further research in this direction. \\

\item \textbf{Heterogeneous Trajectory Prediction}.

Autonomous vehicles must predict trajectories for diverse agent categories, including vehicles of various sizes (e.g., trucks, vans, sports cars), pedestrians, bicycles, and motorcycles. Each agent type exhibits distinct kinematic properties (except pedestrians), typical speeds, and acceleration patterns, which significantly influence their motion behavior. Vehicle trajectories are constrained by factors such as steering limits and acceleration bounds, whereas pedestrian motion is more unconstrained and inherently stochastic. A key challenge is developing a unified prediction framework capable of accommodating these heterogeneous characteristics. One approach involves using a single model trained on large, diverse datasets, with agent category labels and relevant priors provided as inputs, or following a hierarchical design where the model first classifies the agent type and then delegates prediction to category-specific modules. This setup requires carefully curated training data to ensure balanced learning across categories. Alternatively, using category-specific predictors can improve specialization but increases inference time and system complexity.  \\

\item \textbf{Pedestrian Individual and Group-aware Trajectory Prediction}.

Group-aware predictions require modeling group formations, relationships, and dynamics to effectively capture collective pedestrian motion patterns. Prior studies have emphasized the significance of these factors in understanding group behavior \citep{10.1007/978-3-319-46484-8_33, Yoo2016VisualPP, Alahi2014SociallyAwareLC, 6907734, Zhou2015, Shi2019-ut, PEI2019273, 9578212, liao2023batbehaviorawarehumanliketrajectory, kothari2021humantrajectoryforecastingcrowds}. A promising direction lies in developing hybrid models that integrate crowd-level modeling with individual trajectory prediction, especially in dense urban scenes. If the model can accurately detect group formations and estimate collective motion, it may reduce computational costs by avoiding per-agent predictions within the crowd. For pedestrians outside group formations or leaving the crowd, the system should switch to individual-level predictions. This approach requires further research, particularly in building diverse and representative crowd datasets tailored to autonomous driving contexts.  \\

\item \textbf{Integration with Planning.}

The output of the prediction module is passed to the planning block in the autonomous driving stack. However, these two components are often developed by separate communities, leaving the integration aspect \citep{10821596} relatively underexplored. Key integration challenges include suitable predictor's output format (e.g., occupancy grid is least challenging for integration), quantifying uncertainty, and determining the level of prediction error the planner can tolerate. Ultimately, the goal of prediction is to support safe planning through collision avoidance, thus, tighter integration with planning modules can provide valuable insight into performance requirements and system robustness. To enable smoother interaction, planners must be aware of the uncertainty in predicted trajectories. One effective approach is to use multimodal predictions, which present several possible future outcomes along with their associated probabilities. This allows the planner to consider alternative scenarios and make more informed, flexible decisions when selecting a safe course of action. \\

\item \textbf{Collaboration between AVs}. 

To overcome limitations in individual sensor suites, collaboration between vehicles enables the construction of more comprehensive scene representations from complementary viewpoints. One promising research direction is task-specific collaboration, where context-aware fusion is performed using sparse, semantically relevant features tailored to the requirements of downstream tasks. Another research direction to explore is the assumption that shared information is always reliable. However, a vehicle with degraded or malfunctioning sensors may unintentionally transmit erroneous data. Therefore, fusion mechanisms should incorporate verification steps to evaluate the credibility of incoming information and avoid propagating low-quality inputs. \\

\item \textbf{V2V, V2I and V2X systems.} 

Different collaboration modes pose distinct challenges requiring tailored solutions. V2V communication supports rapid, localized data sharing but demands future research on bandwidth-efficient, context-aware protocols that prioritize and transmit only task-relevant information in real time. V2I communication introduces centralized sensing via infrastructure (e.g., roadside units), offering a broader view of the environment. Research should focus on adapting V2V methods to V2I setups, while addressing communication speed and volume challenges. V2X communication, which encompasses V2V, V2I, and interactions with other entities such as pedestrians, networks, and IoT devices, introduces added complexity related to trust, data synchronization, and interoperability. Future research must move beyond simulations and prioritize real-world testing to validate V2X reliability and scalability in dynamic traffic systems.   \\

\item \textbf{Integration of Prediction Models.}

Trajectory prediction models find integration in both fully autonomous vehicles and modern advanced driver-assistance systems (ADAS), contributing significantly to road safety. In full-stack autonomy, prediction modules inform planning about potential collisions. Automotive companies like Waymo and Cruise integrate multi-agent forecasting in their planning pipelines\footnote{\url{https://waymo.com/research/motionlm/}}. In ADAS\footnote{\url{https://www.mobileye.com/blog/}} prediction models help infer lane-change intent and pedestrian movement, enabling early warnings that give human drivers more time to maneuver and avoid collisions. Another application involves integrating prediction with emergency braking systems, as seen in Mercedes-Benz vehicles, which use predictive cues to autonomously stop when drivers respond too late\footnote{\url{https://group.mercedes-benz.com/innovation/product-innovation/technology/pedestrian-emergency-braking-system.html}}, contributing to enhanced vehicle safety \citep{Korneev_Niu_Ibrahim_2024}. Future research should continue on integrating prediction models in traffic management systems to enable proactive safety interventions.

\end{enumerate}  
\section{Conclusion}
\label{sec:conclusion}

In this work, we have conducted an in-depth review of trajectory prediction approaches for autonomous driving, providing a comprehensive taxonomy to classify existing methodologies. This taxonomy distinguishes between learning-based and non-learning methods, with learning-based techniques further categorized into machine learning, deep learning, and reinforcement learning approaches. Additionally, we broadened the scope of the review to include a detailed overview of the prediction pipeline, covering input and output modalities, benchmark datasets used for evaluation, and the performance metrics critical for assessing prediction models.

Beyond the taxonomy and evaluation, this study discussed the pressing challenges and evolving trends in the field. Sections on robustness and generalizability emphasize the need for uncertainty quantification and mitigation, while collaborative perception is discussed as a promising solution for addressing limitation of individual perception such as occlusions, sensor noise, and restricted sensing capabilities. The review also explored planning-conditioned trajectory prediction, which incorporates motion planning principles to model rational agent behavior. Methods such as goal-anchor-based prediction play a pivotal role by estimating potential destinations and generating trajectories aligned with those goals, ensuring more interpretable and goal-driven outputs. Additionally, the integration of VLMs into prediction pipelines offers an exciting frontier, providing common-sense knowledge and contextual understanding to enhance decision-making in uncertain conditions. However, challenges like latency and scalability of these models require further investigation. Furthermore, the study underscores the importance of designing systems that generalize across diverse datasets and scenarios, addressing the existing gap in cross-dataset evaluations and qualitative assessments.

We have addressed the posed research questions, providing a comprehensive summary of the state of the art while identifying potential future research directions. Despite significant advancements over the past two decades, trajectory prediction remains an open and challenging area of research. The field has evolved from early physics-based models to Graph Transformers. The shift toward combining domain knowledge with data-driven methods, such as physics-based hybrid models, represents a step forward in balancing interpretability with predictive capacity. This review not only presented the evolution of trajectory prediction methodologies but also shed the light on key research gaps and directions for further exploration in trajectory forecasting for autonomous driving.



%%       \citet{<label>} ==> Jones et al. [21]
%%       \citep{<label>} ==> [21]
%%

\bibliographystyle{elsarticle-num-names} 
\bibliography{refs}


%\begin{thebibliography}{00}



%\end{thebibliography}
\end{document}